\documentclass{article}

\usepackage[final]{neurips_2020}

\usepackage[utf8]{inputenc} 
\usepackage[T1]{fontenc}    
\usepackage{hyperref}       
\usepackage{url}            
\usepackage{booktabs}       
\usepackage{amsfonts}       
\usepackage{nicefrac}       
\usepackage{microtype}      
\usepackage{footmisc}

\usepackage{graphicx}
\usepackage{amsmath}
\usepackage{amssymb}
\usepackage{pifont}
\usepackage{enumitem}       
\usepackage{algorithm}
\usepackage[noend]{algpseudocode}

\newcommand{\scf}{\textsc{f}}
\newcommand{\scg}{\textsc{g}}
\newcommand{\scw}{\textsc{w}}
\newcommand{\scp}{\textsc{p}}
\newcommand{\scs}{\textsc{s}}

\newcommand{\cmark}{\ding{51}}
\newcommand{\xmark}{\ding{55}}

\title{Neural Topographic Factor Analysis for fMRI Data}

\author{%
  Eli Sennesh \thanks{Equal contribution} \textsuperscript{1,3},  Zulqarnain Khan\footnotemark[1] \textsuperscript{2},  Yiyu Wang\textsuperscript{3} \\
  \textbf{Jennifer Dy\textsuperscript{2}, Ajay Satpute\textsuperscript{3}, J. Benjamin Hutchinson\textsuperscript{4}, Jan-Willem van de Meent\textsuperscript{1}} \\
  \texttt{sennesh.e@northeastern.edu}, \texttt{khan.zu@ece.neu.edu}, \texttt{wang.yiyu@northeastern.edu} \\
  \texttt{jdy@ece.neu.edu}, \texttt{a.satpute@northeastern.edu}, \texttt{bhutch@uoregon.edu}, \\
  \texttt{j.vandemeent@northeastern.edu} \\
  \textsuperscript{1} Khoury College of Computer Sciences, Northeastern University \\
  \textsuperscript{2} Department of Electrical and Computer Engineering, Northeastern University \\
  \textsuperscript{3} Department of Psychology, Northeastern University \\
  \textsuperscript{4} Department of Psychology, University of Oregon
}

\begin{document}

\maketitle

\begin{abstract}
Neuroimaging studies produce gigabytes of spatio-temporal data for a small number of participants and stimuli. Rarely do researchers attempt to model and examine how individual participants vary from each other -- a question that should be addressable even in small samples given the right statistical tools.  We propose Neural Topographic Factor Analysis (NTFA), a probabilistic factor analysis model that infers embeddings for participants and stimuli. These embeddings allow us to reason about differences between participants and stimuli as signal rather than noise. We evaluate NTFA on data from an in-house pilot experiment, as well as two publicly available datasets. We demonstrate that inferring representations for participants and stimuli improves predictive generalization to unseen data when compared to previous topographic methods. We also demonstrate that the inferred latent factor representations are useful for downstream tasks such as multivoxel pattern analysis and functional connectivity.
\vspace{-1em}
\end{abstract}

\section{Introduction}
\vspace{-0.5em}
Analyzing functional neuroimaging studies is both a large data problem and a small data problem.  A single scanning run typically comprises hundreds of full-brain scans that each consist of tens of thousands of spatial locations (known as voxels). At the same time, neuroimaging studies tend to have limited statistical power \citep{10.1371/journal.pone.0184923}; a typical study considers a cohort of 20-50 participants undergoing tens of stimuli from ten (or fewer) stimulus categories. This poses a significant problem for the over fourteen-thousand functional neuroimaging studies that seek to address both fundamental and translational research questions in cognitive neuroscience on individual differences in functional neural activity \citep{Elliott2020}. A largely unsolved challenge in this domain is to develop analysis methods that appropriately account for both the commonalities and variations among participants and stimuli effects, scale to tens of gigabytes of data, and reason about uncertainty.

In this paper, we develop Neural Topographic Factor Analysis (NTFA)\footnote{Source code submitted with paper and available upon request.}, a generative model for neuroimaging data that explicitly represents variation among participants and stimuli. NTFA extends Topographic Factor Analysis (TFA) and Hierarchical Topographic Factor Analysis (HTFA) \citep{manning2014topographic,manning2018probabilistic}. It differs from these models in that it learns a prior that maps embeddings (i.e.~vectors of features) for each participant and stimulus to a conditional distribution over spatial factors and weights, instead of imposing a single global prior. The result is a structured probabilistic model that learns a representation of each participant and each stimulus.

NTFA offers two advantages over other dimensionality reduction methods that project data into a low-dimensional space: Our embeddings factorize the generative contributions of participants from those of stimuli, and they supply uncertainty measures by which we can measure the scale of the embedding space.  Having the embedding space ``scaled'' by uncertainty allows us more confidence in resolving differences: if the means for embeddings of stimuli lie several standard deviations apart from each other, we can be confident they reflect significant differences in the neural repsonse.

We perform a qualitative evaluation of inferred embeddings on four datasets: 
\begin{itemize}[labelwidth=0.5em,labelsep=0.5em,leftmargin=1.0em,topsep=0em,label={\tiny\raisebox{0.7ex}{\textbullet}}]
    \item We show that in a synthetic dataset, simulated from distinguishable clusters of participants and stimuli, inference recovers the underlying cluster structure.
    \vspace{-.5em}
    
    \item We present results for our own pilot study investigating whether threat-relevant stimuli from three categories induce the same or different patterns of neural response. NTFA infers stimulus embeddings that show differences in patterns of neural response between stimulus categories.
    \vspace{-.5em}
    
    \item We analyze and evaluate two publicly available datasets. In the first, participants with major depressive disorder and controls listened to emotionally valenced sounds and music \citep{10.1371/journal.pone.0156859}.  In the second, participants viewed images of faces, cats, five categories of man-made objects, and scrambled pictures \citep{Haxby2425}. In both cases, NTFA infers an embedding structure that is consistent with previously reported findings. 
\end{itemize}
Because NTFA is, to our knowledge, the first model to explicitly infer embeddings for participants and stimuli, we devise two simple baselines as comparisons. The first is to apply PCA directly to the input data, and the second is to compute post-hoc embeddings after training a shared response model (SRM) \citep{chen2015reduced-dimension}. PCA fails to recover participant and stimulus structure, whereas the SRM yields point-estimates that are qualitatively similar but lack uncertainty estimates.

As a sanity check, we also compare predictive performance on a validation set of brain images across NTFA and HTFA.  We hold out trials by their stimulus-participant pairs, requiring our model to generalize from other trials in which the same stimulus or participant were seen.  PCA, the SRM, and TFA cannot recombine representations to predict such novel combinations in this way.
\begin{figure*}[!t]
    \begin{center}
    \includegraphics[width=0.95\textwidth]{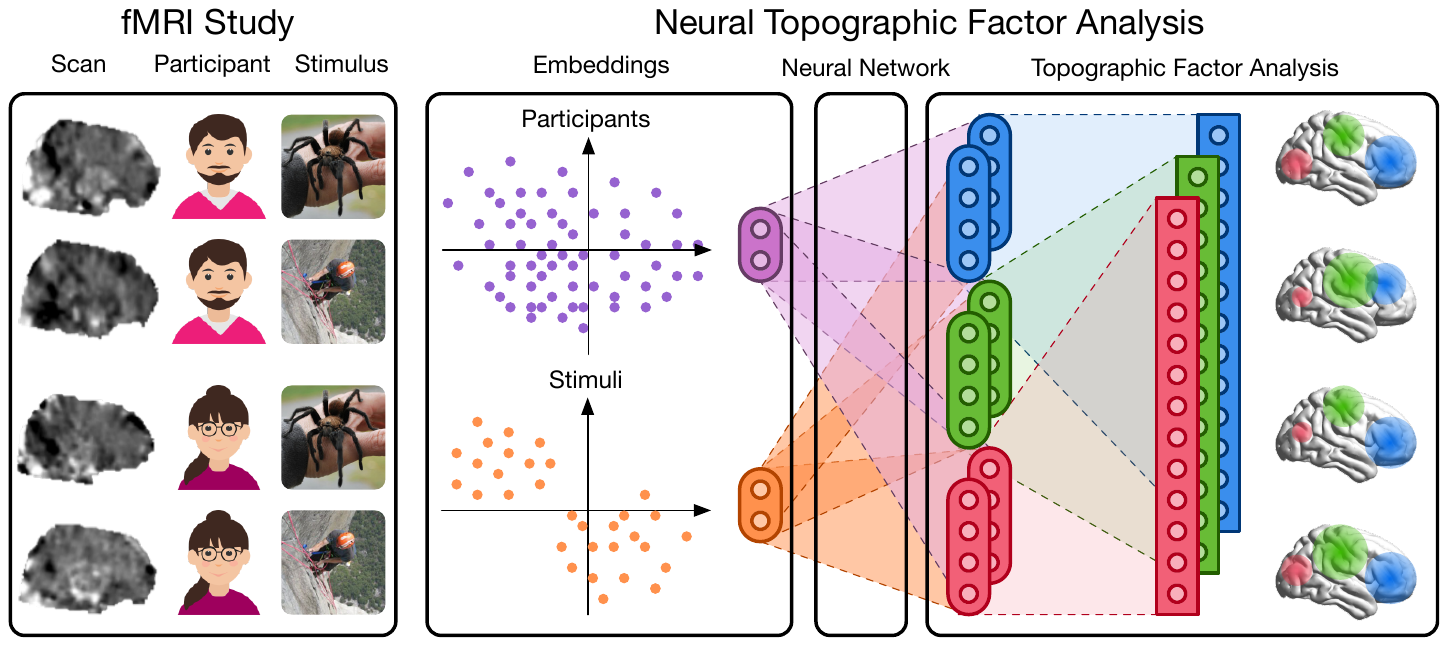}
    \end{center}
    \vspace{-0.75em}
    \caption{\textbf{Overview of Neural Topographic Factor Analysis (NTFA)}: We decompose the fMRI signal into Gaussian factors (shown in red, green and blue in the figure) that correspond to spatially and temporally related brain activity across individuals. A typical fMRI study consists of multiple trials of participants undergoing scans while experiencing different stimuli (or performing different tasks). In our generative model we represent these participants (purple) and stimuli (orange) with embedding vectors. A multilayer perceptron then predicts the factors' location, size, and weights.}
    \label{fig:overview}
\end{figure*}

This work makes both neuroscientific and machine learning contributions. From a machine learning perspective, NTFA is a novel neural extension of probabilistic factor analysis methods. The inferred embeddings capture similarities in the neural response across participants and stimuli. This improves prediction on held-out data, while requiring fewer trainable parameters. From a neuroscientific perspective, the generative model in NTFA contributes to our ability to characterize individual variation in whole-brain analyses. Psychological states (e.g.~emotions and memories) involve patterns of activation distributed widely throughout voxel space \citep{Haxby2425,Satpute2019TICS}. Existing whole-brain analyses such as multivoxel pattern analysis (MVPA) thus often rely on supervised feature selection using labels for stimulus categories or participant groups\citep{Pereira2009}. In contrast, latent factors from NTFA and HTFA enable unsupervised whole-brain MVPA, and can be used to create data-driven functional connectomes.

Figure~\ref{fig:overview} outlines our proposed approach. Section~\ref{sec:background} covers related work in factor analysis for neuroimaging data, primarily the spatially topographic methods on which we build. Section~\ref{sec:ntfa-basic} develops the NTFA model. Section~\ref{sec:evaluation} discusses our architectural details, preprocessing steps, and experiments, then discusses and evaluates experimental results.  Section~\ref{sec:conclusion} concludes.
\vspace{-1em}
\section{Background}
\label{sec:background}
\vspace{-1em}

\begin{table}[!t]
    \caption{Comparison of factor analysis methods for fMRI data.  When a method considers participant and stimulus variations dependently, we consider it to model variation in the independent factor. Our method (NTFA) is shown in the bottom row.}
    \label{tab:fa_features}
    \centering
    \begin{tabular}{c|c|c|c|c}
        \toprule
        Model & Spatial factors & Participant variation & Stimulus variation & Scanning run variation\\
        \midrule
        PCA & \xmark & \xmark & \xmark & \xmark \\
        SRM & \xmark & \cmark & \xmark & \cmark \\
        TFA & \cmark & \xmark & \xmark & \xmark \\
        HTFA & \cmark & \xmark & \xmark & \cmark \\
        TLSA & \cmark & \cmark & \xmark & \xmark \\
        \midrule 
        NTFA & \cmark & \cmark & \cmark & \cmark \\
        \bottomrule
    \end{tabular}
\end{table}

Factor analysis methods are widely used to reduce the dimensionality of neuroimaging data. These methods decompose the fMRI signal for a trial $Y \in \mathbb{R}^{T\text{x}V}$ with $T$ time points and $V$ voxels into a product $Y \simeq W F$ between a lower-rank matrix of weights $W \in \mathbb{R}^{T\text{x}K}$ and a lower-rank matrix of factors $F \in \mathbb{R}^{K\text{x}V}$. The dimension $K \ll V$ is chosen to balance the degree of dimensionality reduction and the reconstruction accuracy. 

Standard methods that are applied primarily to fMRI data include Principal Component Analysis \citep{abdi2010principal} and Independent Component Analysis \citep{hyvarinen2001independent}.
There are also methods that have been specifically developed for fMRI analysis. These include  adaptations of dictionary learning methods for large-scale fMRI datasets \citep{mensch2017learning}, hyper-alignment (HA) \citep{haxby2011common}, the shared response model (SRM) \citep{chen2015reduced-dimension} and the matrix-normal shared response model (MN-SRM) \citep{shvartsman2018matrix} for functional alignment, as well as topographic latent source analysis (TLSA) \citep{gershman2011topographic,Gershman2014}, topographic factor analysis \citep{manning2014topographic}, and hierarchical topographic factor analysis \citep{manning2014hierarchical,manning2018probabilistic} for functional connectivity.  Many of these come in probabilistic varieties \citep{CAI2020107500}.

Topographic models define spatially smooth factors via radial basis functions.  Non-topographical models - such as PCA, ICA, the SRM, the MN-SRM and dictionary learning - learn a $V\times K$ factor-voxel loading matrix\footnote{In some cases for each subject, requiring a total of $V K P$ parameters to be learned for a study with $P$ participants, or each stimulus, requiring $V K S$ parameters, where $K$ is the number of factors and $S$ is the number of unique stimuli.}, with no requirement of spatial smoothness.  Some of these learn factor-loading weights for participants, such as the SRM, while others such as dictionary learning learn factor-loading weights for experimental conditions or stimuli.  HA can be considered a special case of the SRM with $K=V$. The MN-SRM is also very similar to the SRM except it enforces a weaker Gaussianity constraint on the factor-loading weights instead of the orthonormality constraint in SRM. Table~\ref{tab:fa_features} compares our model's latent factorization structure to those of other models.  NTFA is novel in learning independent low-dimensional embeddings for both participants and stimuli.
\vspace{-0.5em}
\subsection{Topographic Factor Analysis}
\label{subsec:tfa-htfa-models}
\vspace{-0.5em}

This work extends TFA and HTFA, two probabilistic models that employ radial basis functions to represent spatial factors. TFA and HTFA model data that comprises $N$ trials (i.e.~continuous recordings), each of which contain $T$ time points for voxels at $V$ spatial positions. TFA approximates each trial separately $Y_{n} \simeq W_{n} F_{n}$ as a product between time-varying weights $W_n \in \mathbb{R}^{T \times K}$ and spatially-varying factors $F_n \in \mathbb{R}^{K \times V}$. To do so, TFA assumes that the data is noisily sampled from the inner product between the weights and factors matrices
\begin{align}
    Y_{n} \sim \mathcal{N}\left(W_{n} F_{n},\sigma^Y\right).
\end{align}
TFA combines this likelihood $p(Y_n \,|\, W_n, F_n)$ with the prior $p(W_n, F_n)$ to define a probabilistic model $p(Y_n, W_n, F_n)$. Black-box methods are then used to approximate the posterior $p(W_n,F_n \,|\, Y_n)$ with a mean-field variational distribution $q_{\lambda}(F_n, W_n)$ on the factors $W_n$ and $F_n$.

The prior $p(W_n,F_n)=p(W_n)\:p(F_n)$ factorizes over $W_n$ and $F_n$. The prior over weights $p(W_n)$ is a hierarchical Gaussian with hyperparameters $\mu^\textsc{w}_{n,k}$ and $\sigma^\textsc{w}_{n,k}$,
\begin{align}
	W_{n,t,k}
    &\sim 
    \mathcal{N}(\mu^\scw_{n,k}, \sigma^\scw_{n,k}),
    &
	\mu^\scw_{n,k}
   	&\sim 
    p(\mu^\scw),
    &
	\sigma^\scw_{n,k} 
   	&\sim 
    p(\sigma^\scw),
\end{align}
To define a prior over factors $p(F_n)$, TFA employs a kernel function that ensures spatial smoothness of factor values $F_{n,k,v}$ at nearby voxel positions $x^\scg_v \in \mathbb{R}^3$. This kernel function $\kappa$ is normally a radial basis function (RBF), which models each factor $k\in \{1\ldots K\}$ as a Gaussian with center at a spatial location $x_{n,k}^\scf \in \mathbb{R}^3$, whose width is determined by the kernel hyper-parameters $\rho^\scf_{n,k}$, 
\begin{align}
	\label{eq:tfa-factor-kernel}
	F_{n,k,v}
    &= 
    \kappa(x^\scg_v, x^\scf_{n,k} \,;\, \rho^\scf_{n,k})
    ,
    &
	x^\scf_{n,k} 
    &\sim 
    p(x^\scf),
    &   
	\rho^\scf_{n,k} 
    &\sim 
    p(\rho^\scf).
\end{align}
Interpreting factor analysis generatively enables us to incorporate additional assumptions to capture  similarities across a set of related trials. HTFA \citep{manning2014hierarchical,manning2018probabilistic}, introduces variables $\bar{x}^\scf_k$ and $\bar{\rho}^\scf_k$ representing each factor's mean positions and widths across trials, 
\begin{align} 
    x^\scf_{n,k} 
    &\sim 
    p(x^\scf_{n,k} \mid \bar{x}^\scf_{k})
    ,
    &
    \bar{x}^\scf_{k} 
    &\sim
    p(\bar{x}^\scf)
    ,
    &
    \rho^\scf_{n,k} 
    &\sim 
    p(\rho^\scf_{n,k} \mid \bar{\rho}^\scf_{k})
    ,
    &
    \bar{\rho}^\scf_{k} 
    &\sim
    p(\bar{\rho}^\scf).
\end{align}
HTFA assumes that brain layouts and activations across trials vary around a shared Gaussian prior. This imposes unimodality upon the distribution of neural responses across trials.

\vspace{-0.5em}
\section{Neural Topographic Factor Analysis}
\label{sec:ntfa-basic}

NTFA extends TFA to model variation across participants and stimuli. We assume the same factor analysis model as TFA, which approximates the fMRI signal as a linear combination of time-dependent weights and spatially varying Gaussian factors. NTFA extends TFA by inferring \emph{embedding vectors} for individual participants and stimuli. We learn a mapping from embeddings to the parameters of the likelihood model, parameterized by a neural network. Instead of HTFA's global template, we introduce factorized latent spaces of participant and stimulus embeddings, and share the neural networks mapping embeddings to factors.  For reference, a complete description of our notation can be found in Table~\ref{app:tb_not} in the Appendix.

The advantage of incorporating neural networks into the generative model is that it enables us to explicitly reason about multimodal response distributions and effects that vary between individual samples.  The network weights $\theta$ are shared across trials, as are the stimulus and participant embeddings $z^\scs_s$ and $z^\scp_p$. This allows NTFA to capture statistical regularities within a whole experiment.  At the same time, the use of neural networks ensures that differences in embeddings can be mapped onto a wide range of spatial and temporal responses.  Whereas the hierarchical Gaussian priors in HTFA implicitly assume that response distributions are unimodal and uncorrelated across different factors $k\in[K]$, the neural network in NTFA can model such correlations by jointly predicting all $K$ factors.

We model $N$ trials in which participants $p_n \in \{1, \ldots, P\}$ undergo a set of stimuli $s_n \in \{1, \ldots, S\}$ and are scanned for $T$ time points per trial. We assume that participant embeddings $\{z^\scp_1, \ldots, z^\scp_P\}$ and stimulus embeddings $\{z^\scs_1, \ldots, z^\scs_S\}$ are shared across trials. For simplicity, we will consider the case where both embeddings have the same dimensionality $D$ and a Gaussian prior
\begin{align}
    \label{eqs:sample_embeddings}
    z_{p}^\scp &\sim \mathcal{N}(0,I), 
    &
    z_{s}^\scs &\sim \mathcal{N}(0,I).
\end{align}
For each participant $p$, we define the RBF center $x^{\scf}_p$ and log-width $\rho^{\scf}_p$ in terms of a neural mapping
\begin{align}
    \label{eq:factor_mean}
    x^{\scf}_p &\sim \mathcal{N}(\mu^{x}_{p}, \sigma^{x}_{p}),
    &
    \mu^{x}_{p}, \sigma^{x}_{p} &\gets \eta^{\scf,x}_\theta (z^\scp_p),
    \\
    \label{eq:factor_width}
    \rho^{\scf}_{p} &\sim \mathcal{N}(\mu^{\rho}_{p}, \sigma^{\rho}_{p}),
    &
    \mu^{\rho}_{p}, \sigma^{\rho}_{p} &\gets \eta^{\scf,\rho}_\theta (z^\scp_p).
\end{align}
Here $\eta^{\scf}_\theta$ is a neural network parameterized by a set of weights $\theta$, which models how variations between participants and stimuli affect the factor positions and widths in brain activations.  This network outputs a $K\times 4\times 2$ tensor, that contains a two-tuple of three-dimensional parameters for each factor center $(\mu^x_{p}, \sigma^x_{p})$ and another two-tuple of one-dimensional parameters for each factor log-width$(\mu^\rho_{p}, \sigma^\rho_{p})$. We use a second network $\eta^\scw_\theta(z^\scp_p, z^\scs_s)$ to parameterize the distribution over weights $W_{n,t}$ with a $K \times 2$ tensor, given the embeddings for each trial $n$ and time point $t$ with $p=p_n,s=s_n$:
\begin{align}
    \label{eq:weight_nn}
    W_{n,t} &\sim \mathcal{N} \left( \mu^{\textsc{w}}_{n}, \sigma^{\textsc{w}}_{n} \right),
    &
    \mu^{\textsc{w}}_{n}, \sigma^{\textsc{w}}_{n} 
    &\leftarrow 
    \eta^\scw_\theta \left( z^\scp_p, z^\scs_s \right).
\end{align}
The likelihood model is the same as that in TFA,
\begin{align}
    \label{eq:image_generate}
    Y_{n,t} &\sim \mathcal{N}\big( W_{n,t} \cdot F_{p}, \sigma^\textsc{y}\big),
    &
    F_{p} &\leftarrow \kappa(x^\scf_{p}, \rho^{\scf}_{p}).
\end{align}

\begin{algorithm}[t]
   \caption{NeuralTFA Generative Model}
   \label{alg:ntfa_generative}
\begin{algorithmic}[1]
    \Statex {($p_1, \ldots, p_N$)} \Comment{Participant for each trial}
    \Statex {($s_1, \ldots, s_N$)} \Comment{Stimulus for each trial}
\For{$p$ \textbf{in} $1, \ldots, P$}
    \State $z_{p}^\scp \sim \mathcal{N}(0,I)$ 
    \Comment{Equation~\eqref{eqs:sample_embeddings}}
\EndFor
\For{$s$ \textbf{in} $1, \ldots, S$}
    \State $z_{s}^\scs \sim \mathcal{N}(0,I)$
    \Comment{Equation~\eqref{eqs:sample_embeddings}}
\EndFor
\For{$n$ \textbf{in} $1, \ldots, N$} 
    \State $p,s \leftarrow p_n, s_n$
    \State $\left( \mu^{x}_{p}, \sigma^{x}_{p}\right), \left( \mu^{\rho}_{p} , \sigma^{\rho}_{p} \right) \leftarrow \eta^\scf_\theta (z^\scp_p)$
    \State $x^\scf_p \sim \mathcal{N}(\mu^{x}_{p}, \sigma^{x}_{p})$
    \Comment{Equation~\eqref{eq:factor_mean}}
    \State $\rho^\scf_{p} \sim \mathcal{N}(\mu^{\rho}_{p}, \sigma^{\rho}_{p})$
    \Comment{Equation~\eqref{eq:factor_width}}
    \State $\mu^{\textsc{w}}_{n}, \sigma^{\textsc{w}}_{n} \leftarrow \eta^\scw_\theta \left( z^\scp_p, z^\scs_s \right)$
    \Comment{Equation~\eqref{eq:weight_nn}}
    \For{$t$ \textbf{in} $1 \ldots T$}
        \State $W_{n,t} \sim \mathcal{N}(\mu^{\textsc{w}}_{n}, \sigma^{\textsc{w}}_{n})$ 
        \Comment{Equation~\eqref{eq:weight_nn}}
        \State $F_{p} \leftarrow \kappa(x^\scf_p, \rho^\scf_p)$
        \State $Y_{n,t} \sim \mathcal{N}(W_{n,t}\cdot F_p, \sigma^Y)$
        \Comment{Equation~\eqref{eq:image_generate}}
    \EndFor
  \EndFor
\end{algorithmic}
\end{algorithm}

We summarize the generative model for NTFA in Algorithm~\ref{alg:ntfa_generative}. This model defines a joint density $p_{\theta}(Y, W, x^\scf, \rho^\scf, z^\scp, z^\scs)$, which in turn defines a posterior $p_{\theta}(W, x^\scf, \rho^\scf, z^\scp, z^\scs \mid Y)$ when conditioned on $Y$. We approximate the posterior with a fully-factorized variational distribution,
\begin{align}
    \label{eq:mean-field-variational}
    \begin{split}
    q_{\lambda}(W, \rho^\scf, x^\scf, z^\scp, z^\scs)
    &=
    \prod_{n,t} 
    q_{ \lambda^\scw_{n,t}}(W_{n,t})
    \prod_{s}
    q_{\lambda^\scs_{s}}(z^\scs_s) 
    \prod_{p}
    q_{\lambda^{\textsc{x}^\scf_{p}}}(x^\scf_p) \:
    q_{\lambda^{\rho^\scf_{p}}}(\rho^\scf_p) \:
    q_{\lambda^\scp_{p}}(z_p).
    \end{split}
\end{align}
We learn the parameters $\theta$ and $\lambda$ by maximizing the evidence lower bound (ELBO)
\begin{align*}
    \mathcal{L}(\theta, \lambda)
    = 
    \mathbb{E}_{q}
    \left[
    \log 
    \frac{p_{\theta}(Y, W, x^\scf, \rho^\scf, z^\scp, z^\scs)}
         {q_{\lambda}(W, x^\scf, \rho^\scf, z^\scp, z^\scs)}
    \right]
    \le 
    \log p_\theta(Y).
\end{align*}
We optimize this objective using black-box methods provided by Probabilistic Torch, a library for deep generative models that extends the PyTorch deep learning framework  \citep{narayanaswamy2017learning}. Specifically, we maximize an importance-weighted bound \citep{Burda2016} using a doubly-reparameterized gradient estimator \citep{Tucker2019}. This objective provides more accurate estimates for the gradient of the log marginal likelihood.

While neural network models can have thousands or even millions of parameters, we emphasize that  NTFA in fact has a \emph{lower} number of trainable parameters than HTFA. This follows from the fact that TFA and HTFA assume fully-factorized variational distributions that have $O(NK+NTK)$ parameters for $N$ trials with $T$ time points. In NTFA, the networks $\eta^\scf$ and $\eta^\scw$ have $O(D(D+K))$ parameters each, whereas the variational distribution has $O(D(P+S)+PK+NTK)$ parameters.

In practice, scanning time limitations impose a trade-off between $N$ and $T$. For this reason $NTK$ does not always dominate $NK$, since often $T\propto O(10)$.  We can then choose $D\propto O(1)$ and $K \propto O(100)$, and if we label constant factors $c$, the total number of parameters becomes $O(cD^2 + cDK)$, making $O(cDK)$ the dominant term.  When $P \ll N$, as is usually the case, NTFA can therefore have orders of magnitude fewer parameters than HTFA for $D=2$.

\vspace{-0.5em}
\section{Evaluation}
\label{sec:evaluation}
\vspace{-0.75em}
\subsection{Datasets}
\label{subsec:datasets}
\vspace{-0.5em}

We consider four datasets in our experiments. First, we create a simulated dataset to verify that NTFA can recover a ground-truth structure in data that, by construction, contains clearly distinguishable participant and stimulus clusters (labelled ``Synthetic''). Second, we analyze a previously unpublished data from a pilot study, conducted by one of the authors, that measures the neural response to threat-relevant stimuli (labelled ``ThreatVids'').
Third, we analyze a publicly available dataset on valenced sounds and music, with participants divided into a control group and a depressed group \citep{10.1371/journal.pone.0156859} (labelled ``Lepping''). Finally, we verify that NTFA can reconstruct a popular publicly available dataset of participants watching pictures of animate and inanimate objects \citep{Haxby2425} (labelled ``Haxby'').  These experimental datasets vary in their number of participants, time points, voxels, and task variables. A detailed description of each of these datasets can be found in Appendix~\ref{sec:si_datasets}, and our standard neuroimaging preprocessing pipeline is discussed in Appendix~\ref{sec:si_preprocessing}.  

\vspace{-0.5em}
\subsection{Model Architecture and Training}
\label{subsec:model_architecture}
\vspace{-0.5em}

We employ participant and stimulus embeddings with $D=2$ in all experiments.  For the synthetic dataset, we analyze the data with the same number of factors as were used to generate it, $K=3$. For non-simulated data we use $K=100$ factors. This is somewhat fewer than previously reported for HTFA ($K=700$) \citep{manning2018probabilistic} owing to GPU memory limitations.  
We report parameter counts for HTFA and NTFA in Table~\ref{table:reconstruction_errors}, and provide details on network architectures in Appendix~\ref{subsec:neural-net-architectures}.

\vspace{-0.75em}
\subsection{Generalization to Held-out Images}
\label{subsec:generalization-quality}
\vspace{-0.5em}

To evaluate generalization, we split datasets into training and test sets, ensuring the training set contains at least one trial for each participant $p \in \{1,\ldots,P\}$ and each stimulus $s \in \{1,\ldots,S\}$.  To do so, we construct a matrix of $(p,s) \in \{1,\ldots,P\}\times \{1,\ldots,S\}$ with participants as rows and stimuli as columns.  We then choose all trials along the matrix's diagonals $\{n : p_n \bmod S = s_n \}$ as our test set. All other trials are used as the training set.

We evaluate generalization to held-out data in terms of the posterior-predictive probability 
\begin{align*}
    p_\theta(\tilde{Y} \mid Y) = \int p_\theta(\tilde{Y} \mid z^\scp, z^\scs) \: p_\theta(z^\scp, z^\scs \mid Y) \: d z^\scp \: dz^\scs .
\end{align*}
Like the marginal likelihood, this quantity is intractable. We approximate it by computing a VAE-style lower bound  $\mathbb{E}[\tilde{\mathcal{L}}] \le \log p_\theta(\tilde{Y} \mid Y)$ from $L$ samples (see Appendix~\ref{app:log-predictive-bound} for a derivation),
\begin{equation}
\begin{aligned}
    \tilde{\mathcal{L}}
    =
    \frac{1}{L} \sum_{l=1}^L
    \log 
    p \big(
        \tilde{Y} 
        \mid 
        \tilde{W}^{(l)} \!, 
        \tilde{x}^{\scf \, (l)}  \!, 
        \tilde{\rho}^{\scf \, (l)} \!, 
        z^{\scp \, (l)} \!, 
        z^{\scs \, (l)}
    \big).
\end{aligned}
\label{eq:post_pred_lb}
\end{equation}
We sample embeddings from the variational distribution and remaining variables from the prior
\vspace{-.1em}
\begin{align*}
    z^{\scp \, (l)} &\sim q(z^\scp),
    &
    z^{\scs \, (l)} &\sim q(z^\scs),
    &
    \tilde{W}^{(l)}, \tilde{x}^{\scf \, (l)}, \tilde{\rho}^{\scf \, (l)}
    &\sim 
    p_\theta\big(
        \tilde{W}, \tilde{x}^{\scf}, \tilde{\rho}^{\scf}
        \mid  
        z^{\scp \, (l)},
        z^{\scs \, (l)}
    \big)
    .
\end{align*}

\begin{table}[!b]
\caption{\textbf{Generalization performance} (in log predictive probability) and \textbf{parameter counts}. We approximate the log predictive with a VAE-style lower bound. We evaluate on a test set of held out subject-stimuli pairs, and use $K=100$ factors across datasets and models.  For NTFA we set $D=2$.}
\label{table:reconstruction_errors}
\centering
\begin{tabular}{lcccc}
\toprule
 & \textbf{Log-predictive} & \textbf{Log-predictive} & \textbf{Parameter count} & \textbf{Parameter count} \\
 & \textbf{HTFA} & \textbf{NTFA} & \textbf{HTFA} & \textbf{NTFA} \\
\midrule
~Synthetic ($K=3$) & $-4.72 \times 10^6$ & $\mathbf{-4.68 \times 10^6}$ & $2.16\times10^4$ & $1.90\times10^4$ \\
~ThreatVids & $-2.23\times 10^9$ & $\mathbf{-2.19 \times 10^9}$ & $1.64\times10^8$ & $8.88\times 10^6$ \\
~Lepping & $-2.54 \times 10^9$ &  $\mathbf{-2.47 \times 10^9}$ & $2.53\times 10^7$  & $2.61 \times 10^6$ \\
~Haxby & $-7.17 \times 10^8$ & $\mathbf{-7.10 \times 10^8}$ & $1.44 \times 10^6$ & $1.01 \times 10^6$ \\
\bottomrule
\end{tabular}
\end{table}

In Table~\ref{table:reconstruction_errors}, we compare NTFA to HTFA in terms of log predictive probability for held-out data, computing an importance-weighted bound over each dataset's test set. Across datasets, NTFA exhibits a higher log-likelihood and log-predictive probability than HTFA, with the same number ($K=100$) of latent factors. We observe larger improvements by NTFA over HTFA in datasets such as ThreatVids, in which NTFA shares statistical strength across trials since $N \gg P$ and $N \gg S$.

We visualize the posterior-predictive means for held-out trials in Appendix~\ref{app:additional-predictions}.  HTFA shares its predictive distribution across all trials, because it lacks any explicit representation of participants and stimuli.  By contrast, NTFA's predictive representation more closely resembles the actual data.

\vspace{-0.75em}
\subsection{Inferred Embeddings}
\label{subsec:inferred-embeddings}
\vspace{-0.5em}

NTFA infers embeddings for both participants and stimuli, along with estimates of uncertainty for those embeddings.  Participant embeddings appeared to primarily reflect idiosyncratic differences among participants, without mapping clearly to any participant conditions or behavior available in the datasets. We discuss these participant embeddings in detail in Appendix~\ref{app:participant_embeddings}. Here we discuss the extent to which stimulus embeddings align with with experimenter-defined categories.

\begin{figure*}[!t]
    \centering
    \begin{tabular}{cc}
        \textsf{\textbf{Underlying}} & \textsf{\textbf{Inferred}} \\
    \includegraphics[width=0.45\textwidth]{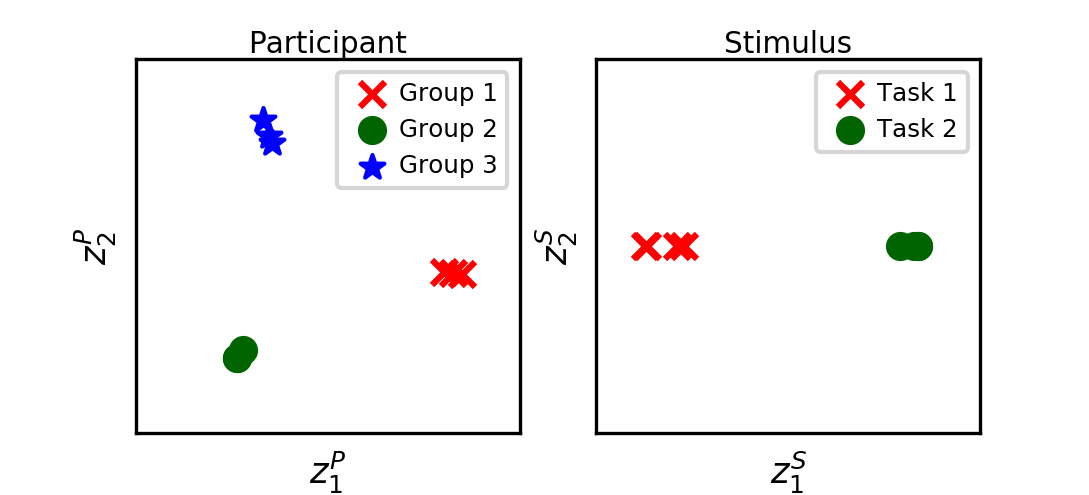} &
    \includegraphics[width=0.45\textwidth]{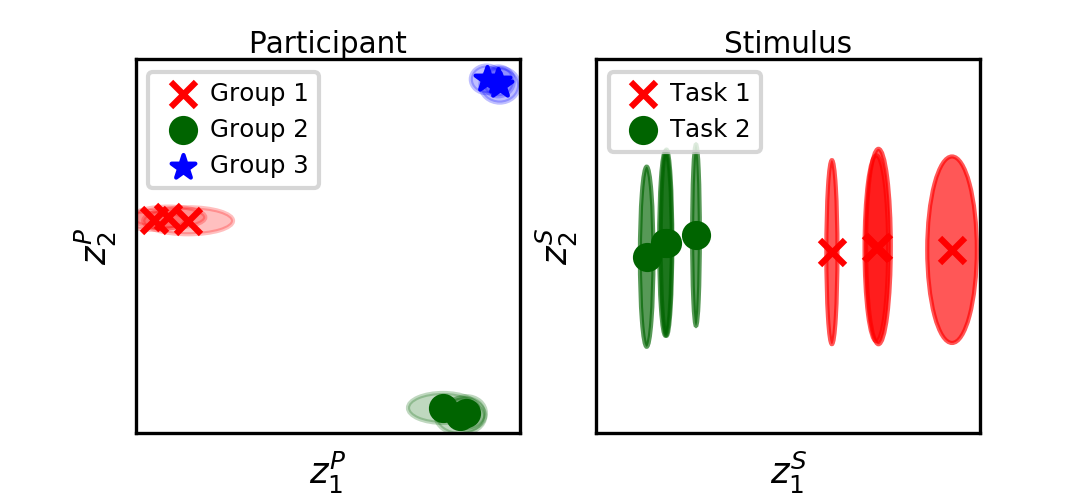}
    \end{tabular}
    \vspace{-1em}
    \caption{\textbf{Inferred embeddings for synthetic data}: \textbf{Left:} In the simulated data, three groups of participants exhibit varying levels of response in three different brain regions to \emph{Task 1} and \emph{Task 2} stimuli, depending on the locations of underlying participant and stimulus embeddings used to generate the data. \textbf{Right:} NTFA recovers these conditions in participant and stimulus embeddings without prior knowledge. Only the relative spatial arrangement is of interest. Since the original embeddings vary relative to each other only along the horizontal axis, NTFA learns a distribution for these embeddings with very high variance along the vertical axis. }
    \label{fig:synthetic_embeddings}
    \vspace{-1em}
\end{figure*}

\textbf{Synthetic Data}: For synthetic data, NTFA recovers stimulus and participant embeddings that are qualitatively similar to the embeddings that we used to generate the data (Figure~\ref{fig:synthetic_embeddings}). We emphasize that embeddings are learned directly from the synthetic data in an entirely unsupervised manner, which means that there is in principle no reason to expect embeddings to be exactly the same. However, we do observe that learned embeddings for participants and stimuli are well-separated, appear to have some variance, and are invariant under linear transformations. Moreover, given the ``true'' number of factors ($K=3$), NTFA reconstructs synthetic data better than HTFA.

\begin{figure*}[!t]
    \begin{center}
        \includegraphics[width=0.33\textwidth]{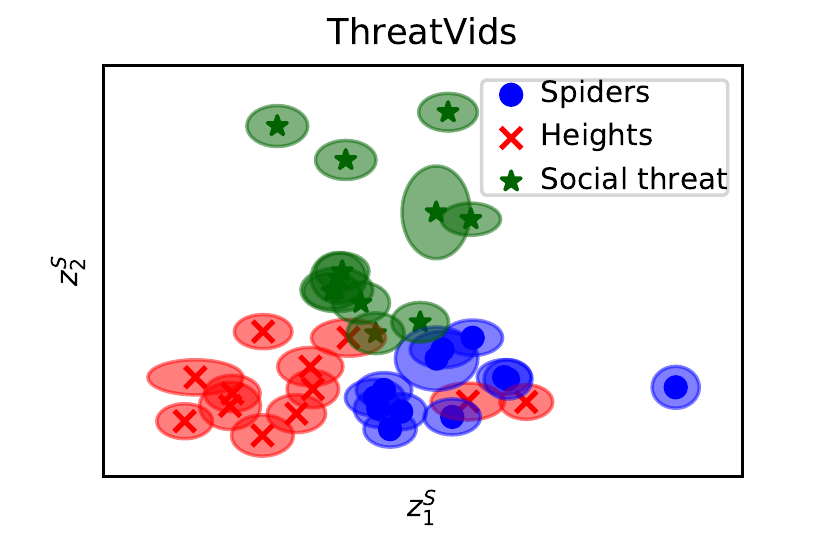}
        \includegraphics[width=0.33\textwidth]{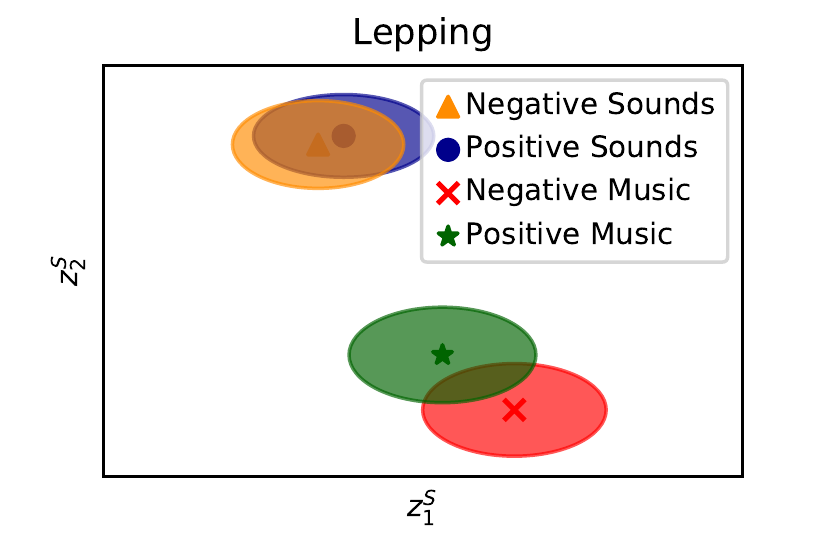}
        \includegraphics[width=0.32\textwidth]{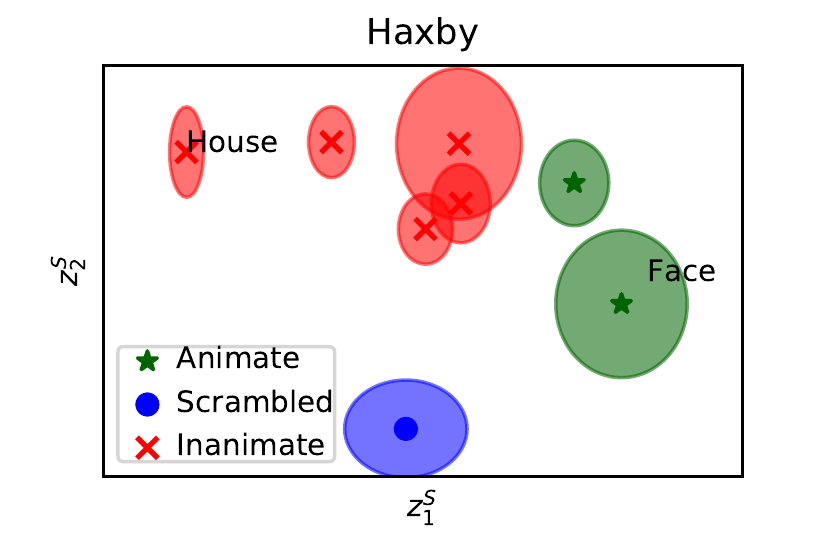}
    \end{center}
    \vspace{-1em}
    \caption{\textbf{Inferred distributions of stimulus embeddings.} \textbf{Left}: On ThreatVids, stimulus embeddings recovered groups of fear stimuli.
    \textbf{Middle}: On the Lepping dataset, stimulus embeddings clearly distinguish the musical stimuli from the non-musical sounds; positive vs negative music show less overlap than positive vs negative sounds. \textbf{Right}: On the Haxby dataset, stimulus embeddings for animate objects are separated from various inanimate objects. Faces were among the animate embeddings and clearly distinct from the embedding for houses, as expected \citep{Haxby2425}.}
    \label{fig:stimulus-embeddings}
    \vspace{-0.5em}
\end{figure*}

\textbf{ThreadVids Dataset}: In this dataset, metadata provided individual stimulus labels as well as stimulus categories to group them together. Table~\ref{table:reconstruction_errors} shows that NTFA generalizes better than HTFA to held-out participant-stimulus pairs, without additional inference.  In an analysis without resting-state data, NTFA uncovers stimulus embeddings that clearly correlate with stimulus categories. While ``Heights'' and ``Spiders'' show some overlap, ``Social Threat'' is clearly separated (Figure~\ref{fig:stimulus-embeddings}, left column).

\textbf{Lepping Dataset}: We here infer one embedding per stimulus category, since the metadata does not label individual stimuli. The embeddings (Figure~\ref{fig:stimulus-embeddings}, middle column) display a clear separation between music and nonmusical sounds, with positive and negative music showing a greater probability of differing from one-another.  Positive and negative sounds overlap in the embedding space. This is consistent with previous findings \citep{10.1371/journal.pone.0156859}.

\textbf{Haxby Dataset}: We here again infer category embeddings. Inference resulted in stimulus embeddings spread throughout the embedding space, with animate and inanimate stimuli segregated (Figure~\ref{fig:stimulus-embeddings}, right column).  This reflects the evidence for distinct processing of animate and inanimate objects in scenes \citep{NASELARIS2012239,doi:10.1002/hbm.24212}.

\vspace{-1em}
\subsection{Point-estimate Embeddings from the SRM, the MN-SRM and PCA}
\vspace{-0.5em}

Evaluating the results above poses inherent challenges in the sense that we lack ground truth. Moreover, NTFA is, to our knowledge the first method that infers low-dimensional embeddings for participants and stimuli directly from the data. To provide some point of comparison, we devise two ad-hoc baselines that compute point estimates of embeddings directly from the input data. 

The first baseline applies PCA, to see how a simple model might still capture meaningful structure in embeddings. We vectorized each trial $Y_n$ to obtain $N$ vectors of $TV$ dimensions. We then time-averaged these vectors, performed PCA upon them, and retained the first two principal components. This linear projection of the data did not capture any meaningful structure, as shown in Figure~\ref{fig:stimulus-embeddings-pca} in Appendix~\ref{app:pca_srm_embeddings}. 

The second baseline computes post-hoc embeddings from the SRM. The SRM learns a shared response matrix $S \in \mathbb{R}^{T \times K}$ and a participant-specific orthonormal weight matrix $W_p \in \mathbb{R}^{K \times V}$ to approximate the signal as $S \cdot W_p$. Since our datasets comprise unaligned stimuli, we reorder blocks in each scanning run to align stimuli across participants. We then compute participant embeddings by vectorizing the $W_p$ and projecting to the first two principal components with PCA. We split the shared-response matrix $S$ into stimulus blocks $S_s$ and then project to the first two principal components. We show results for this analysis in Figure~\ref{fig:stimulus-embeddings-srm} and Figure~\ref{fig:synthetic_embeddings_si} in Appendix~\ref{app:pca_srm_embeddings}.

The third baseline computes post-hoc embeddings from the Matrix Normal SRM in exactly the same fashion as done for SRM. The MN-SRM  is similar to the SRM, though it assumes a weaker Gaussian prior for $W_p$ with a shared spatial covariance across subjects. It also assumes that all subjects share the same temporal noise covariance in addition to the shared response $S$. We show results for this analysis in Figure~\ref{fig:stimulus-embeddings-mn-srm} in Appendix~\ref{app:pca_srm_embeddings}.

The SRM- and MN-SRM-derived point estimates are qualitatively similar to those obtained with NTFA, but do not provide any notion of uncertainty. This makes them difficult to interpret, particularly in cases with few stimulus categories such as the Lepping and Haxby datasets.

\vspace{-0.5em}
\subsection{Multivoxel Pattern and Functional Connectivity Analysis}
\label{subsec:downstream-applications}
\vspace{-0.5em}

One of the advantages of learning a deep generative model is that we can use the learned latent representations in downstream tasks. To illustrate this use case, we consider two types of post-scan analyses that are commonly performed on full fMRI data. As features in these analyses we use the low-dimensional representation learned by NTFA: the inferred factor locations, widths, and weights.

\begin{figure*}[!t]
    \begin{center}
    \includegraphics[width=0.324\textwidth]{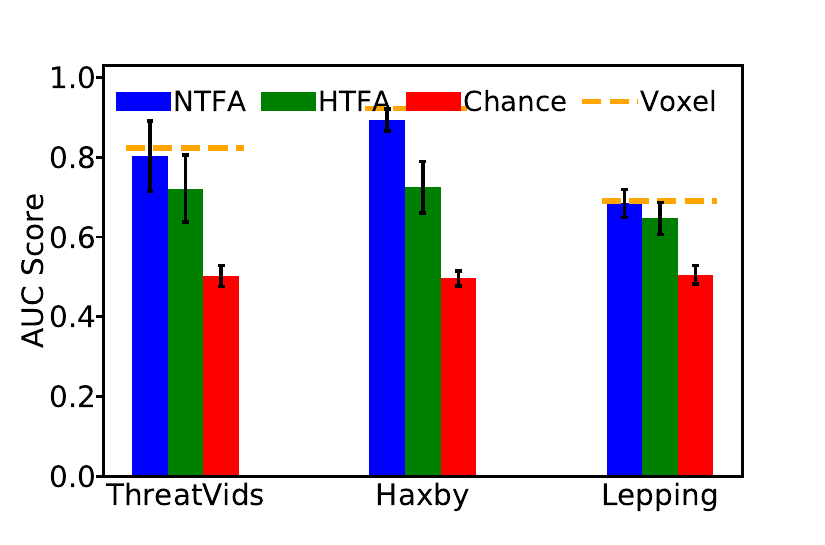} 
    \includegraphics[width=0.324\textwidth]{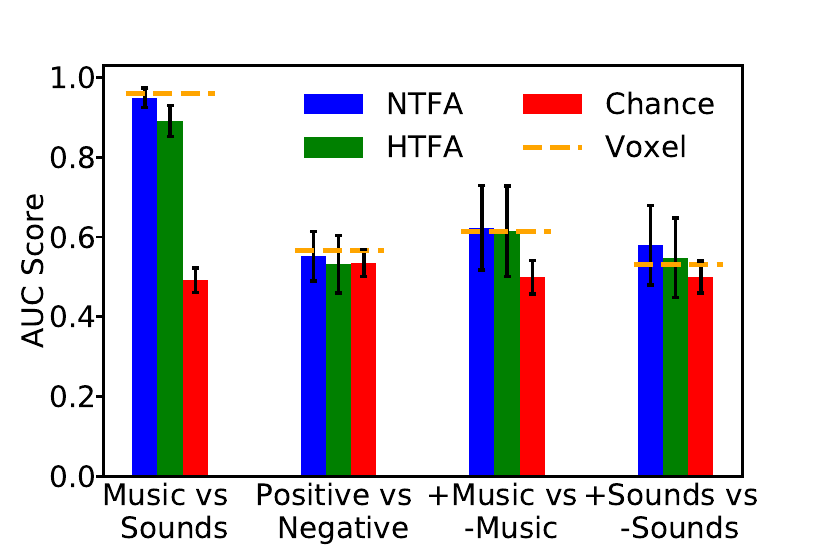}
    \includegraphics[width=0.324\textwidth]{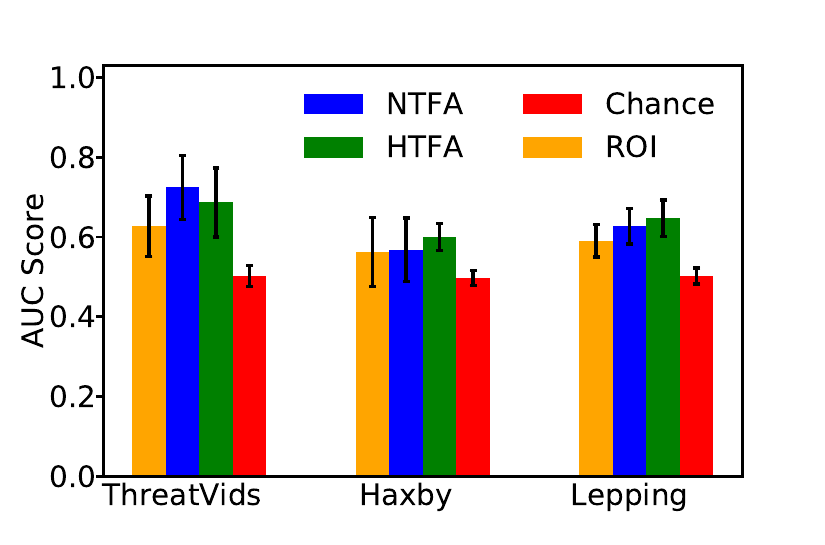}\\
    \end{center}
    \vspace{-1em}
    \caption{\textbf{Classification performance} measured by Accuracy Under the Curve (AUC).  We show mean AUC scores with $95\%$ CI across categories. \textbf{Left:} For each dataset, we compare supervised voxel selection (``Voxel''), NTFA and HTFA. \textbf{Middle:} AUC scores for Lepping dataset across different stimulus categories. The embeddings in Figure~\ref{fig:stimulus-embeddings} (middle) qualitatively match these results.
    \textbf{Right:} Classification using pairwise time-correlation matrices. Functional connectomes derived from NTFA and HTFA's representations outperform those from the data-agnostic regions of interest (ROIs).
    }
    \label{fig:svm_accuracy}
    \vspace{-0.5em}
\end{figure*}
\textbf{Multivoxel Pattern Analysis (MVPA)}:
In MVPA, a regularized linear classifier is trained to predict experimental variables from distributed patterns of mean voxel intensities.  This is usually preceded by a supervised feature selection step to select voxels most relevant to the classification task \citep{Pereira2009}. We apply this standard method to our datasets and compare it to using time-averaged weight matrices derived without supervision from NTFA and HTFA.  We show the resulting classification accuracy scores, measured using Area Under the (receiver operating) Curve (AUC) on the left in Figure~\ref{fig:svm_accuracy}.  While all three methods perform significantly better than chance, NTFA outperforms HTFA, and performs almost as well as supervised voxel selection. We also note that the stimulus embeddings qualitatively predict classification performance on different stimulus categories, as seen in the middle of Figure ~\ref{fig:svm_accuracy} for Lepping dataset. NTFA learns a latent representation useful for MVPA stimulus classification, without supervision. We detail the methods and results in Appendix~\ref{app:mvpa}.

\textbf{Functional Connectivity (FC)}: Functional connectivity analyses study the co-activation of brain areas during resting-state or during a task, regardless of their apparent physical distance.  A variety of studies have shown FC, and changes in FC, to correlate with behavior \citep{Elliott2019}.  Voxels, however, capture neither single neurons, nor functional brain regions that could hypothetically share an activation pattern. NTFA's latent factor representations provide a data-driven alternative to standard regions of interest (ROIs) that maintains the spatial locality crucial to functional connectivity. In Figure~\ref{fig:svm_accuracy} (right), we see that linear classifiers trained on NTFA's latent factor representations perform better at a stimulus classification task than those trained on ROIs.  NTFA-derived FC patterns perform comparably to HTFA-derived patterns, despite NTFA's lower parameter count.

\vspace{-1em}
\section{Conclusion}
\label{sec:conclusion}
We have introduced Neural Topographic Factor Analysis, an unsupervised model for fMRI data that characterizes individual variation in the neural response by inferring low-dimensional embeddings for participants and stimuli.  NTFA is a first step in a line of approaches that employ deep generative models to incorporate inductive biases into unsupervised analyses of neuroimaging experiments. By designing models whose structure reflects a particular experimental design, or potentially even a neuroscientific hypothesis, we can hope to appropriately account for the uncertainties that arise from limitations in statistical power and sample sizes.  This provides a path towards analyses that reason about individual variation  in a manner that is data-efficient and mitigates risks of overfitting the data.

\section*{Broader Impact}

While this paper reports on NTFA in terms of its characteristics as a general-purpose machine learning method for the analysis of neuroimaging data, we envision downstream impacts in the context of specific neuroscientific research questions. There is a need in neuroscience research to develop formal computational approaches that capture individual differences in neural function.

The embedding space yields a simple, visualizable model to inspect individual differences that has the potential to, at least in a qualitative manner, provide insights into fundamental questions in cognitive neuroscience. One such question is whether neural responses to stimuli are shared across individuals, vary by pre-defined participants groups (e.g.~depressed vs.~non-depressed participants), or are unique to participants or subgroups (e.g.~as suggested by calls for “precision medicine” approaches). 

Going forward, we will use our pilot data to address whether the neural basis of fear, for example, is shared across individuals and situations (i.e.~there is a single ``biomarker'' or ``neural signature'' for fear), or as we expect, whether it varies by person or situation (suggesting that biomarkers for fear are idiographic) \citep{Satpute2019TICS}. With further developments, we plan to perform more extensive neuroimaging experiments that probe individual variation in additional fMRI datasets including in house datasets and publicly available datasets. Our hope is that the work presented in this paper will form a basis for developing probabilistic factor-analysis models with structured priors that will allow testing and development of specific neuroscientific hypotheses regarding individual variation in the functional neural organization of psychological processes.

\begin{ack}
The authors thank the anonymous reviewers for their constructive feedback.  We also thank Jeremy Manning for insightful conversations, and Michael Shvartsman for sharing his Matrix-Normal SRM code with us.
This work was supported by startup funds from Northeastern University and the University of Oregon, as well as the Intel Corporation, the National Science Foundation (NCS 1835309), and the US Army Research Institute for the Behavioral and Social Sciences (ARI W911N-16-1-0191).
\end{ack}

\bibliography{neurips_2020}
\bibliographystyle{plainnat}

\clearpage\newpage
\appendix
\section{Appendix}

\begin{table}[!h]
\caption{\textbf{Description of Notations.} This table explains notations used in the paper, in the order they appear in the main text.}
\centering
\begin{tabular}{l|p{100mm}}
\toprule
\textbf{Symbol} &
\textbf{Description}\\
\midrule
$T$ 
& Number of TRs (in a block).
\\
$V$ 
&  Number of voxels in a brain image.
\\
$K$ 
&  Number of factors used to approximate the input data using factor analysis (usually $K << V$).
\\
$Y_n \in \mathbb{R}^{T\text{x}V}$
& $n^{\text{th}}$ block of the dataset under analysis, organized as a number of TRs times number of voxels matrix.
\\
$W_n \in \mathbb{R}^{T\text{x}K}$
& A lower-rank matrix of weights that specifies the time varying weights of each factor.
\\
$F_n \in \mathbb{R}^{K\text{x}V}$
& A lower-rank matrix of factors, such that an element $f_{kv}$ at row $k$ and column $v$ specifies the contribution of $k^{\text{th}}$ factor to the activation of voxel $v$. 
\\
$\sigma^Y$
& Gaussian noise variance, assumed constant across a given dataset.
\\
$\mu^\scw_{n,k}, \sigma^\scw_{n,k}$
& Mean and variance for $k^{\text{th}}$ row of $W_n$. 
\\
$\mu^\scw, \sigma^\scw$
& Hyperparameters for $\mu^\scw_{n,k}, \sigma^\scw_{n,k}$.
\\
$x^\scg_v \in \mathbb{R}^3$
& Coordinates for voxel $v$.
\\
$x_{n,k}^\scf \in \mathbb{R}^3, \rho^\scf_{n,k}$
& center and log-width of the $k^{\text{th}}$ factor for block $n$.
\\
$\kappa(x^\scg_v, x^\scf_{n,k} \,;\, \rho^\scf_{n,k})$
& Radial basis function with center at $x^\scf_{n,k}$ and log-width $\rho^\scf_{n,k}$ evaluated at location $x^\scg_v$. 
\\
$x^\scf, \rho^\scf$
& Hyperparameters for factor centers and factor log-widths.
\\
$P$
& Total number of participants in a given dataset.
\\
$S$
& Total number of unique stimuli in a given dataset.
\\
$p_n \in \{1, \ldots, P\}$
& Participant involved in block $n$.
\\
$s_n \in \{1, \ldots, S\}$
& Stimulus involved in block $n$.
\\
$z^\scp_p \in \mathbb{R}^D$
& $D-$dimensional participant embedding associated with a specific participant $p$.
\\

$z^\scs_s \in \mathbb{R}^D$
& $D-$dimensional stimulus embedding associated with a specific stimulus $s$.
\\
$x^{\scf}_p \in \mathbb{R}^{K \times 3}, \rho^{\scf}_{p} \in \mathbb{R}^{K}$
& Factor centers and log-widths for a specific participant $p$.
\\
$\mu^{x}_{p} \in \mathbb{R}^{K \times 3}, \sigma^{x}_{p}\in \mathbb{R}^{K}$
& Means and variances for the $K$ factor centers for participant $p$.
\\
$\mu^{\rho}_{p} \in \mathbb{R}^{K \times 3}, \sigma^{\rho}_{p}\in \mathbb{R}^{K}$
& Means and variances for the $K$ factor log-widths for participant $p$.
\\
$\eta^{\scf}_\theta$
& Neural network that takes $z^\scp_p$ as input and outputs $\mu^{x}_{p}, \sigma^{x}_{p}, \mu^{\rho}_{p}, \sigma^{\rho}_{p}$ for participant $p$.
\\
$\mu^{\textsc{w}}_{n} \in \mathbb{R}^K, \in \mathbb{R}^K$
& Mean and variance for each of the $K$ rows of the matrix of weights $W_n$ for block $n$. 
\\
$\eta^\scw_\theta$
& Neural network that takes a concatenation of $z^\scp_p$ and $z^\scs_s$ for participant $p$ and stimulus $s$ in block $n$ and outputs $\mu^{\textsc{w}}_{n}$ and $\sigma^{\textsc{w}}_{n}$.
\\
$\theta$
& Learnable parameters in the generative model, that is, the neural network weights.
\\
$\lambda$
& Learnable parameters of the variational distributions.
\\
$\lambda^\scw_{n,t}$
& Parameters of posterior distribution over weights $W_{n,t}$.
\\
$\lambda^\scs_{s}, \lambda^\scp_{p}$
& Parameters of posterior distribution over stimulus embedding $z^\scs_s$ for stimulus $s$ and participant embedding $z^\scp_p$ for participant $p$.
\\
$\lambda^{\textsc{x}^\scf_{p}}, \lambda^{\rho^\scf_{p}}$
& Parameters of posterior distribution over 
factor centers $x^\scf_p$ and factor log-widths $\rho^\scf_p$ for participant $p$.
\\
$\tilde{Y}$
& Held-out validation data.
\\

\bottomrule
\end{tabular}
\label{app:tb_not}
\end{table}

\subsection{Participant embedding results}
\label{app:participant_embeddings}
\begin{figure}
    \begin{center}
    \includegraphics[width=0.6\columnwidth]{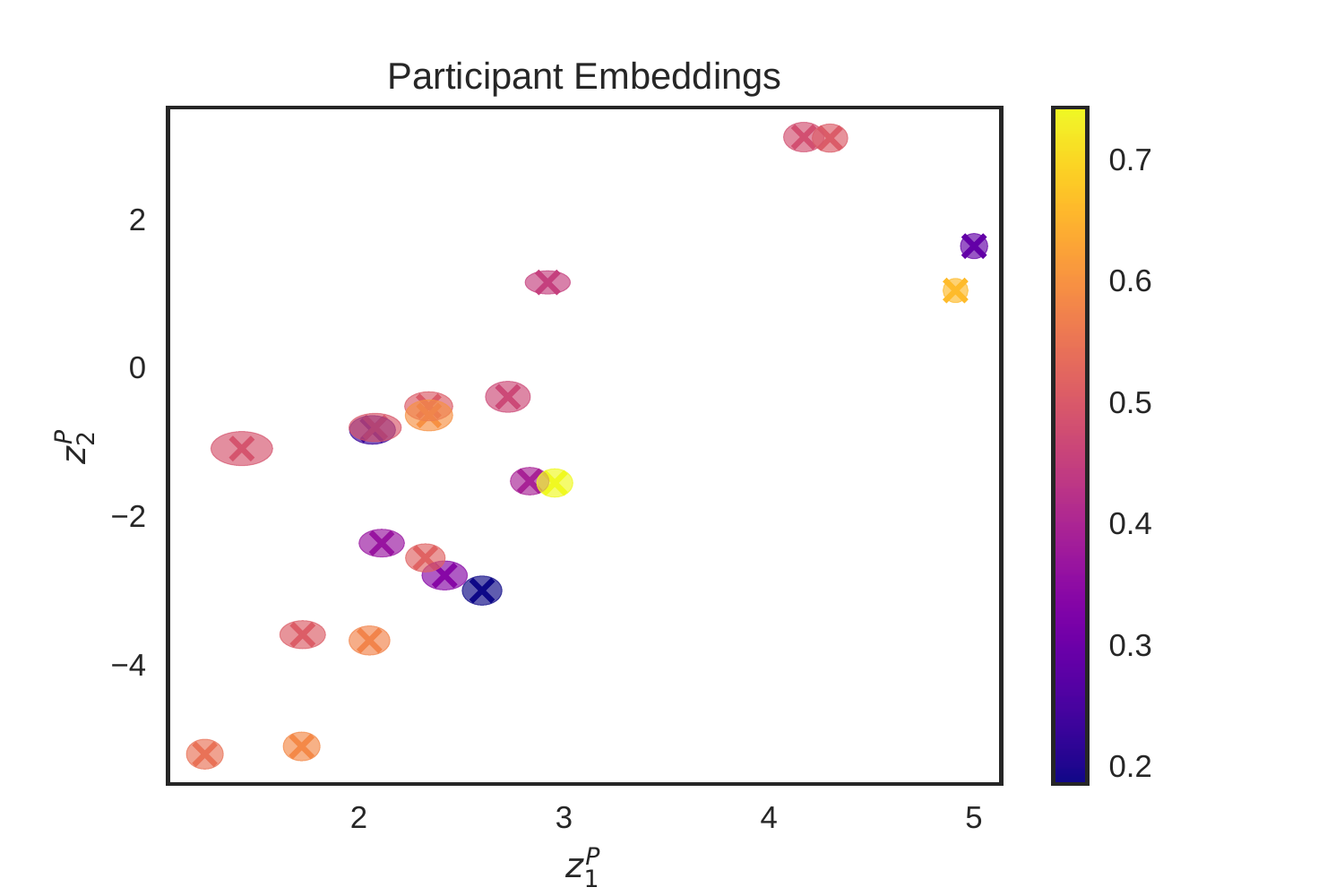}
    \end{center}
    \caption{Participant embeddings from the ThreatVids dataset. Crosses indicate the location of the (approximate) posterior mean, and ellipses display (approximate) posterior covariance. The labels are used only for visualization purposes.  Participant embeddings are color-coded by their reported level of fear across all stimulus categories (heights, spiders, and social threats). Cooler colors indicate lower mean fear ratings, while warmer colors indicate higher mean fear ratings.}
    \label{fig:affvids-participant-embeddings}
\end{figure}

In the ThreatVids dataset, the participant embeddings uncovered three groups: the more frightened, the less frightened, and those sensitive to particular fears (Figure~\ref{fig:affvids-participant-embeddings}). Participant embeddings for individual fear categories are shown in Figure~\ref{fig:affvids-participant-embeddings}. Participants were not recruited in specific groups (e.g.~arachnophobes and acrophobes), and stimuli could be categorized multiple ways (e.g.~by kind or degree).  We observe that most participants carried a greater fear of heights (left) and social threat (right) than of spiders (middle).  A scattering of individuals in the mid-left of the embedding space appeared to suffer little overall fear in any stimulus category, while those further out from the centroid had more varied fear experiences across categories.  Few individuals showed high mean fear ratings across stimulus categories.

The participant embeddings do not seem to predict the self-reported fear ratings. However as shown in Figure ~\ref{fig:affvids-participant-embeddings} they do seem to uncover variations among participants in the latent space. Note, for example the participant groups breaking away from the central ``cluster'' towards top-right and bottom-left. This suggests that there are factors not explained by the self-reported fear ratings that might be driving the individual variation in response among participants.

\begin{figure}
    \begin{center}
    \includegraphics[width=0.6\columnwidth]{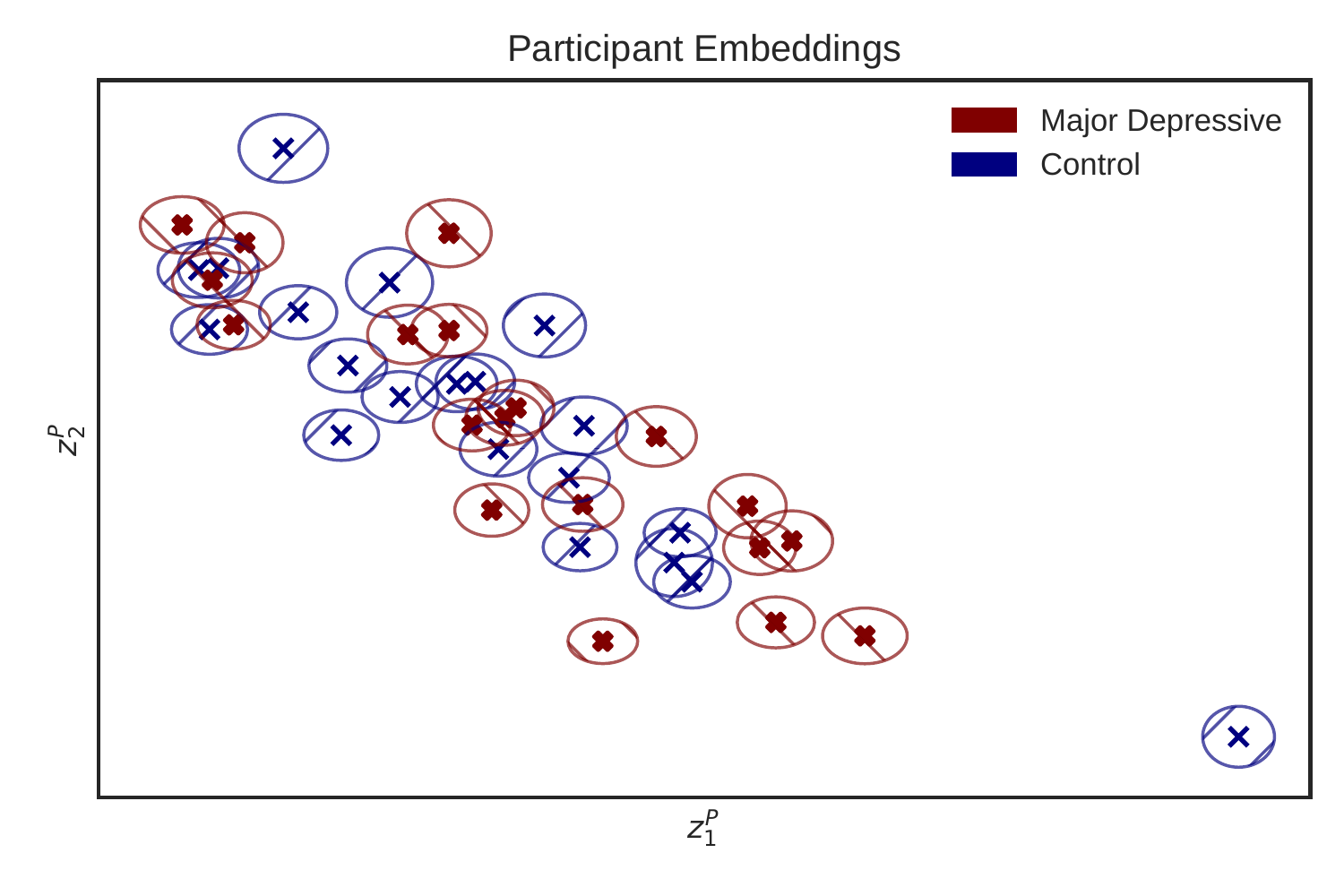}
    \end{center}
    \caption{Participant embeddings for the Lepping dataset \citep{10.1371/journal.pone.0156859}.  Crosses indicate the location of the posterior mean, and ellipses display posterior covariance. The labels are used only for visualization purposes. Participant embeddings did not show a clear difference between control and major depressive groups, but did appear scattered around a linear trend in the latent space, which we have yet to interpret.}
    \label{fig:lepping-participant-embeddings}
\end{figure}

\subsection{``Embeddings'' from PCA, SRM and MN-SRM}
\label{app:pca_srm_embeddings}

\begin{figure*}[!t]
    \begin{center}
        \includegraphics[width=0.324\textwidth]{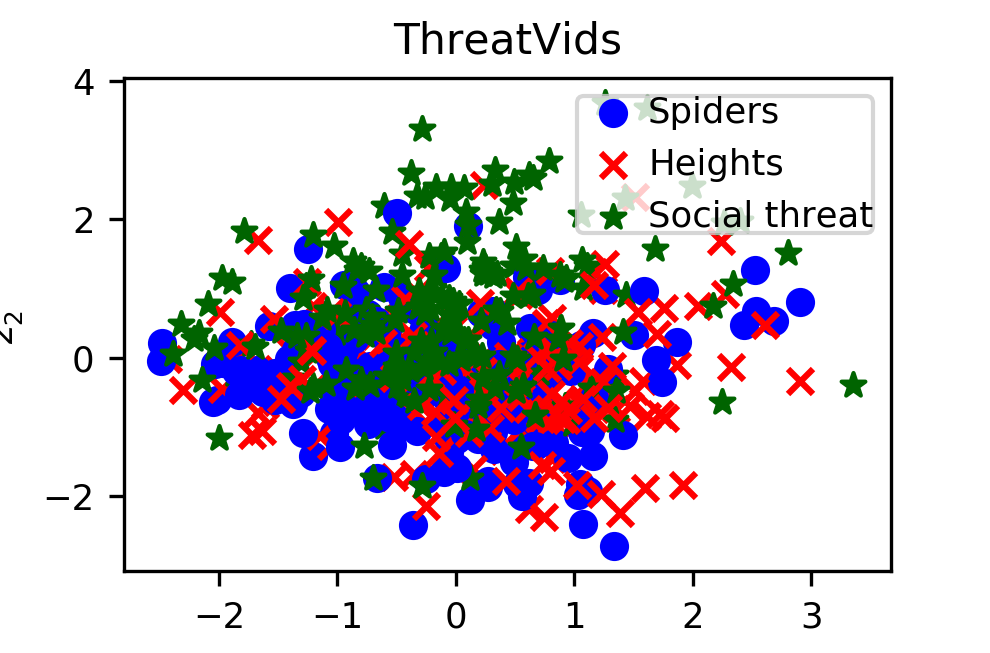}
        \includegraphics[width=0.324\textwidth]{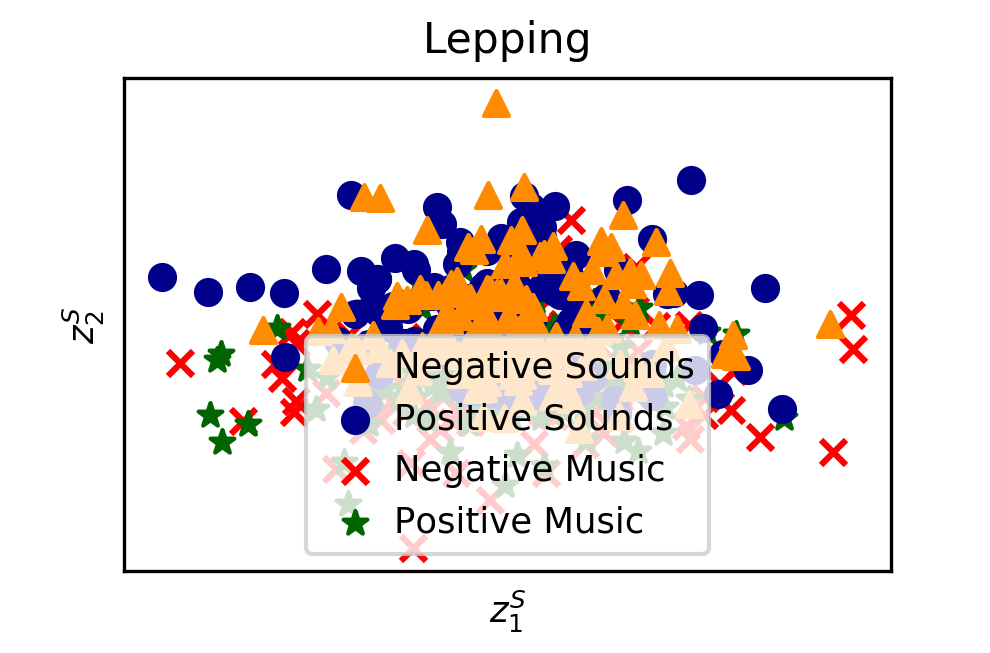}
        \includegraphics[width=0.324\textwidth]{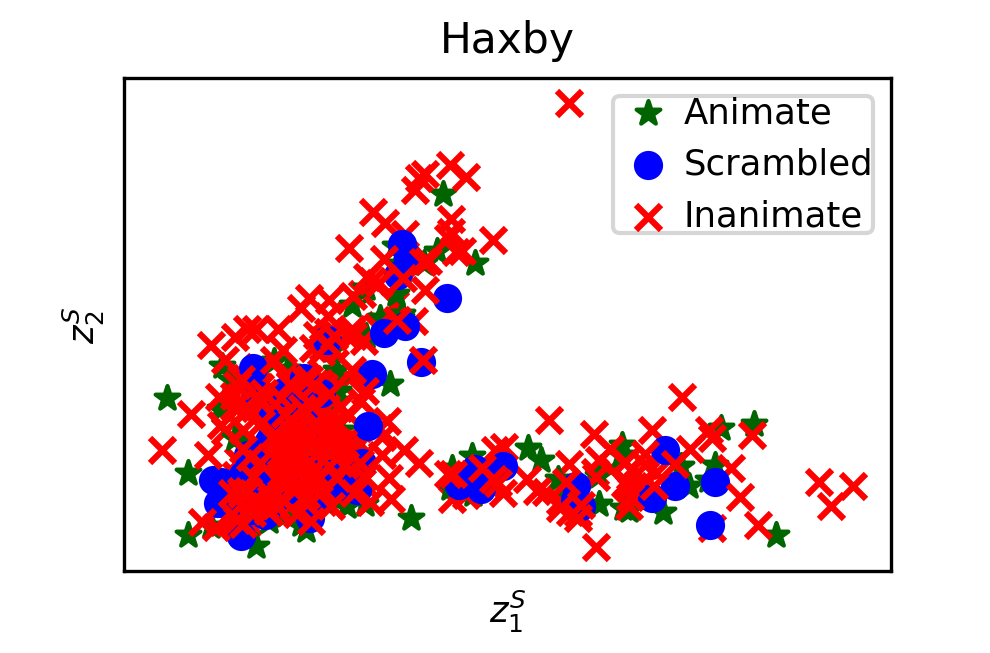}
    \end{center}
    \vspace{-1em}
    \caption{\textbf{PCA projections of input data as embeddings} The baseline embeddings recovered directly from input data by projecting each trial to the first two principal components. The embeddings don't seem to capture any meaningful structure.}
    \label{fig:stimulus-embeddings-pca}
\end{figure*}

\begin{figure}[!t]
    \centering
    \includegraphics[width=0.45\textwidth]{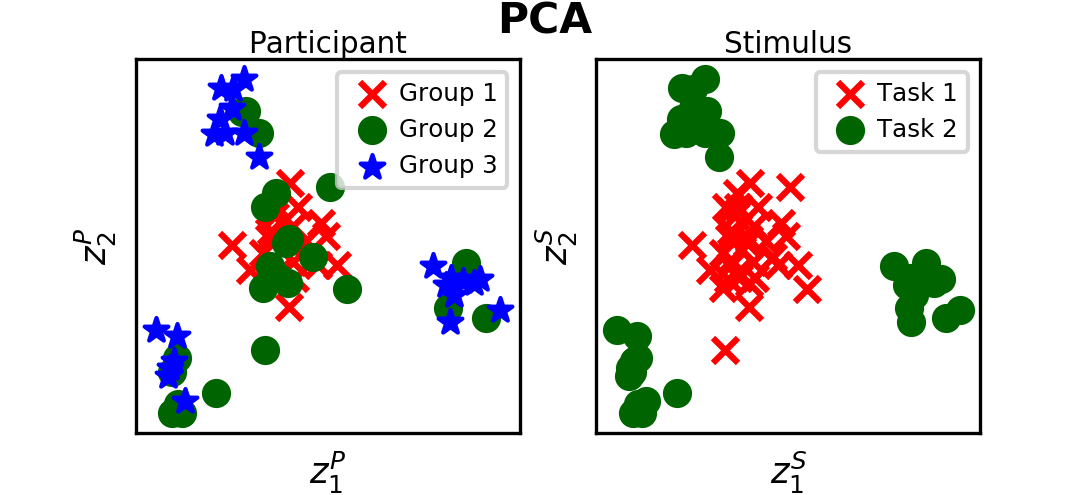}
    ~~
    \includegraphics[width=0.45\textwidth]{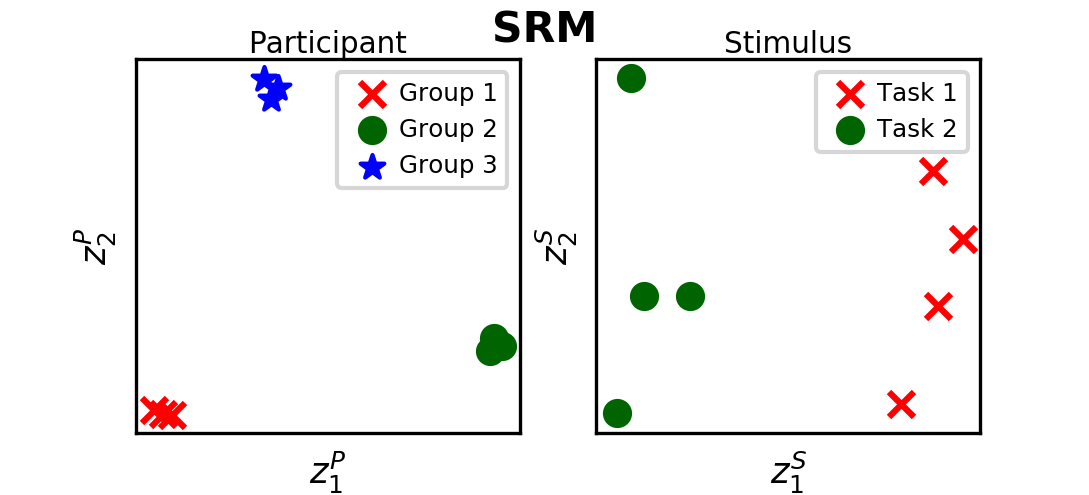}
    
    \includegraphics[width=0.45\textwidth]{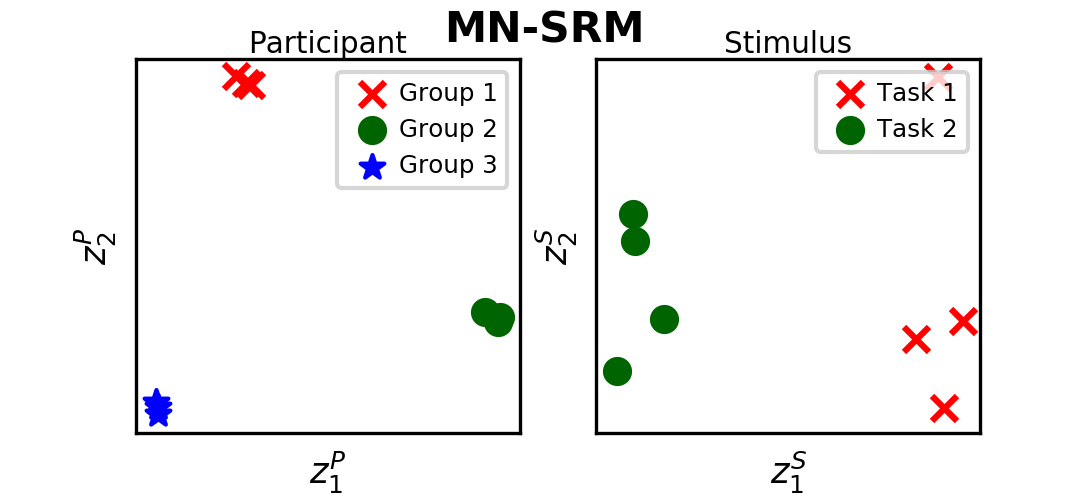}

    \vspace{-1em}
    \caption{\textbf{Embeddings for synthetic data using PCA, SRM and MN-SRM}: On simulated data PCA fails to capture the participant or stimulus groups accurately. Post-hoc embeddings from the SRM and MN-SRM look qualitatively similar to NTFA albeit without uncertainty estimates.}
    \label{fig:synthetic_embeddings_si}
    \vspace{-1em}
\end{figure}

To establish a rough baseline. We performed PCA on the input data. The data $Y_n$ from each trial $n$ for each dataset was vectorized and these $N$ ($T \times V$ dimensional) vectors were projected to the first two principal components. Figure~\ref{fig:synthetic_embeddings_si}(left) shows the result overlaid with labels for each trial. Figure~\ref{fig:stimulus-embeddings-pca} shows the same for the three real datasets with task labels overlaid. PCA, perhaps unsurprisingly fails to capture any meaningful structure. 

\begin{figure*}[!t]
    \begin{center}
        \includegraphics[width=0.324\textwidth]{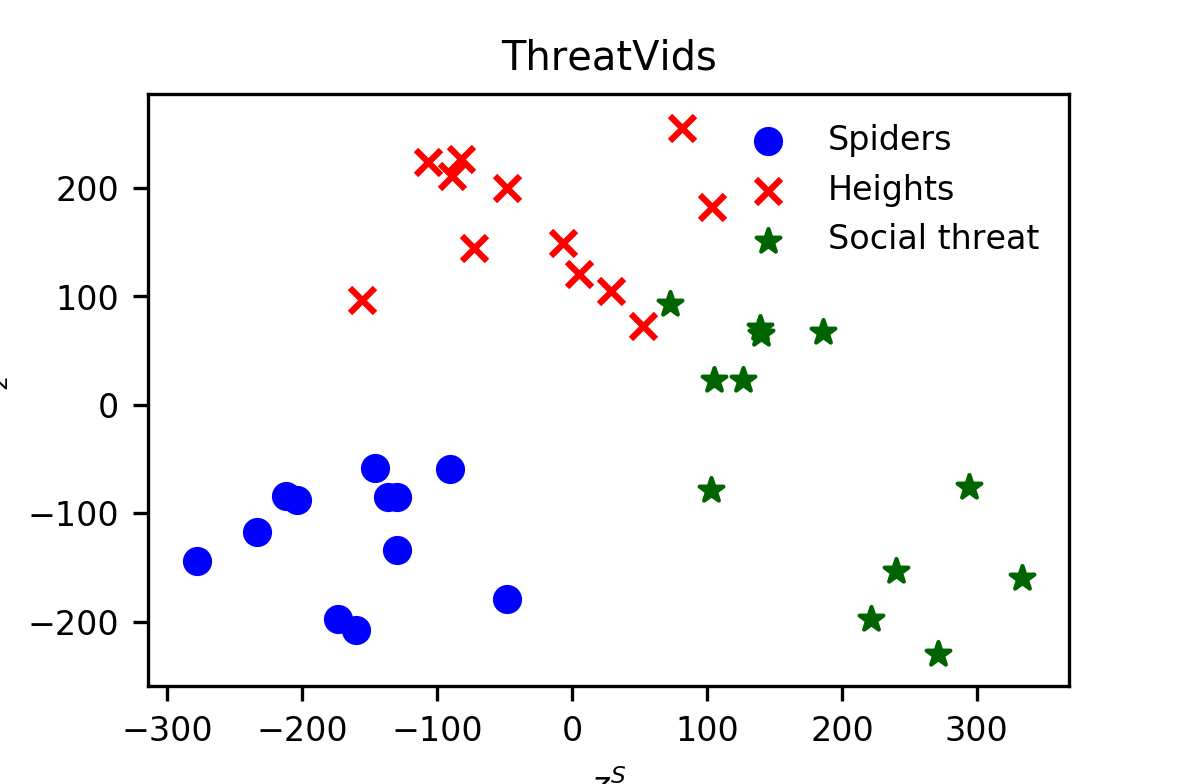}
        \includegraphics[width=0.324\textwidth]{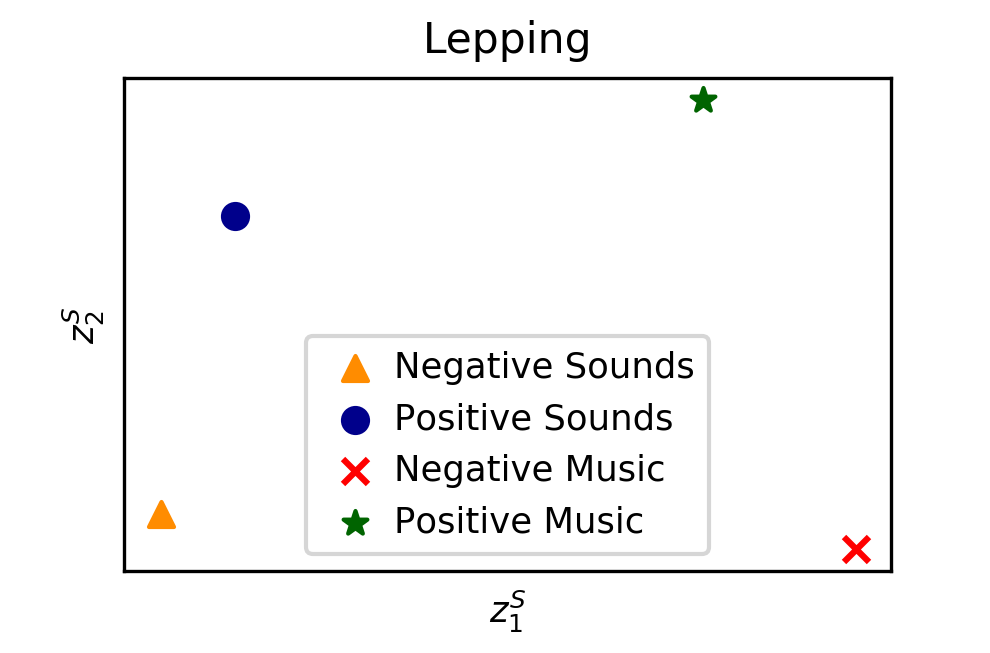}
        \includegraphics[width=0.324\textwidth]{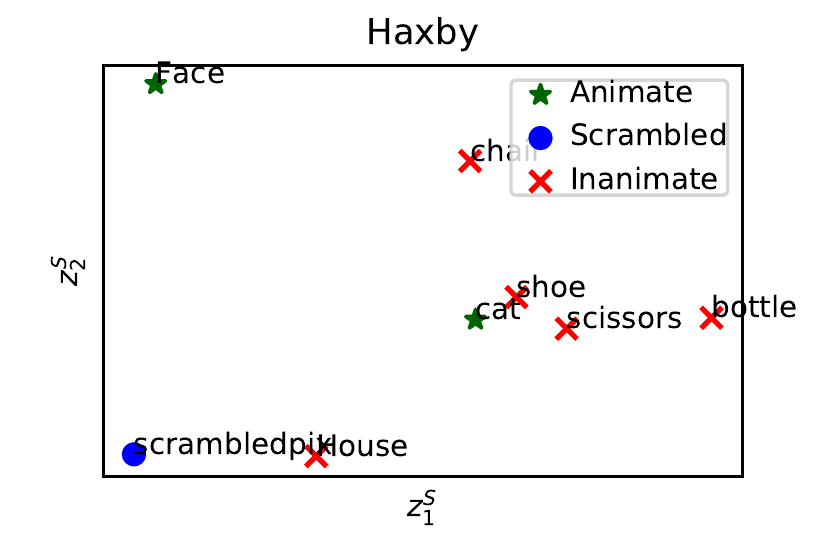}
    \end{center}
    \vspace{-1em}
    \caption{\textbf{Post-hoc embeddings from SRM 
    :} Stimulus embeddings recovered post-hoc from SRM. The embeddings look qualititavely similar to NTFA, but lack uncertainty quantification which makes it difficult to meaningfully reason about the distances in the space.}
    \label{fig:stimulus-embeddings-srm}
\end{figure*}

\begin{figure*}[!t]
    \begin{center}
        \includegraphics[width=0.324\textwidth]{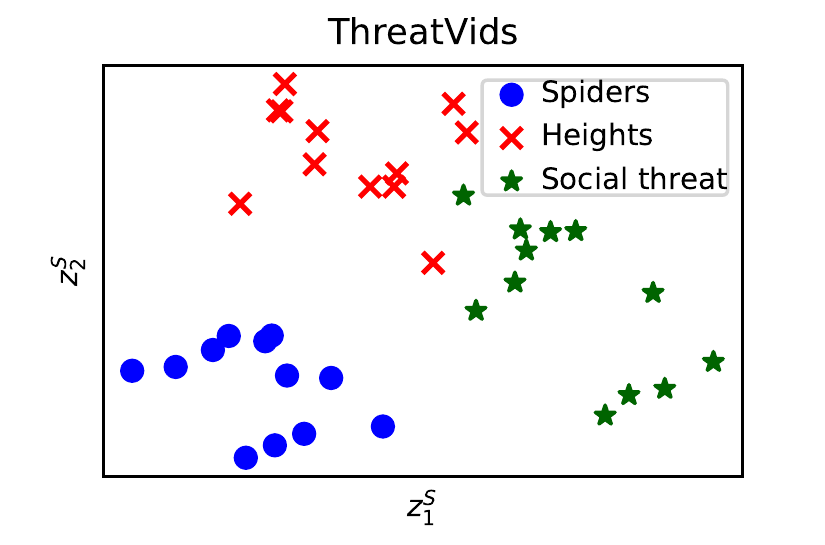}
        \includegraphics[width=0.324\textwidth]{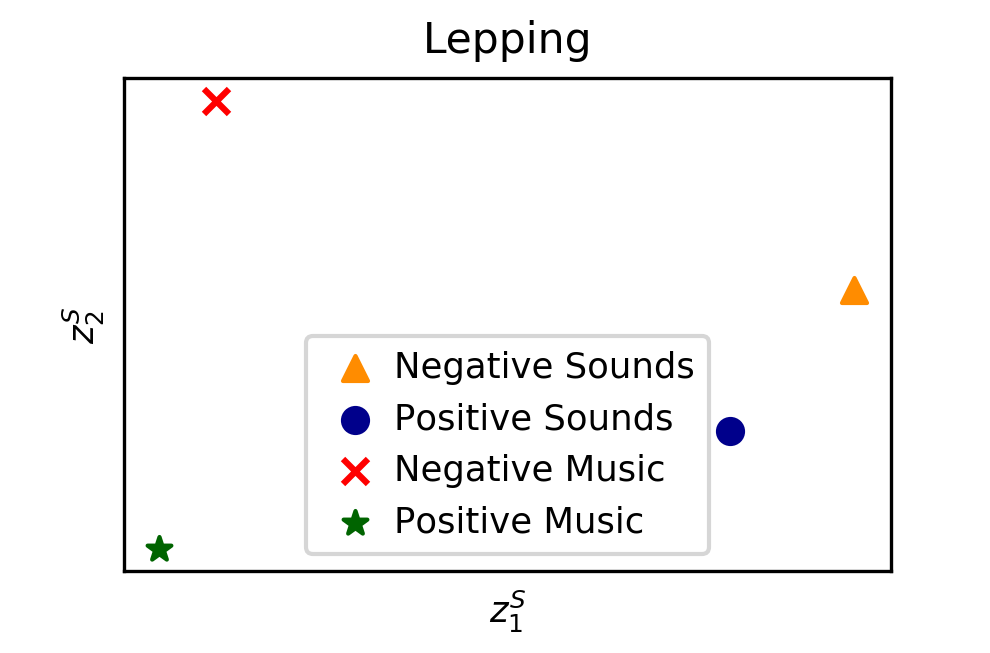}
        \includegraphics[width=0.324\textwidth]{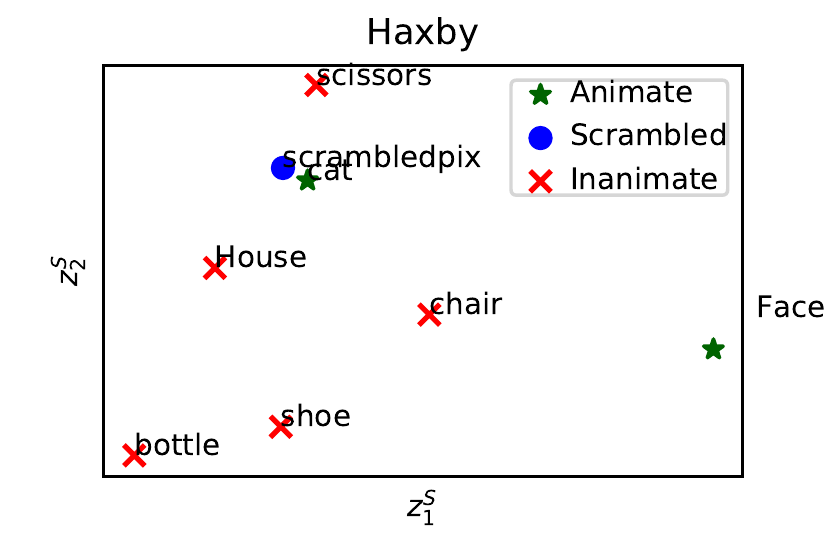}
    \end{center}
    \vspace{-1em}
    \caption{\textbf{Post-hoc embeddings from MN-SRM 
    :} Stimulus embeddings recovered post-hoc from MN-SRM. The embeddings look qualititavely similar to NTFA (and SRM), but lack uncertainty quantification which makes it difficult to meaningfully reason about the distances in the space.}
    \label{fig:stimulus-embeddings-mn-srm}
\end{figure*}

We also acquired a notion of post-hoc ``embeddings'' from SRM and MN-SRM by following these with PCA. SRM and MN-SRM learn a single shared response matrix $S \in \mathbb{R}^{K \times T}$ for all participants in an experiment and are ideally suited to experiments where the stimuli are time aligned across participants. We mimic this structure in our datasets by artificially aligning the trials in the same order of stimuli for all participants. Then the shared response matrix was split into matrices corresponding to each stimulus. These matrices were then vectorized and PCA was done on these as mentioned before. Similarly participant embeddings can be obtained by vectorizing the participant dependent weights learned by SRM and projecting them using PCA. Figure~\ref{fig:synthetic_embeddings_si}(right), and Figure~\ref{fig:synthetic_embeddings_si}(middle) show the results of this procedure for SRM and MN-SRM on synthetic data. Similarly Figure~\ref{fig:stimulus-embeddings-srm} and Figure~\ref{fig:stimulus-embeddings-mn-srm} show the stimulus embeddings for the three real world dataset, using SRM and MN-SRM respectively. While this procedure seems to capture reasonable embeddings for the simulated data and ``ThreatVids''. We notice that a lack of uncertainty around these point estimates means it becomes difficult to interpret them, specially in situations where there's only a limited number of unique stimuli. As is the case for Lepping and Haxby in the middle and right of Figure~\ref{fig:stimulus-embeddings-srm} and Figure~\ref{fig:stimulus-embeddings-mn-srm}.

\subsection{Test-set predictions}

\label{app:additional-predictions}
To visualize the predictive distribution, we compare the time-average of $\tilde{Y}$ to a prediction $\bar{Y} = \bar{\mu}^\scw \cdot \kappa(\bar{\mu}^{x}_p,\bar{\mu}^{\rho}_p)$, where  $\bar{\mu}^\scw, \bar{\mu}^\scw$, $\bar{x}^\scf_p$, $\bar{\rho}^\scf_p$ are computed from the expected values $\bar{z}^\scp$ and $\bar{z}^\scs$ of the embeddings in the variational distribution.

\begin{figure*}[!t]
    \begin{center}
    \begin{tabular}{ccc}
        \textsf{\small Held-out data} & \textsf{\small Posterior predictive (NeuralTFA)} & \textsf{\small Posterior predictive (HTFA)} \\
        \includegraphics[width=0.28\textwidth]{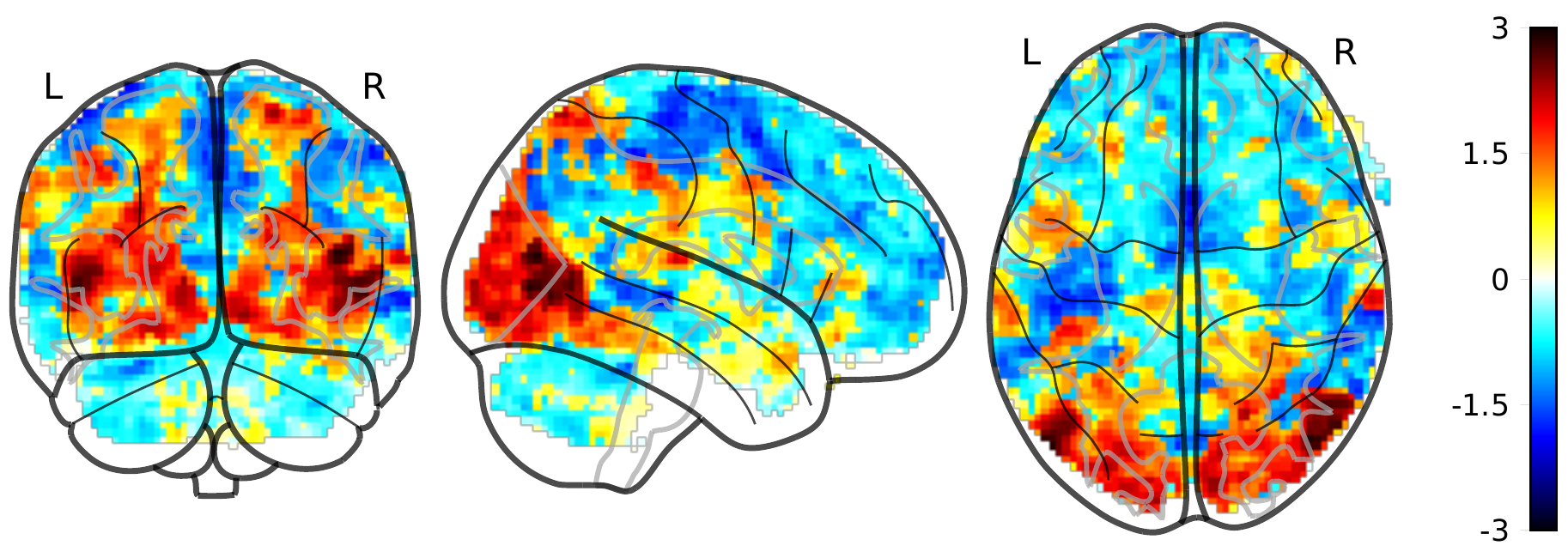} &
        \includegraphics[width=0.28\textwidth]{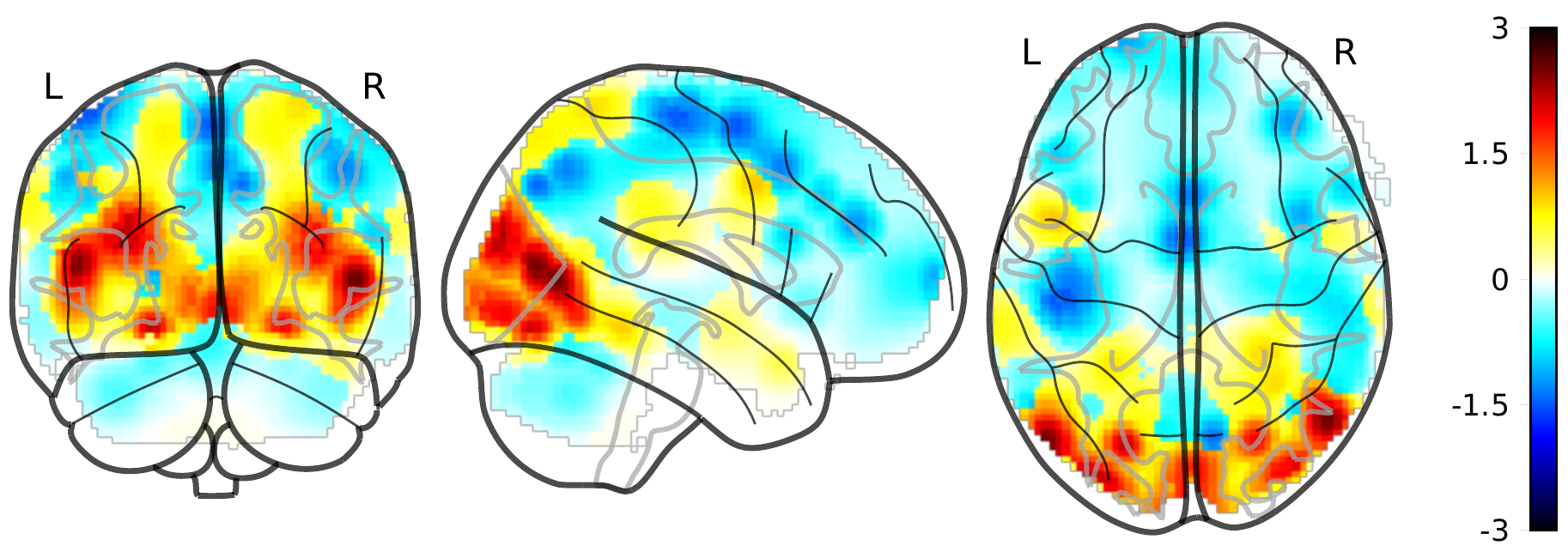} &
        \includegraphics[width=0.28\textwidth]{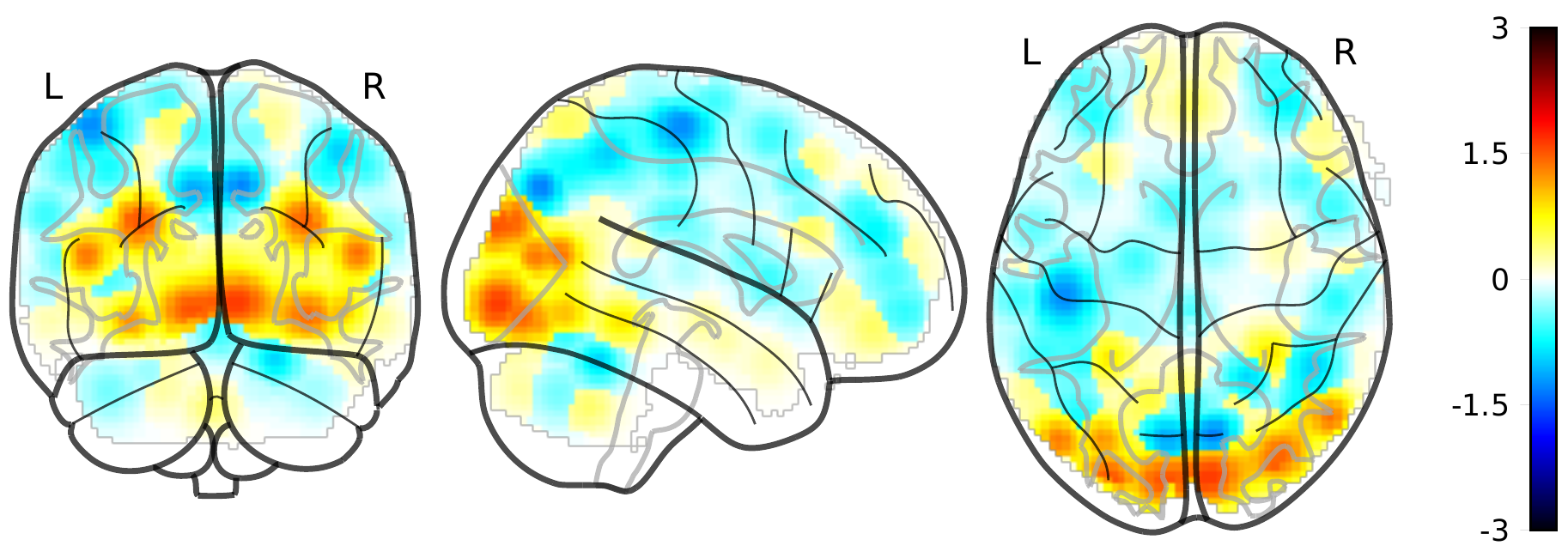} \\
        \includegraphics[width=0.28\textwidth]{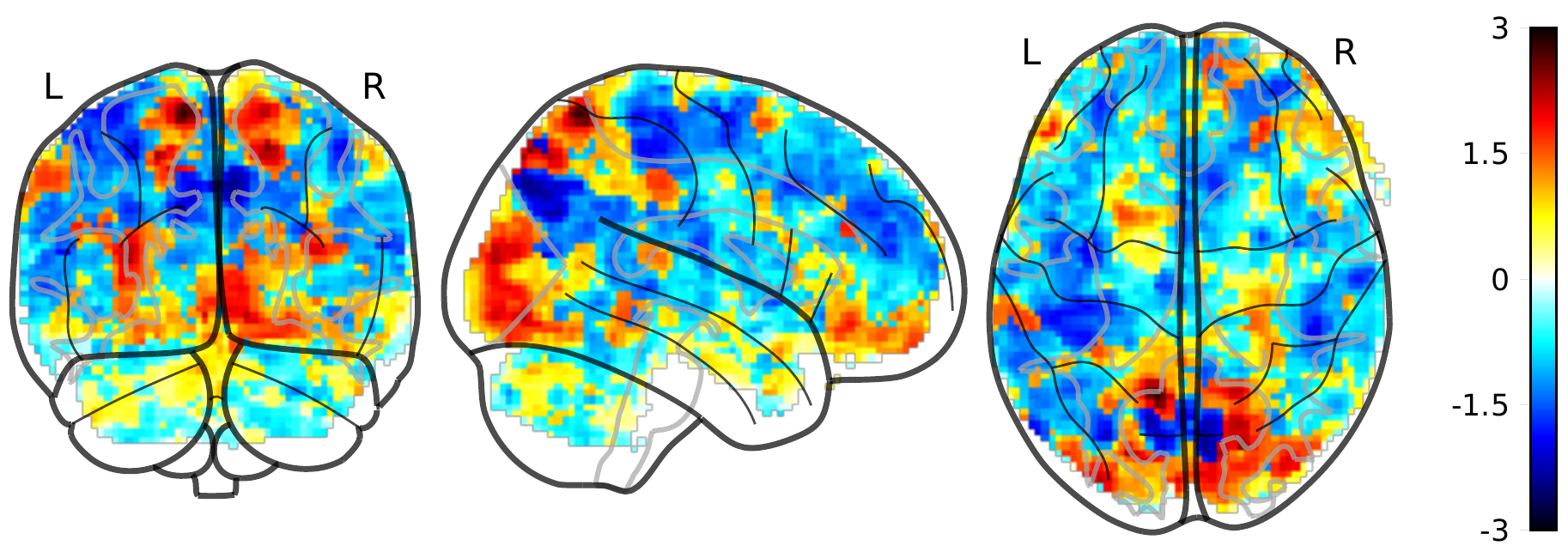} &
        \includegraphics[width=0.28\textwidth]{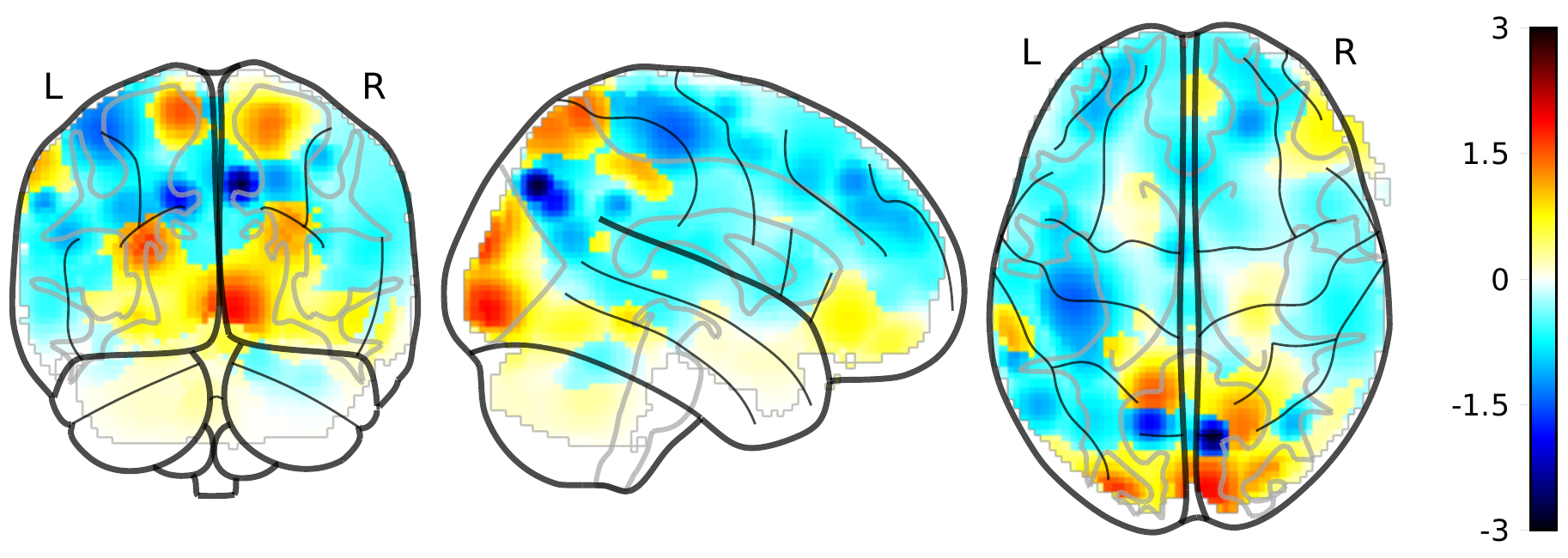} &
        \includegraphics[width=0.28\textwidth]{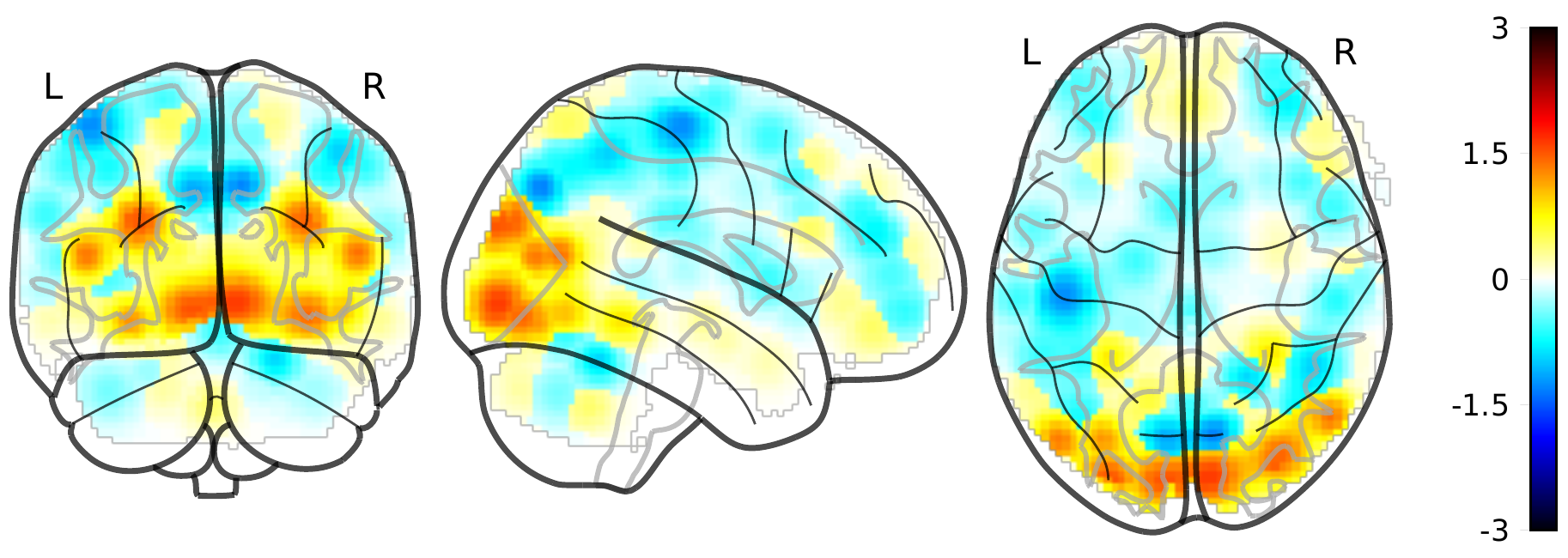} \\
        \includegraphics[width=0.28\textwidth]{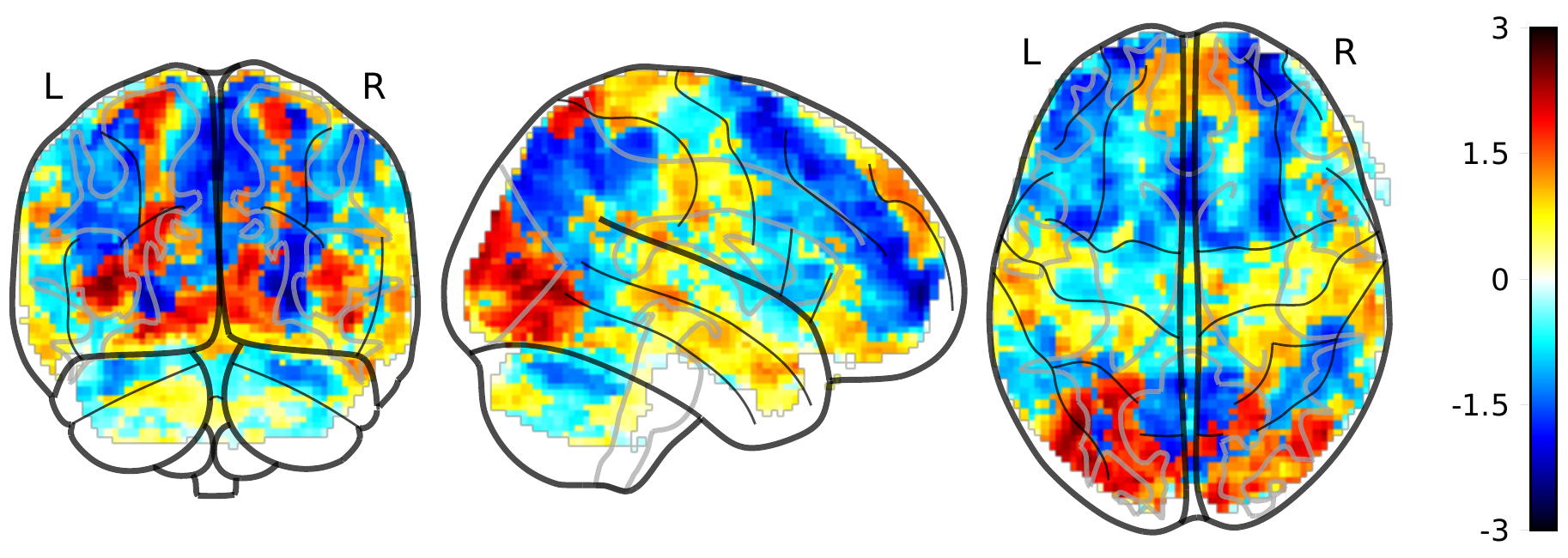} &
        \includegraphics[width=0.28\textwidth]{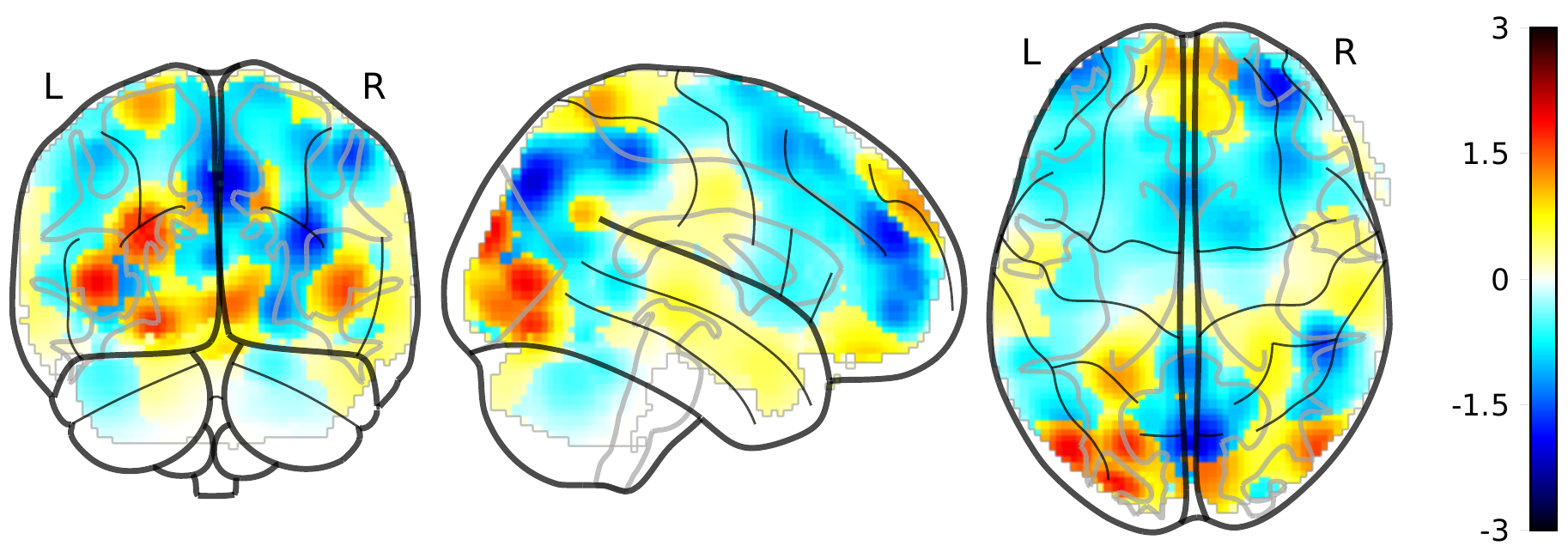} &
        \includegraphics[width=0.28\textwidth]{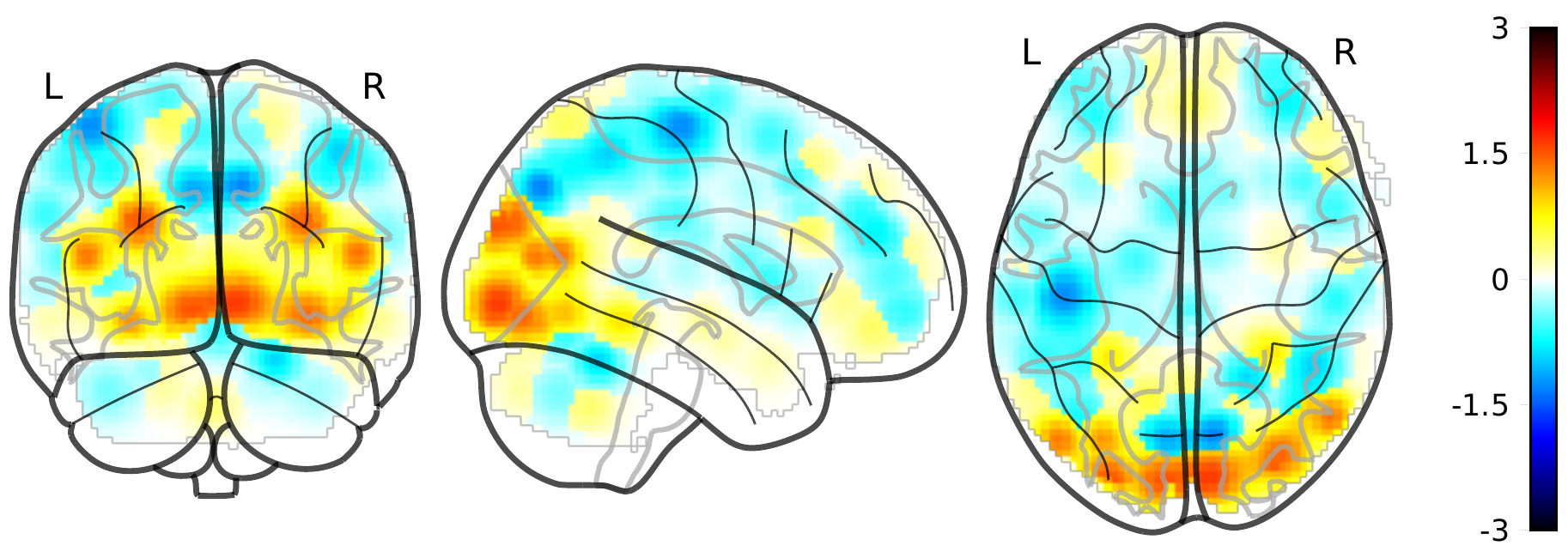}
    \end{tabular}
    \end{center}
    \vspace{-0.5em}
    \caption{\textbf{Test predictions for the ThreatVids dataset}: We compare time-averaged held-out data (\emph{left}) to the posterior-predictive mean for three trials. In NTFA (\emph{center}), the learned generative model and inferred embeddings inform the distribution for unseen participant-stimulus combinations. In HTFA (\emph{right}) the predictive distribution is the same for trials, since the shared global template in this model does not differentiate between participants and stimuli.}
    \label{fig:affvids-reconstruction}
    \vspace{-0.5em}
\end{figure*}

\begin{figure*}[!h]
    \begin{tabular}{ccc}
        \textsf{\small Held-out data} & \textsf{\small Posterior predictive (NTFA)} & \textsf{\small Posterior predictive (HTFA)} \\
        \includegraphics[width=0.31\textwidth]{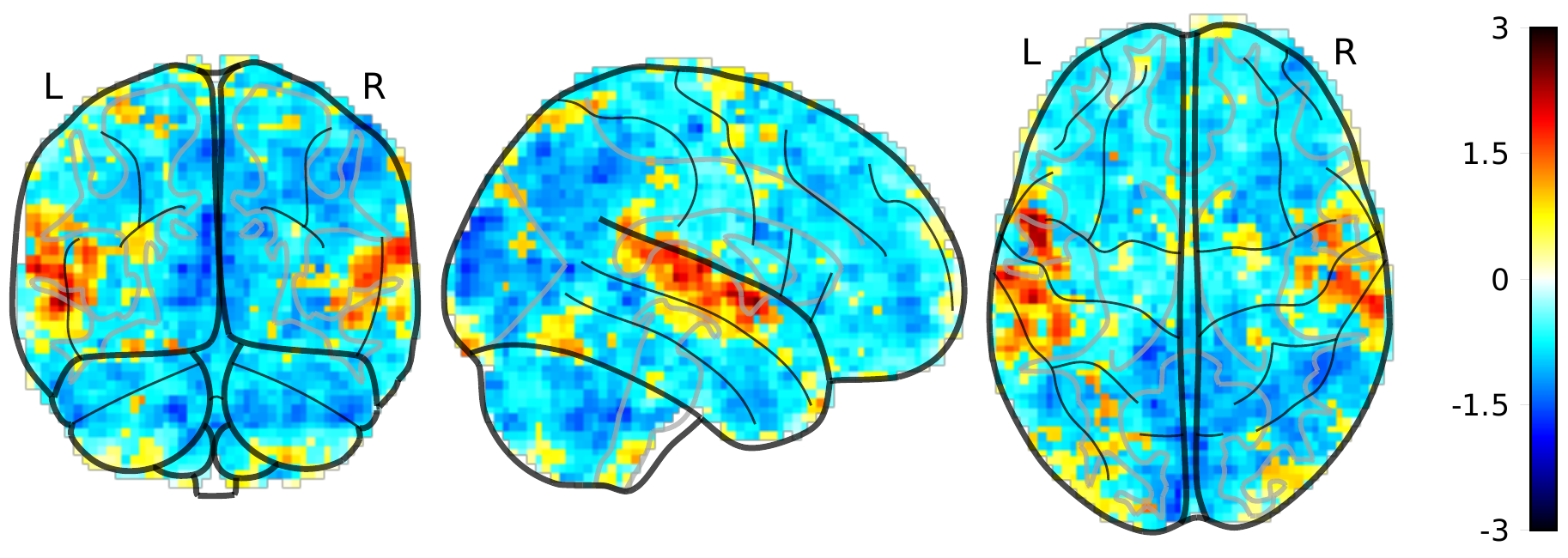} &
        \includegraphics[width=0.31\textwidth]{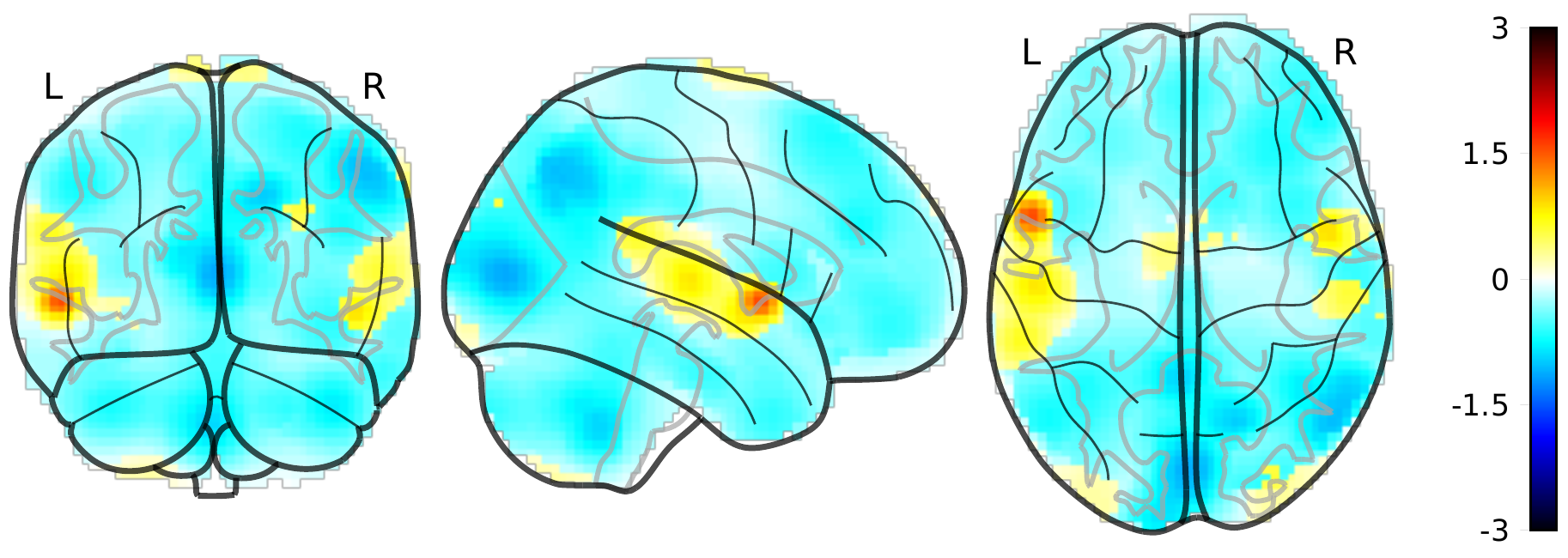} &
        \includegraphics[width=0.31\textwidth]{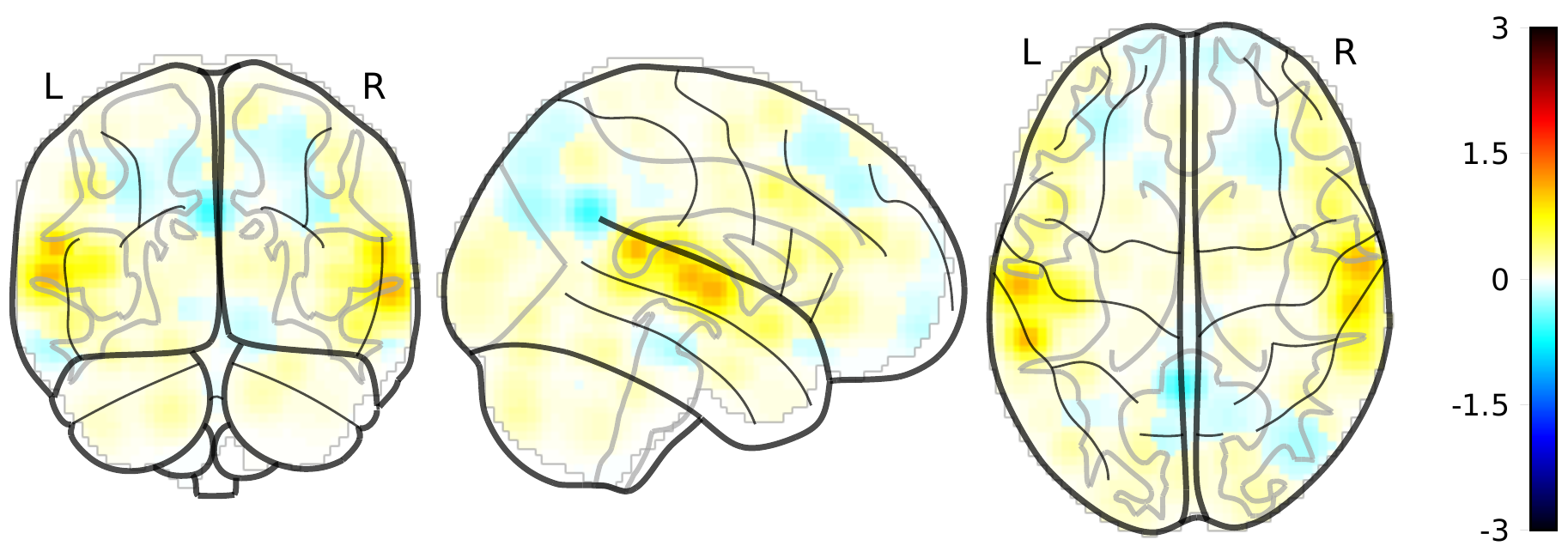} \\
        \includegraphics[width=0.31\textwidth]{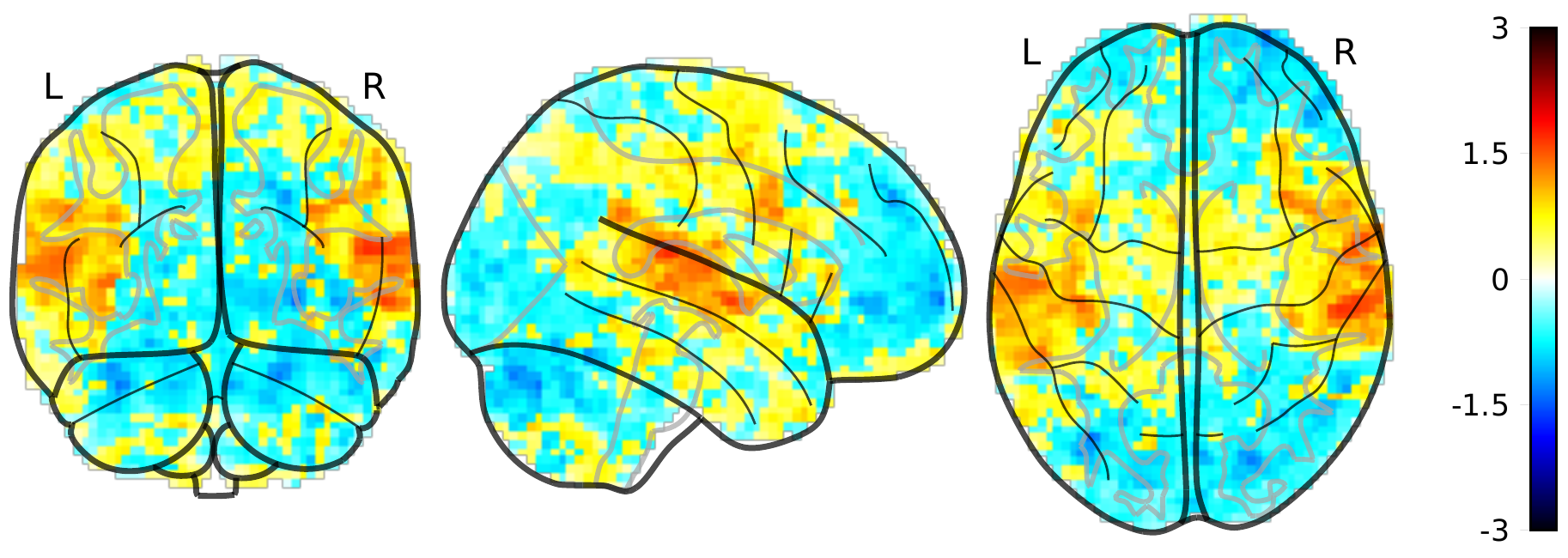} &
        \includegraphics[width=0.31\textwidth]{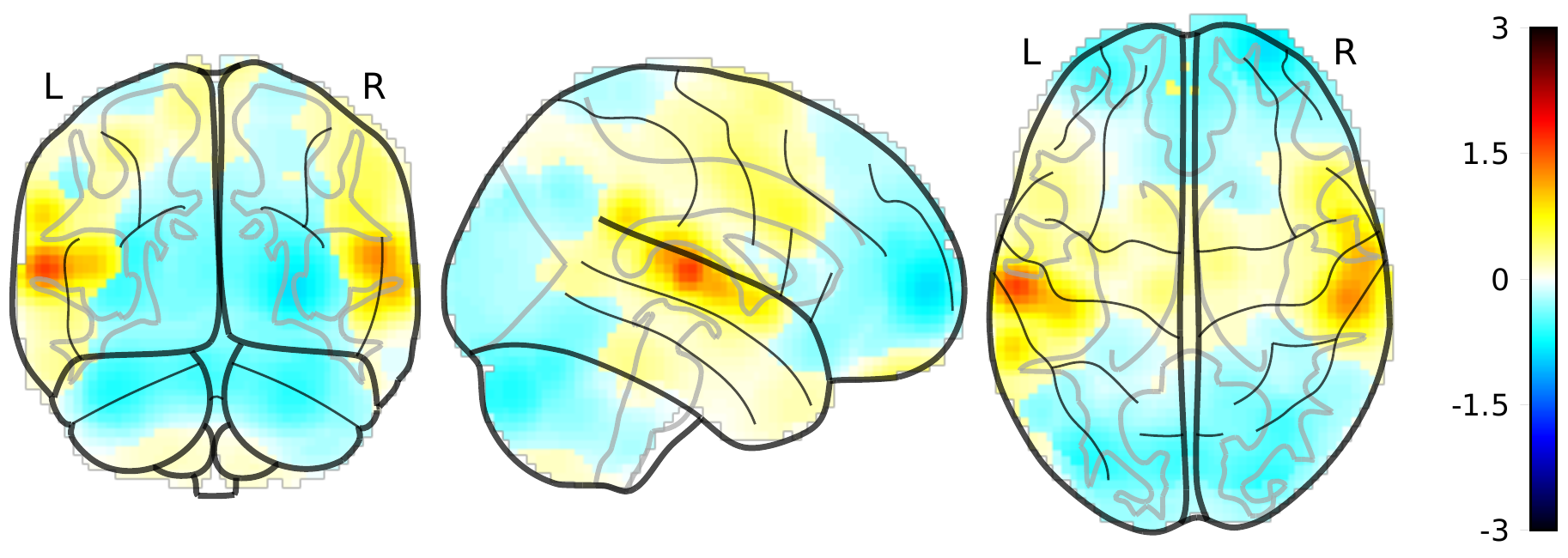} &
        \includegraphics[width=0.31\textwidth]{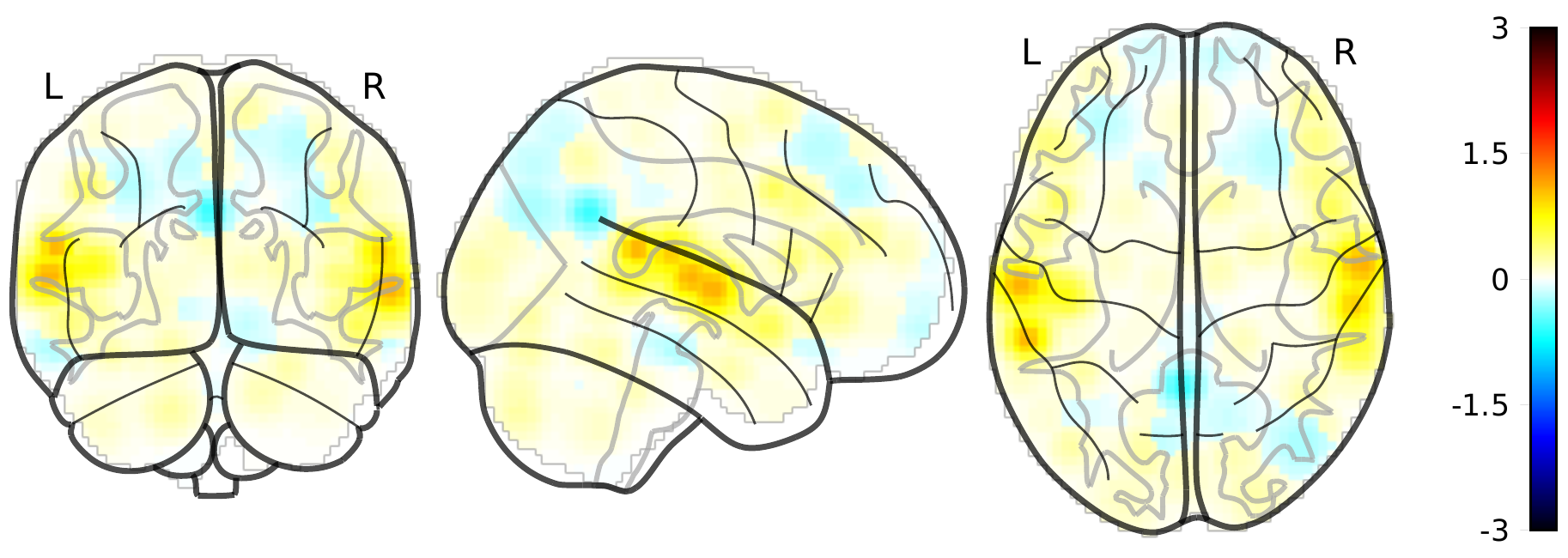} \\
        \includegraphics[width=0.31\textwidth]{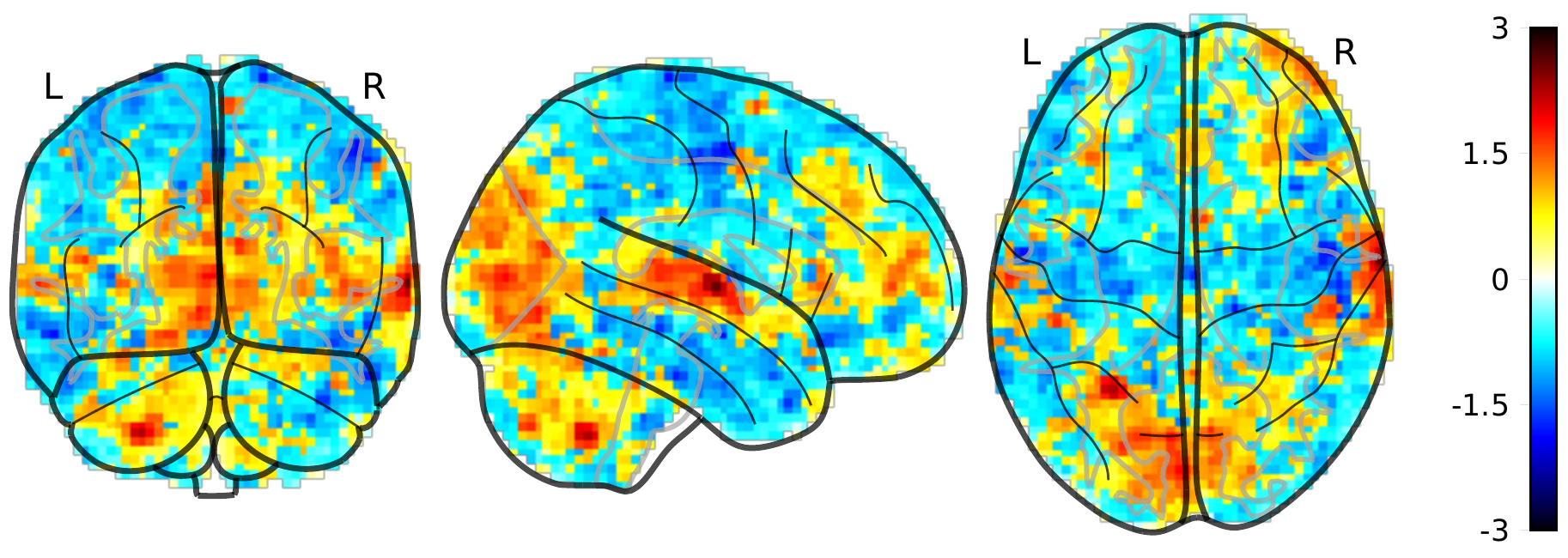} &
        \includegraphics[width=0.31\textwidth]{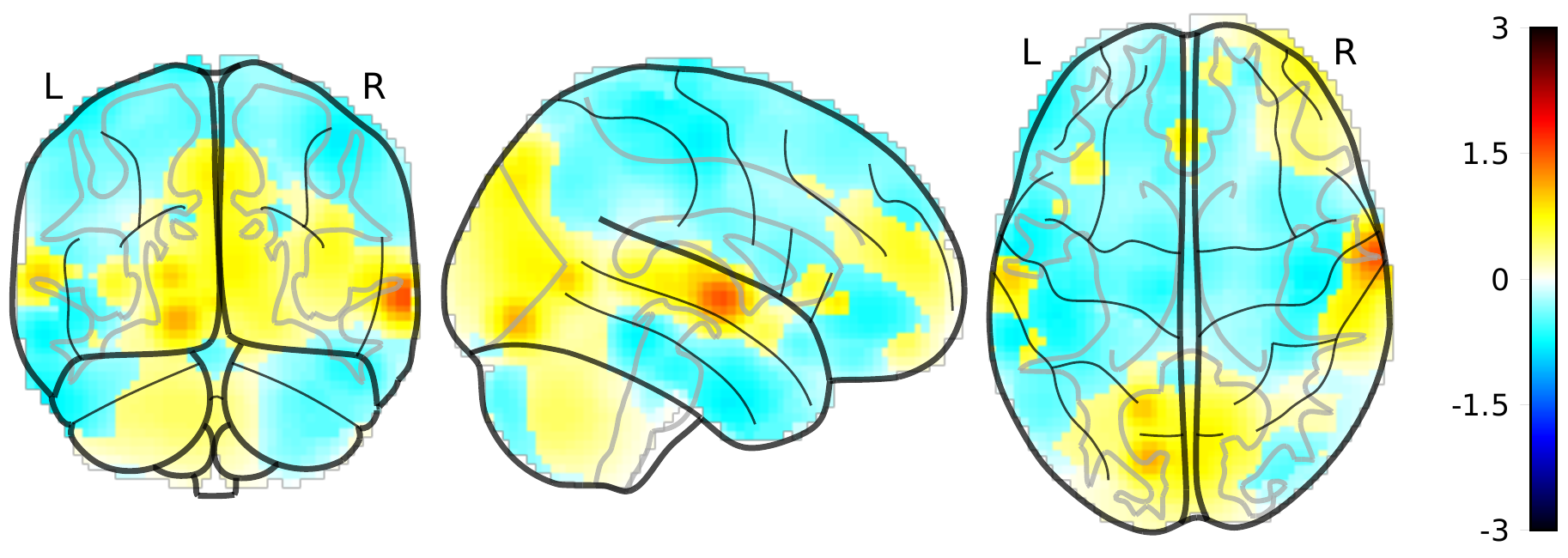} &
        \includegraphics[width=0.31\textwidth]{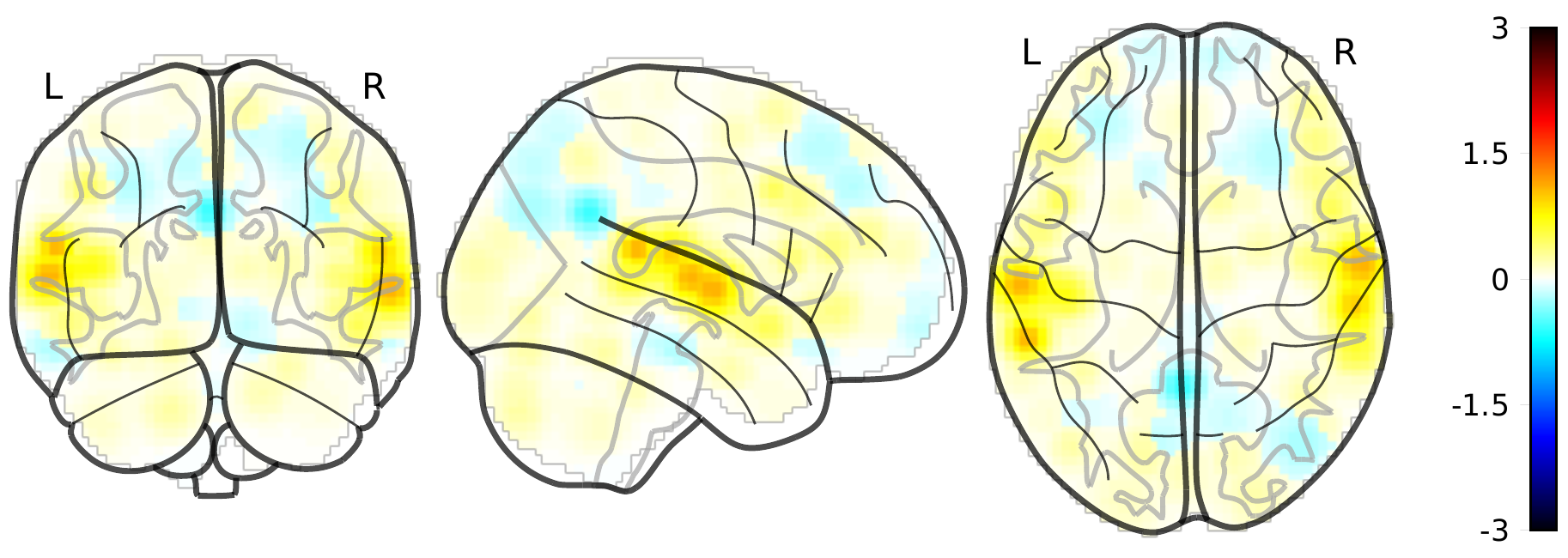}
    \end{tabular}
    \caption{\textbf{Test predictions for the Lepping dataset}: We show average images for three trials, with participant-stimulus pairs held out from the training set.   Posterior predictive estimates under NTFA (center) capture meaningful trial-specific variation in the original images (left), whereas HTFA can only re-use its same global template across differing trials (right).}
    \label{fig:lepping-reconstruction}
\end{figure*}

\begin{figure*}[!h]
    \begin{tabular}{ccc}
        \textsf{\small Held-out data} & \textsf{\small Posterior predictive (NTFA)} & \textsf{\small Posterior predictive (HTFA)} \\
        \includegraphics[width=0.31\textwidth]{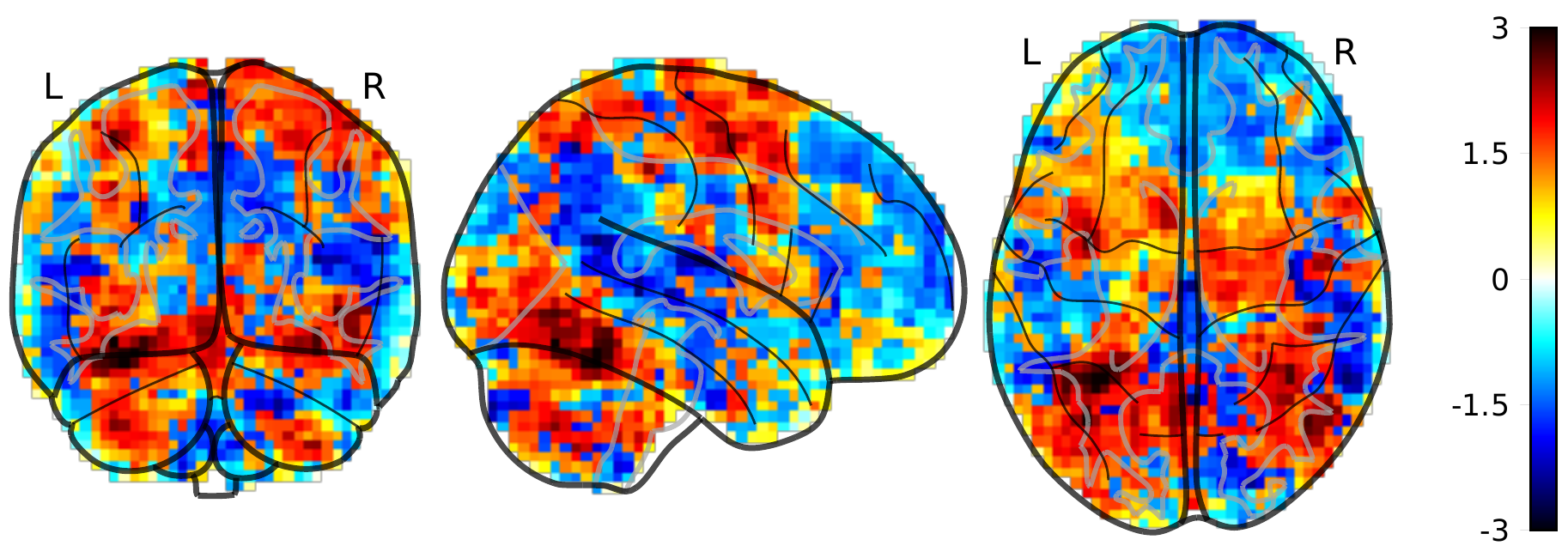} &
        \includegraphics[width=0.31\textwidth]{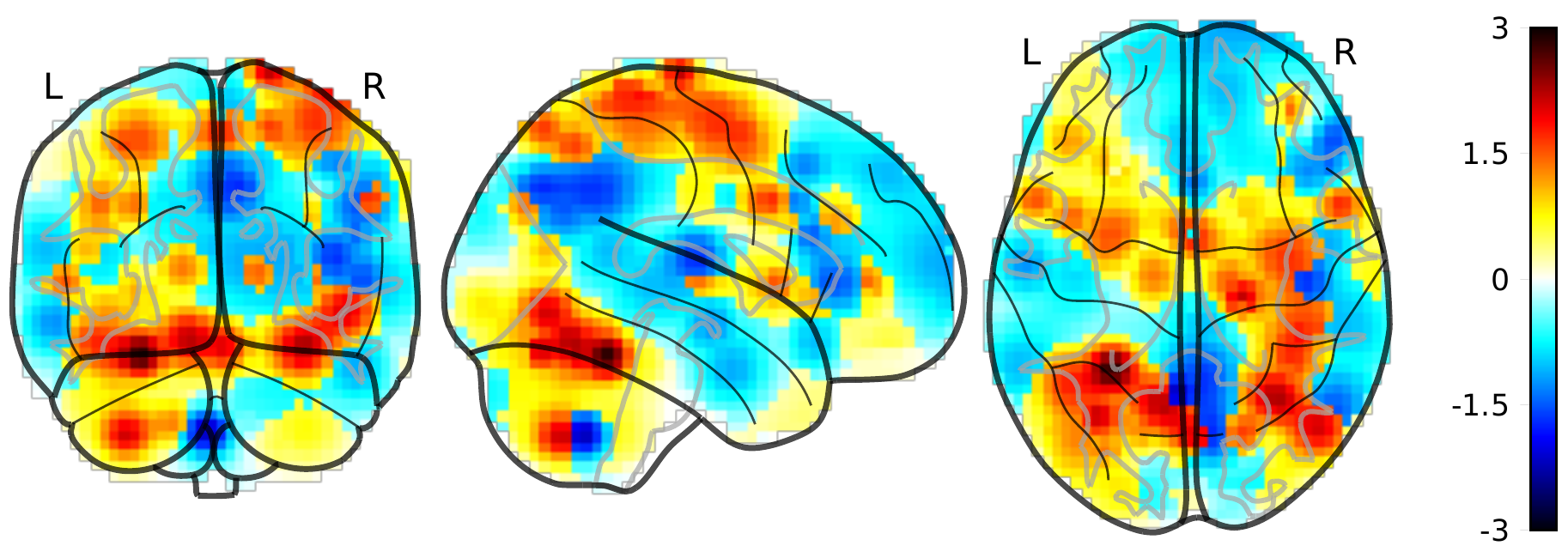} &
        \includegraphics[width=0.31\textwidth]{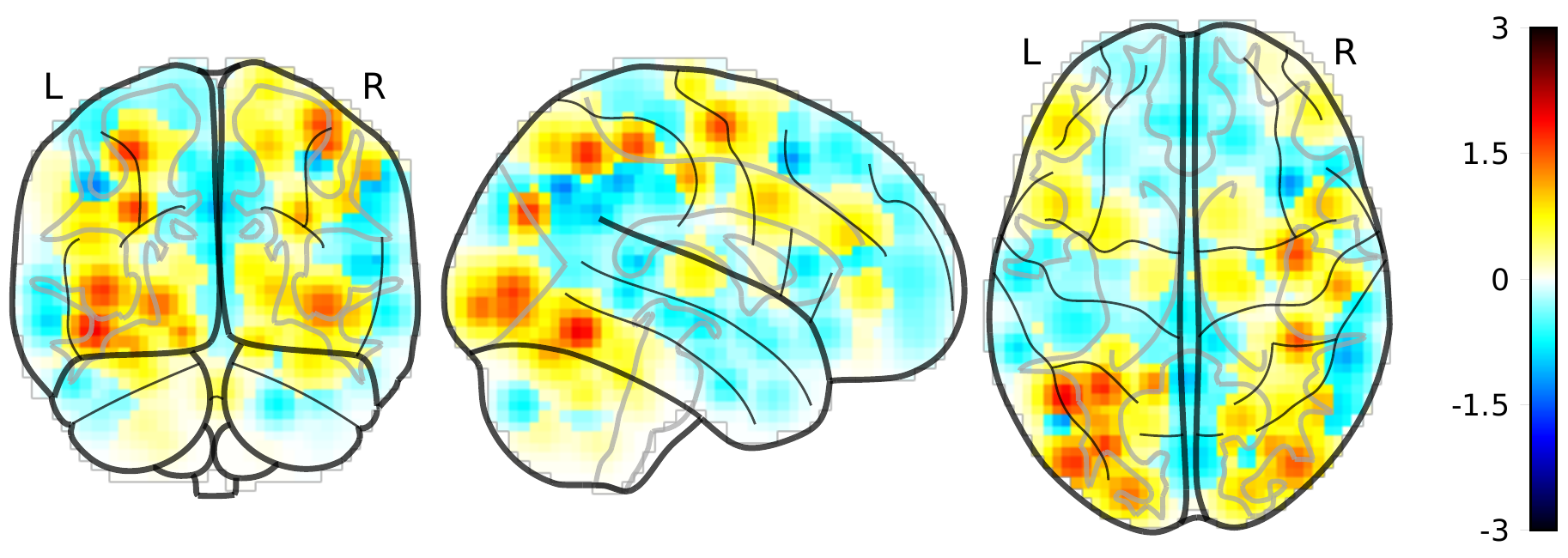} \\
        \includegraphics[width=0.31\textwidth]{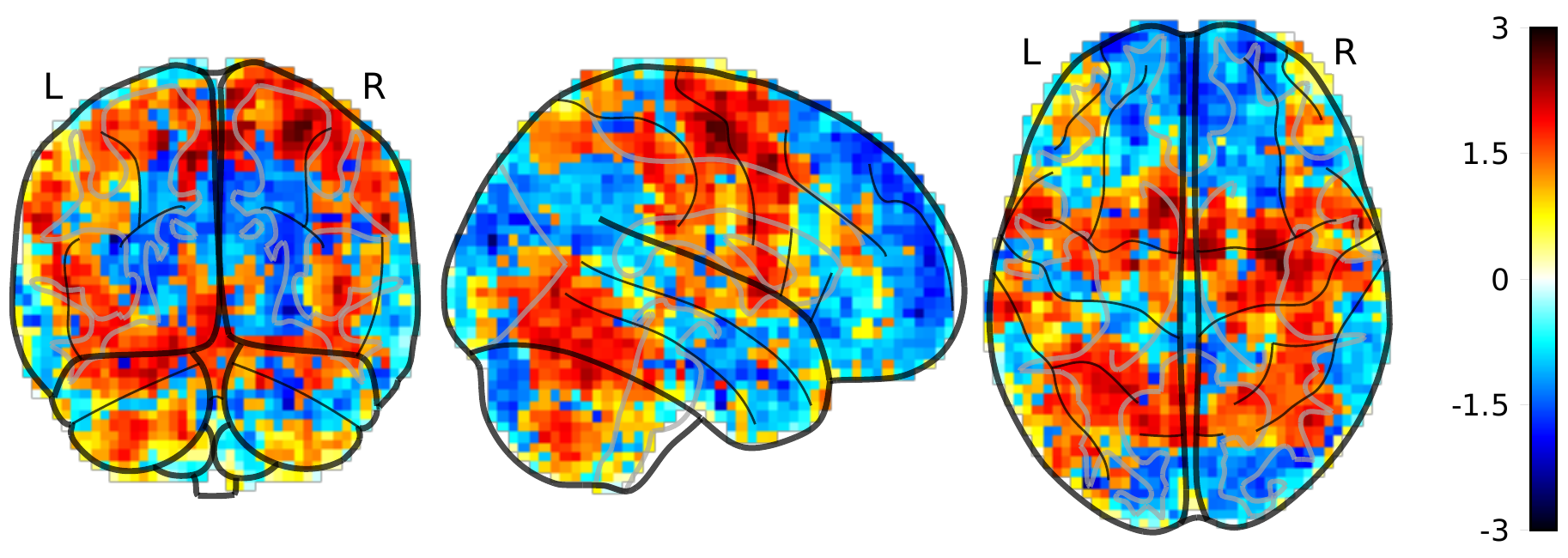} &
        \includegraphics[width=0.31\textwidth]{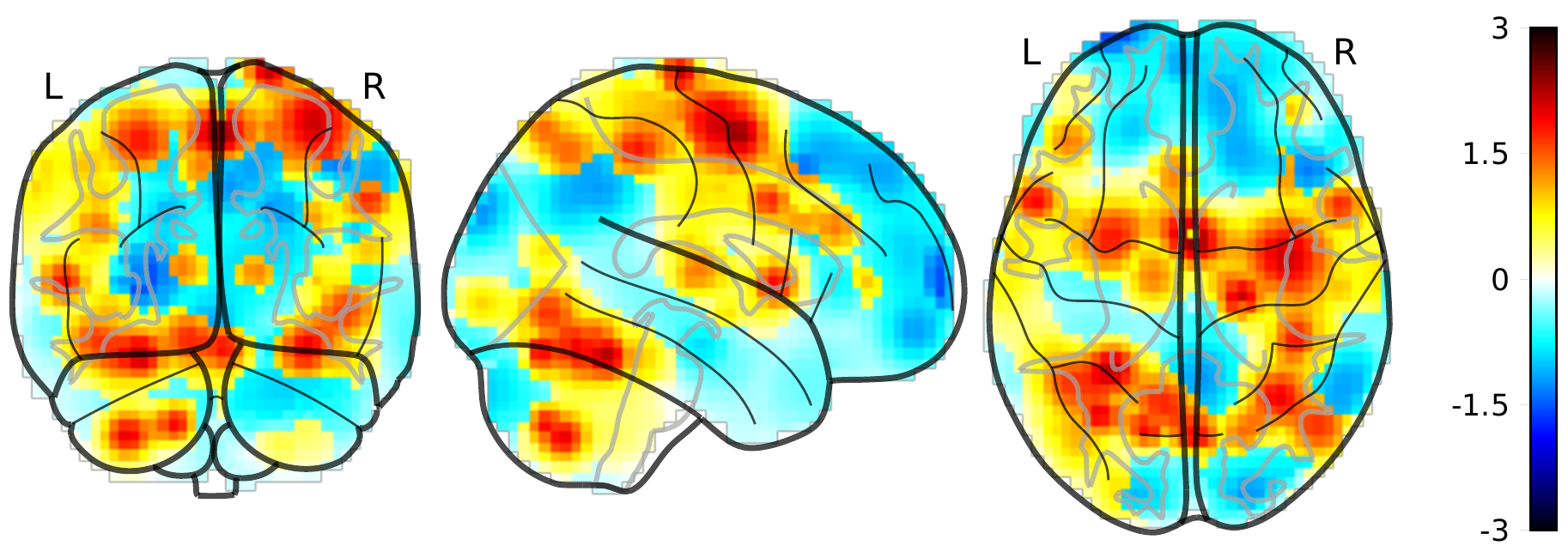} &
        \includegraphics[width=0.31\textwidth]{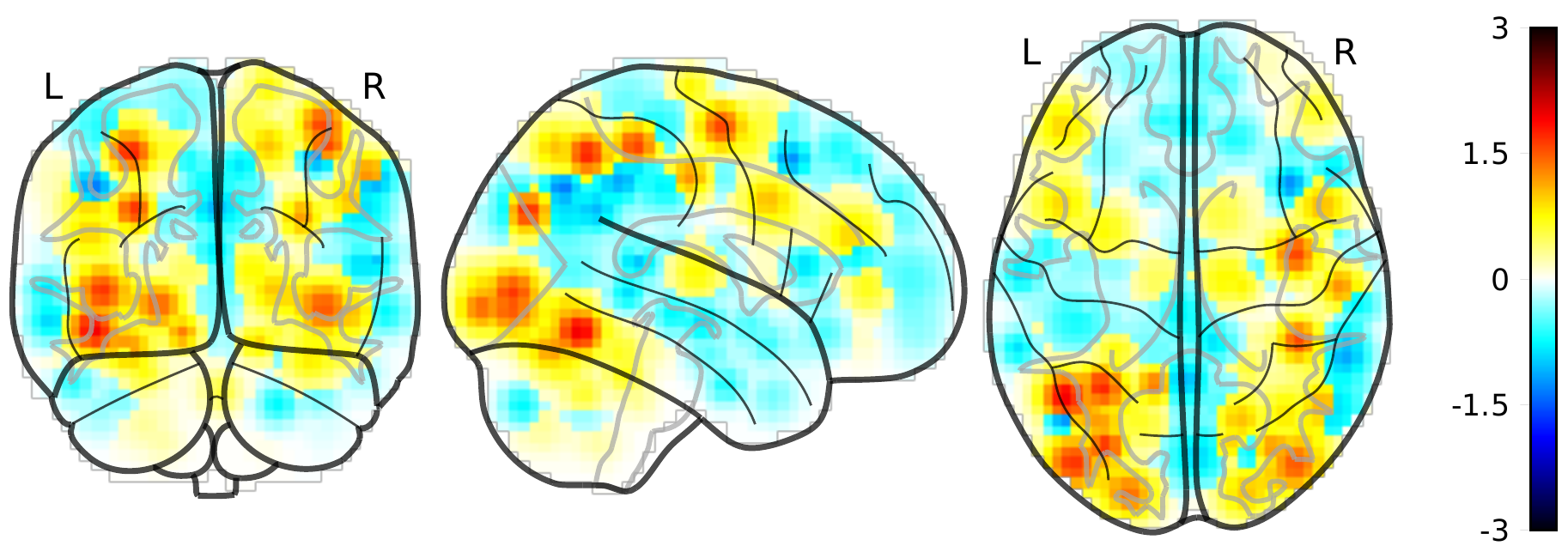} \\
        \includegraphics[width=0.31\textwidth]{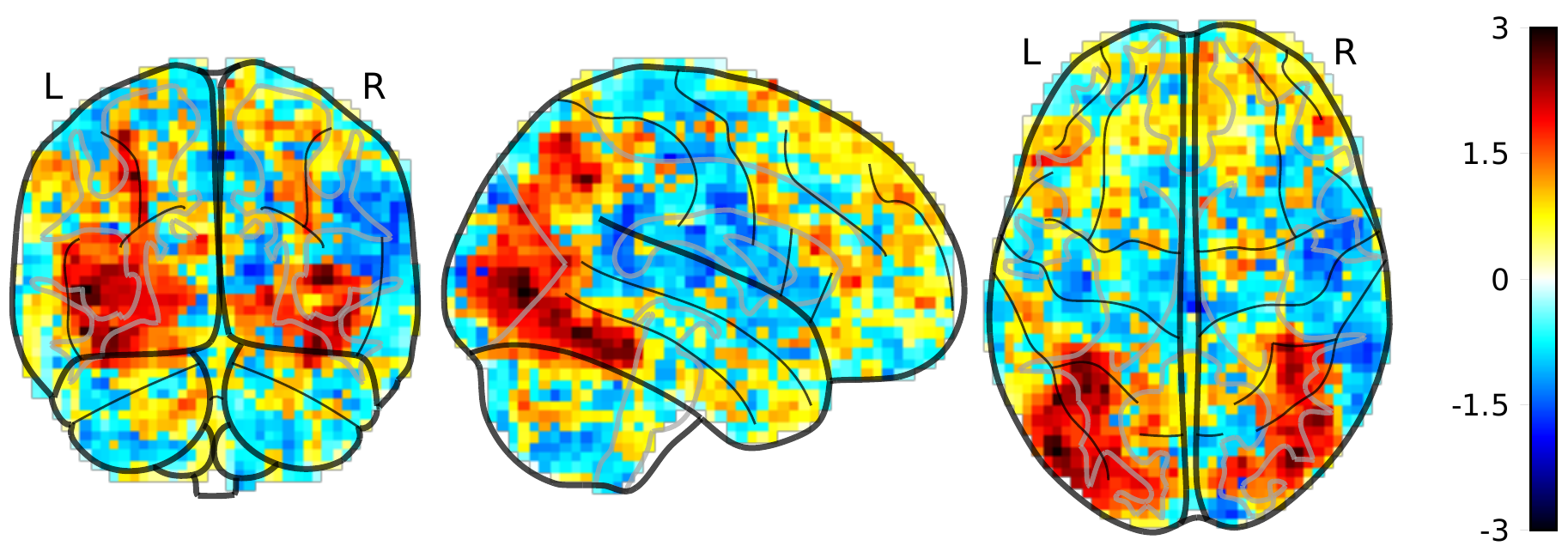} &
        \includegraphics[width=0.31\textwidth]{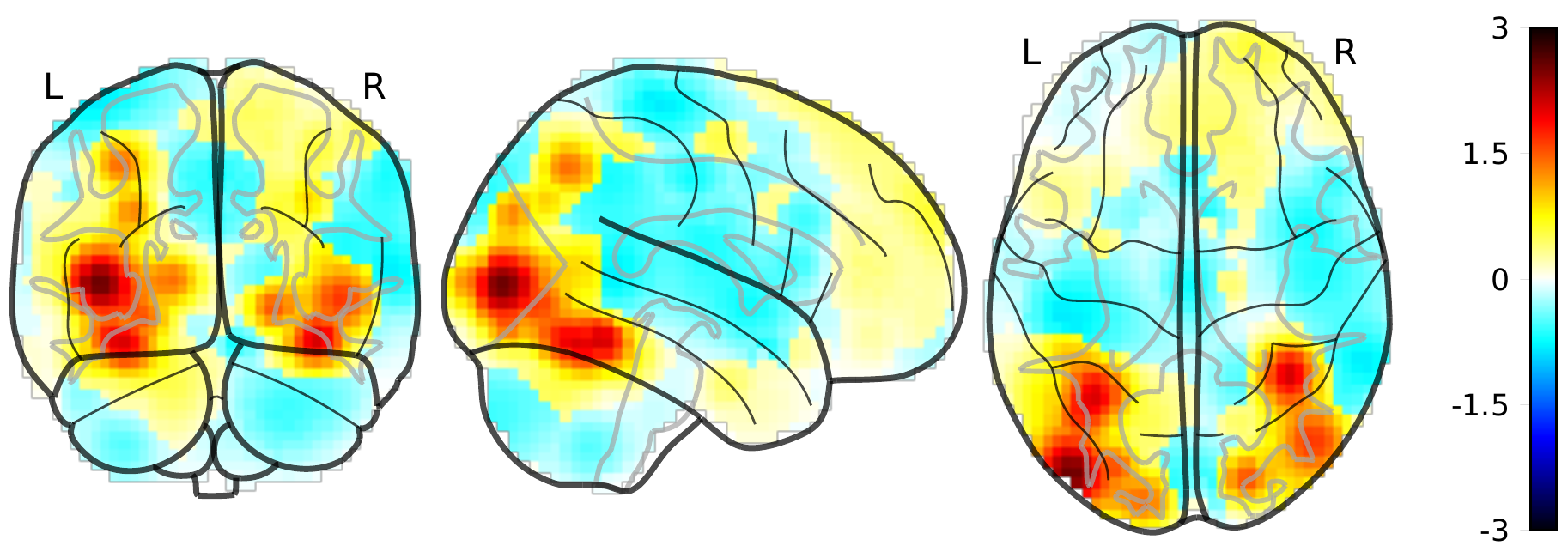} &
        \includegraphics[width=0.31\textwidth]{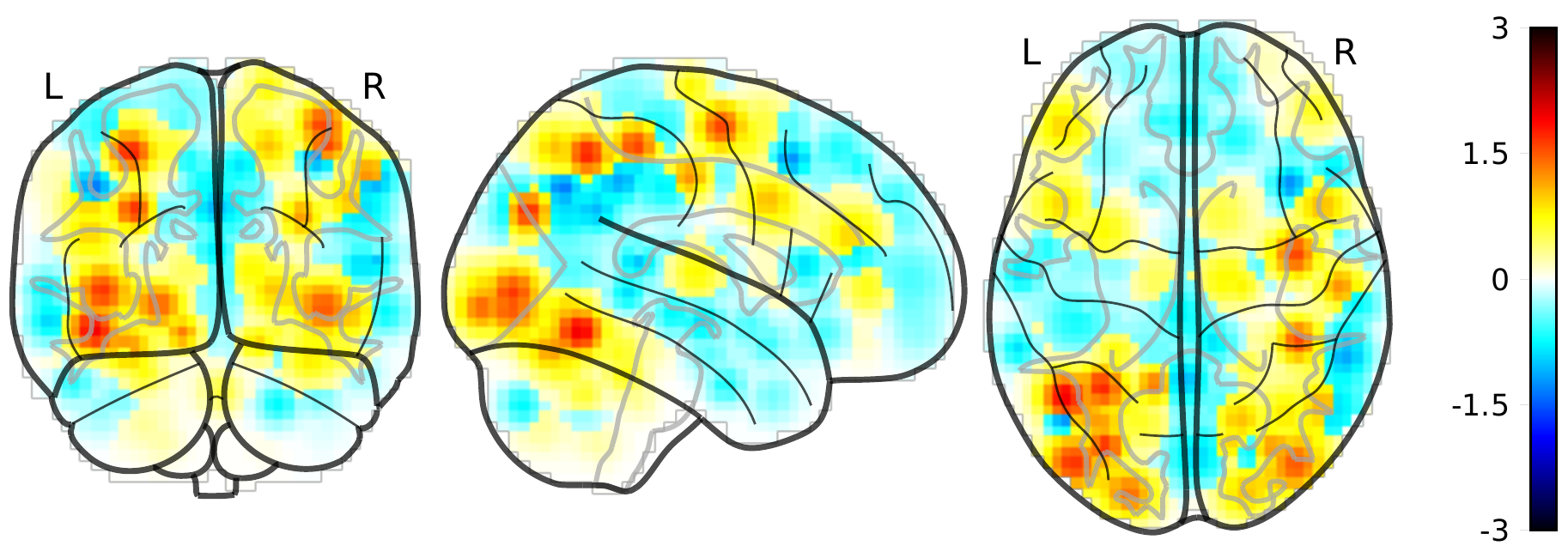}
    \end{tabular}
    \caption{\textbf{Test predictions for the Haxby dataset}: We show average images for three trials, with participant-stimulus pairs held out from the training set.  Posterior predictive estimates under NTFA (center) capture meaningful trial-specific variation in the original images (left), whereas HTFA can only re-use its same global template across differing trials (right).}
    \label{fig:haxby-reconstruction}
\end{figure*}

Here we show test predictions from both NTFA and HTFA for our real datasets, ThreadVids (Figure~\ref{fig:affvids-reconstruction}), Lepping (Figure~\ref{fig:lepping-reconstruction}) and Haxby (Figure~\ref{fig:haxby-reconstruction}).  HTFA's pale experiment-wide averages on the Lepping dataset show its inability to capture the participant- and stimulus-wise variations captured clearly by NTFA.  While HTFA does make clear predictions for its average across the Haxby dataset, nonetheless, it does not capture the variation across trials that NTFA does.

\subsection{Derivation of the lower bound to the log posterior predictive distribution}
\label{app:log-predictive-bound}

We begin by showing how to use the variational distribution to approximate the posterior predictive distribution via importance sampling, and then convert the resulting importance weight into a lower bound on the log posterior predictive.
Posterior sampling from the NTFA generative model, conditioned upon the posterior distribution over embeddings, would yield the joint distribution
\begin{align*}
    p \big(
        \tilde{Y},
        \tilde{W},
        \tilde{x}^{\scf},
        \tilde{\rho}^{\scf},
        z^{\scp},
        z^{\scs}
        \mid
        Y
    \big) &= p \big(
        \tilde{Y},
        \tilde{W},
        \tilde{x}^{\scf},
        \tilde{\rho}^{\scf}
        \mid 
        z^{\scp},
        z^{\scs}
    \big) p \big(z^{\scp}, z^{\scs} \mid Y \big),
\end{align*}
which factorizes according to the generative model as,
\begin{align*}
    p \big(
        \tilde{Y},
        \tilde{W},
        \tilde{x}^{\scf},
        \tilde{\rho}^{\scf},
        z^{\scp},
        z^{\scs}
        \mid
        Y
    \big) &= p \big(
        \tilde{Y}
        \mid
        \tilde{W},
        \tilde{x}^{\scf},
        \tilde{\rho}^{\scf},
        z^{\scp},
        z^{\scs}
    \big)
    p \big(
        \tilde{W}
        \mid 
        z^{\scp},
        z^{\scs}
    \big)
    p \big(
        \tilde{x}^{\scf},
        \tilde{\rho}^{\scf}
        \mid 
        z^{\scp}
    \big)
    p \big(
        z^{\scp}, z^{\scs} \mid Y 
    \big).
\end{align*}
The marginal of this joint distribution, that being the posterior predictive distribution, can be defined by importance weighting, where the learned variational distributions $q(z^\scp)$, $q(z^\scs)$ serve as proposals for $z^\scp$ and $z^\scs$ while the generative model serves as its own proposal for the other latent variables, yielding
\begin{align*}
    p\big(\tilde{Y} \mid Y\big) 
    &= 
    \mathbb{E}_{q,p} \left[\frac{p \left(
        \tilde{Y},
        \tilde{W},
        \tilde{x}^{\scf},
        \tilde{\rho}^{\scf},
        z^{\scp},
        z^{\scs}
        \mid
        Y
    \right)}{p\left(\tilde{W}, \tilde{x}^{\scf}, \tilde{\rho}^{\scf} \mid z^\scp, z^\scs \right) q\left(z^{\scp}\right) q\left(z^{\scs}\right)}\right]\\
    &= \mathbb{E}_{q,p} \left[\frac{p \left(
        \tilde{Y}
        \mid
        \tilde{W},
        \tilde{x}^{\scf},
        \tilde{\rho}^{\scf},
        z^{\scp},
        z^{\scs}
    \right)
    p \left(
        \tilde{W}
        \mid 
        z^{\scp},
        z^{\scs}
    \right)
    p \left(
        \tilde{x}^{\scf},
        \tilde{\rho}^{\scf}
        \mid 
        z^{\scp}
    \right)
    p \left(z^{\scp}, z^{\scs} \mid Y \right)}{p\left(\tilde{W} \mid z^\scp, z^\scs \right) p\left(\tilde{x}^{\scf}, \tilde{\rho}^{\scf} \mid z^{\scp}\right) q\left(z^{\scp}\right) q\left(z^{\scs}\right)} \right]\\
    &= \mathbb{E}_{q,p} \left[\frac{p \left(
        \tilde{Y}
        \mid
        \tilde{W},
        \tilde{x}^{\scf},
        \tilde{\rho}^{\scf},
        z^{\scp},
        z^{\scs}
    \right)
    p \left(z^{\scp}, z^{\scs} \mid Y \right)}{q\left(z^{\scp}\right) q\left(z^{\scs}\right)}\right].
\end{align*}
We can then apply Jensen's inequality to define a lower bound
\begin{align*}
    & \text{ELBO}_{\tilde{Y}\mid Y} \\
    &=
    \mathbb{E}_{q,p} \left[\log \frac{p \left(
        \tilde{Y}
        \mid
        \tilde{W},
        \tilde{x}^{\scf},
        \tilde{\rho}^{\scf},
        z^{\scp},
        z^{\scs}
    \right)
    p \left(z^{\scp}, z^{\scs} \mid Y \right)}{q\left(z^{\scp}\right) q\left(z^{\scs}\right)}\right] \\ 
    &\leq \log p\left(\tilde{Y} \mid Y\right).
\end{align*}
This is a standard definition of the ELBO, albeit for the posterior predictive distribution rather than the marginal likelihood (i.e.~the prior predictive). By converting the log of a product of densities into a sum of log-density terms and noting that the expectations are over proposal distributions $p = p\left(\tilde{W}, \tilde{x}^{\scf}, \tilde{\rho}^{\scf} \mid z^\scp, z^\scs \right)$ and $ q = q\left(z^{\scp}\right) q\left(z^{\scs}\right)$, we can write this ELBO as:
\begin{align*}
    \text{ELBO}_{\tilde{Y}\mid Y} &= \mathbb{E}_{q,p} \left[\log p \left(
        \tilde{Y}
        \mid
        \tilde{W},
        \tilde{x}^{\scf},
        \tilde{\rho}^{\scf},
        z^{\scp},
        z^{\scs}
    \right) - \log \frac{q(z^\scp)q(z^\scs)}{p(z^\scp,z^\scs \mid Y)}\right] \\
    \text{ELBO}_{\tilde{Y}\mid Y} &= \mathbb{E}_{q,p} \left[\log p \left(
        \tilde{Y}
        \mid
        \tilde{W},
        \tilde{x}^{\scf},
        \tilde{\rho}^{\scf},
        z^{\scp},
        z^{\scs}
    \right)\right] - \text{KL}\left(q\left(z^{\scp}\right) q\left(z^{\scs}\right) \,||\, p \left(z^{\scp}, z^{\scs} \mid Y \right)\right),
\end{align*}
From the standard decomposition of the ELBO we can also reason that,
\begin{align*}
    \text{ELBO}_{\tilde{Y}\mid Y} &= \log p(\tilde{Y} \mid Y) - \text{KL}\left(q\left(z^{\scp}\right) q\left(z^{\scs}\right) \,||\, p \left(z^{\scp}, z^{\scs} \mid \tilde{Y}, Y \right)\right),
\end{align*}
and therefore
\begin{align*}
        \mathbb{E}_{q,p} \left[\log p \left(
        \tilde{Y}
        \mid
        \tilde{W},
        \tilde{x}^{\scf},
        \tilde{\rho}^{\scf},
        z^{\scp},
        z^{\scs}
    \right)\right] &- \text{KL}\left(q\left(z^{\scp}\right) q\left(z^{\scs}\right) \,||\, p \left(z^{\scp}, z^{\scs} \mid Y \right)\right) =\\ \log p(\tilde{Y} \mid Y) &- \text{KL}\left(q\left(z^{\scp}\right) q\left(z^{\scs}\right) \,||\, p \left(z^{\scp}, z^{\scs} \mid \tilde{Y}, Y \right)\right),\\
    \mathbb{E}_{q,p} \left[\log p \left(
        \tilde{Y}
        \mid
        \tilde{W},
        \tilde{x}^{\scf},
        \tilde{\rho}^{\scf},
        z^{\scp},
        z^{\scs}
    \right)\right] &= \log p(\tilde{Y} \mid Y)\\ &- \text{KL}\left(q\left(z^{\scp}\right) q\left(z^{\scs}\right) \,||\, p \left(z^{\scp}, z^{\scs} \mid \tilde{Y}, Y \right)\right)\\ &+ \text{KL}\left(q\left(z^{\scp}\right) q\left(z^{\scs}\right) \,||\, p \left(z^{\scp}, z^{\scs} \mid Y \right)\right).
\end{align*}
Since the variational distributions $q$ were already optimized during training to minimize the KL divergence in the third term on the right hand side of the above equation, we can reason that it will be small compared to the KL divergence in the second term (between the variational distribution and the true posterior given the test data).  The difference of KL's should therefore remain nonnegative, allowing us to use the expected log-likelihood as a lower bound to the log posterior predictive probability of the test data.  Additionally, the low dimensionality ($D=2$ in our experiments) of $z^\scp$ and $z^\scs$ compared to $Y$ led the log likelihood to dominate the ELBO in all our experiments, a fact which should not be changed by passing to $ELBO_{\tilde{Y}\mid Y}$. This leads to Equation~\eqref{eq:post_pred_lb} in Section~\ref{sec:evaluation}.

\subsection{Description of datasets}
\label{sec:si_datasets}

\textbf{Threat Videos (``ThreatVids'')}\footnote{This dataset is currently in preparation for online repository pending deidentification and submission of an empirical report.}: 
A fundamental question in affective neuroscience is whether threat-relevant stimuli from different categories involve a single or multiple distinct systems \citep{wager2015bayesian}. To evaluate whether NTFA can provide insight into this fundamental research question, we conducted and analyzed our own study. 21 participants each watched 36 videos depicting threat-related content involving spiders, looming heights, and social evaluative threat (12 videos per category). Each video was approximately 20 seconds long and was followed by a set of self-report ratings and a rest period. The videos were chosen to vary in how much fear they normatively evoke within each category. This data contains 81,638 voxels (white matter removed) and 552 time points per scanning run for three runs. Using NTFA, we examine whether neural activity justified organization of the stimuli into three categories.

\textbf{Emotional Musical and Nonmusical Stimuli in Depression (``Lepping'')}
\citep{10.1371/journal.pone.0156859}\footnote{This data was obtained from the OpenfMRI database. Its accession number is ds000171.}: 19 participants with major depressive disorder and 20 control participants (P=39) underwent musical and nonmusical stimuli to examine neural processing of emotionally provocative auditory stimuli in depression. In each trial, participants listened to music or nonmusical valenced (positive or negative) sounds, interleaved with trials in which they heard neutral tones. The fMRI data had 62,733 voxels (white matter removed) and 105 time points in each scanning run for five runs.

\textbf{Face and Objects Image Viewing (``Haxby'')} \citep{Haxby2425}\footnote{\scriptsize \url{https://openneuro.org/datasets/ds000105/}}: fMRI was used to measure whole brain response while subjects viewed faces, cats, five categories of man-made objects, and scrambled pictures.  The study consisted of six subjects (P=6) and 12 scanning runs per subject, with 32,233 voxels (white matter removed) and 121 time points for each scanning run.
\begin{table}[!h]
\caption{\textbf{Dataset Summary}}
\label{app:tb_data}
\centering
\begin{tabular}{lcccc}
\toprule
 & \textbf{No. Participants} & \textbf{No. of Stimuli} & \textbf{TRs per block} & \textbf{No. of Voxels} \\
 & \textbf{P} & \textbf{S} & \textbf{T} & \textbf{V} \\
\midrule
~Synthetic & 9 & 8 & 20 & 5,000 \\
~ThreatVids & 23 & 36 & 20 & 81,638 \\
~Lepping & 39 &  4 & 10 & 62,733 \\
~Haxby & 6 & 8 & 12 & 32,233 \\
\bottomrule
\end{tabular}
\end{table}

\subsection{Preprocessing}
\label{sec:si_preprocessing}

The raw BOLD signal collected in fMRI is generally not usable for analysis.  It contains both physiological (cerebrospinal fluids, global signal, and white matter) and motion artifacts.  We employ standard neuroimaging preprocessing - including slice timing correction, high pass filtering and spatial smoothing - for all fMRI data using fMRIPrep \citep{esteban2019fmriprep}.  The processed data still has units that are incomparable across scanning runs. For the ThreatVids and Haxby datasets, we z-scored each task trial with respect to the entire set of rest trials (trials in which no stimulus was presented) within each run. For the Lepping dataset, we treated neutral tones as rest trials and performed the same z-scoring procedure.  This provides a common scale of units across trials within a dataset, capturing meaningful difference in activation intensities relative to neutral conditions (``rest'' and ``tones''). Since neural activity peaks about three seconds after the task onset \citep{aguirre1998variability}, we make sure to account for this delay when loading each dataset by offsetting the stimulus onsets by three seconds.  We input the resulting z-scored data to NTFA, and use it for all evaluations.
\subsection{Neural network architectures and initialization}
\label{subsec:neural-net-architectures}
The network $\eta^\scf_\theta(\cdot)$ is a multilayer perceptron (MLP) with one hidden layer and PReLU activations.  We extract the factor parameters $\scf, x$ and $\scf, \rho$ by viewing the $8K$-dimensional result as a $K\times 4 \times 2$ tensor. The network $\eta^\scw_\theta(\cdot)$ is similarly an MLP with one hidden layer, though operating over both embeddings.  We extract the weight parameters $\scw_{n}$ by casting its $2K$-dimensional result as a $K\times 2$ matrix. Architectural details for both networks are given in Table~\ref{table:network-architectures}.  The neural network weights $\theta$ specify the linear layers of the networks.

We train the parameters $\theta$ and $\lambda$ on all models using the Adam optimizer \citep{Kingma2015} for 1000-1500 epochs per dataset. We use one particle to calculate the IWAE-style bound to the log-evidence and its gradient estimator at training time, with a learning rate $\eta_\lambda = 0.01$ and $\eta_\theta = 0.0001$.  We anneal the learning rate with a patience of 100 epochs, and a multiplicative decline of 0.5. 

Similarly to \citet{manning2018probabilistic}, we employ a K-means initialization in all experiments across models, initializing the variational parameters for HTFA and the bias in the final layer of the generative model for NTFA.
\begin{table}[!t]
\centering
\caption{\textbf{Network architectures for} $\eta^\scf_\theta$ and $\eta^\scw_\theta$}
\label{table:network-architectures}
\small
\begin{tabular}{ccc}
    \toprule
    \textbf{Layer} 
    & 
    $p_\theta\left(x^\scf_p, \rho^\scf_p \mid z^\scp_p \right)$ 
    &
    $p_\theta\left(W_{n,t} \mid z^{\scp}_p, z^{\scs}_s \right)$
    \\
    \midrule
    Input
    & 
    $z^{\scp}_p \in\mathbb{R}^{D}$
    &
    $z^{\scp}_p, z^{\scs}_s \in\mathbb{R}^{D}$
    \\
    \midrule
    1 & 
    FC $D \!\times\! 2D$ PReLU &
    FC $2D \!\times\! 4D$ PReLU
    \\
    2 & 
    FC $2D \!\times\! 4D$ PReLU &
    FC $4D \!\times\! 8D$ PReLU 
    \\
    3 & 
    FC $4D \!\times\! 8K$ &
    FC $8D \!\times\! 2K$ 
    \\
    \midrule
    Output & 
    $\!\!\left(\mu^{x}_{p}, \sigma^{x}_{p}, \mu^{\rho}_{p}, \sigma^{\rho}_{p}\right) \in \mathbb{R}^{8K}\!\!$ &
    $\!\!\left(\mu^{\textsc{w}}_{n}, \sigma^{\textsc{w}}_{n}\right) \in \mathbb{R}^{2K}\!\!$
    \\
    \bottomrule
\end{tabular}
\end{table}

\subsection{Synthetic data generation}
\label{app:synthetic-data}
We consider a simulated dataset in which there are three participant groups (\emph{Group 1}, \emph{Group 2} and \emph{Group 3}) of three participants each. All participants underwent two categories of hypothetical stimuli, called \emph{Task 1} and \emph{Task 2}, with four stimuli within each category.  Each participant underwent one hypothetical scanning run with rest trials interleaved between stimuli. We manually defined three distinct factors in a standard MNI\_152\_8mm brain. We then sampled participant embeddings $\{z_1^\scp,...,z_{9}^\scp \}$ and stimulus embeddings $\{ z_1^\scs,...,z_{8}^\scs\}$, from mixtures of three and two distinct Gaussians respectively. We set the means for these Gaussians to meet the following conditions under noisy combination. \textbf{1.} All participants show no whole-brain response during rest except random noise. \textbf{2.} Under \emph{Task 1} stimuli, Group 1 exhibits approximately half the response in the first region as compared to under \emph{Task 2} stimuli.  The rest of the brain shows no response. Similarly, Group 2 and Group 3 exhibit a response in the second and third regions respectively, while the rest of the brain shows no response. \textbf{3.} Each stimulus in \emph{Task 1} and \emph{Task 2} provokes a response lower or higher than the stimulus category's average based on the stimulus embedding's location.
Algorithm~\ref{alg:synthetic_generative} shows the pseudocode for generating synthetic datasets similar to this synthetic data used in this paper. We will also include the exact script with the code repository for our method.

\begin{algorithm}[!h]
   \caption{Generating a simple Synthetic data to test NTFA using Nilearn \citep{abraham2014machine}. For a dataset with $T$ time points per block, $K$ factors, $C$ stimulus categories, $N_{C}$ stimuli per category, $G$ participant groups, and $N_G$ participants per group. This leads to $S=C*N_C+(2N_C+1)$ stimuli (to allow for interleaved rest blocks) and $P=G*N_G$ participants.}
   \label{alg:synthetic_generative}
\begin{algorithmic}[1]
    \State {Load Template Brain} \Comment{e.g.~MNI\_152\_8mm}
    \State Define $\mu^S_{1 \ldots C}$, $\Sigma^S_{1 \ldots C}$ \Comment{means and covariances for embeddings for each stimulus category.}
    \State Define $\mu^P_{1 \ldots G}$, $\Sigma^P_{1 \ldots G}$ \Comment{means and covariances for each participant group.}
    
    \State $x^F_{1,\ldots,K} \leftarrow K$ \Comment{manually selected voxels.}
    
    \State Define $\sigma^x,\mu^\rho,\sigma^\rho$ \Comment{variance for factor centers, means and variance for log-width}
    \State $\rho_{1,\ldots,K} \sim \mathcal{N}(\mu^\rho,\sigma^\rho)$
    
    \State $F \leftarrow \textsc{rbf}(x,\rho)$ \Comment{create factor matrix using radial basis functions}

    \Statex Order the total stimuli according to required experiment design and save indices accordingly. e.g.~Category 1, Rest, Category 2, Rest, and so on.

\For{$c$ \textbf{in} $ 1, \ldots, C$ }
    \For{$s$ \textbf{in} $1, \ldots N_C$}
        $z_{s}^{c} \sim \mathcal{N}(\mu_{c}^S,\Sigma_{c}^S)$ 
    \EndFor
\EndFor
\Comment{generate stimulus embeddings}

\For{$g$ \textbf{in} $ 1, \ldots, G$ }
    \For{$p$ \textbf{in} $1, \ldots N_G$}
        $z_{p}^{g} \sim \mathcal{N}(\mu_{g}^P,\Sigma_{g}^P)$ 
    \EndFor
\EndFor
\Comment{generate participant embeddings}

\For{$g$ \textbf{in} $1, \ldots,G$ }
 \For{$p$ \textbf{in} $1, \ldots N_G$}
  \For{$s$ \textbf{in} $1,\ldots,S$}
   \For {$k$ \textbf{in} $1,\ldots,K$}
   \State $W_{[k,s:s+T]} \sim \mathcal{N}(0,\sigma^W)$ \Comment{if $s$ is the start of a rest block}
  \State $W_{[k,s:s+T]} \sim \mathcal{N}(z_p^Tz_s,\sigma^W)$
   \EndFor
  \EndFor
 \EndFor
$Y_{p + (g-1)*N_G} = WF  $ \Comment{data for one participant}
\EndFor    

\end{algorithmic}
\end{algorithm}

\subsection{MVPA Classification}
\label{app:mvpa}
In this section we provide details of the classification pipeline as well as results beyond those presented in Figure \ref{fig:svm_accuracy}. The traditional pipeline outlined in \citet{Pereira2009} was used to do classification on the input data. One classical approach is to first select a subset of voxels with reliably different mean intensities between the experimental variables being tested.  Usually this is done by selecting $500$ voxels based on the $f$-statistic from an analysis of variance (ANOVA). After this supervised feature selection step, a linear support-vector machine (SVM) is usually trained (without hyperparameter tuning) over some combination of cross-validation scans.

For each block (an instance of a participant undergoing a stimulus), only the mean voxel activity was considered. We employed a leave-out-one cross-validation approach with respect to scanning runs for each subject, with a one-vs-all linear SVM trained and tested for each stimulus category.  For Haxby and ThreatVids, we used a leave-one-out cross-validation approach on scanning runs, while for Lepping (in which the experimental design did not support leaving whole scanning sessions out) we applied a stratified three-fold cross-validation scheme across all trials with the same stimulus.  We then ran the same classification pipeline again, substituting NTFA's and HTFA's MAP estimates of the weight matrix $W$ for the label-supervised voxels.

For each cross validation run, the feature selection of voxels was done by keeping the top $500$ of the voxels with the most reliable differences in the mean intensity for the stimulus category the classifier was being trained for vs the remaining categories. The linear SVM trained on these selected voxels on training runs was then tested on the held-out runs. Since this is a one-vs-all scheme, the classes are unbalanced, and raw accuracy can be inflated just by predicting the most frequent class.  We therefore report Area Under the ROC Curve (AUC) instead.

For NTFA and HTFA, we use the same pipeline, except there is no supervised feature selection step. Instead, we used MAP estimates of generated $W \in \mathbb{R}^{T \times K}$ matrices, averaged across time.  These were employed as the training features for classifiers with respect to the cross-validation scheme above. As is evident from Figure~\ref{fig:svm_accuracy}, and Tables \ref{table:svm_affvids} and \ref{table:svm_lepping}, NTFA performs similarly to the supervised pipeline above, and often outperforms HTFA.

Based on suggestion from a reviewer, we also considered including NTFA training within the cross validation folds. However, we note that this is computationally very expensive, since MVPA here is done with separate cross-validation folds for each subject. This would result in training NTFA more than 50 times for the three real datasets. Moreover, given the unsupervised nature of NTFA, we believe that this shortcut of training NTFA on all data only once to extract features is unlikely to be problematic.

\begin{table*}[!t]
    {\centering
\caption{\textbf{Classification Details} on ThreatVids dataset. ``Voxel'' indicates the ANOVA-SVM strategy on input data. NTFA performs consistently better than HTFA and closer to performing a completely supervised feature selection + classification pipeline on the input data.}
\label{table:svm_affvids}
\scriptsize
    \begin{center}
    \setlength\tabcolsep{3.0pt}
        \begin{tabular}{c|c|c|c||c|c|c|c|c|c}
        & \multicolumn{3}{c||}{Voxel}
        & \multicolumn{3}{c}{NTFA}
        & \multicolumn{3}{|c}{HTFA}\\
        \midrule
            \textbf{Subject}      
            & \textbf{Heights}
            & \textbf{Social}
            & \textbf{Spiders}
            & \textbf{Heights}
            & \textbf{Social}
            & \textbf{Spiders}
            & \textbf{Heights}
            & \textbf{Social}
            & \textbf{Spiders}
            \\           
        \midrule
            	4
            	&  .90 $\pm$ .03	
            	&  .94 $\pm$ .04
            	&  .96 $\pm$ .06 
            	& \textbf{.86 $\pm$ .01}	
            	& .89 $\pm$ .09	
            	& \textbf{1.00 $\pm$ .00}  
            	& .84 $\pm$ .07
            	& \textbf{.90 $\pm$ .10}
            	& .86 $\pm$ .06
            	\\   
            	5
            	&  .91 $\pm$ .08	
            	&  .97 $\pm$ .03
            	&  1.00 $\pm$ .00 
            	& \textbf{.85 $\pm$ .03}
            	& \textbf{.91 $\pm$ .11}	
            	& \textbf{.97 $\pm$ .03}  
            	& .58 $\pm$ .10
            	& .62 $\pm$ .17
            	& .76 $\pm$ .15
            	\\   
            	6
            	&  .42 $\pm$ .08
            	&  .26 $\pm$ .09 	
            	&  .54 $\pm$ .08 
            	& \textbf{.45 $\pm$ .11}	
            	& .27 $\pm$ .06
            	& \textbf{.61 $\pm$ .09}  
            	& .30 $\pm$ .14
            	& \textbf{.35 $\pm$ .13}
            	& .58 $\pm$ .15
            	\\             	
            	7           
            	&  .44 $\pm$ .03
            	&  .38 $\pm$ .06 	
            	&  .39 $\pm$ .05 
            	& .44 $\pm$ .00	
            	& \textbf{.44 $\pm$ .09}	
            	& \textbf{.36 $\pm$ .00}  
            	& \textbf{.59 $\pm$ .06}
            	& .25 $\pm$ .00
            	& .30 $\pm$ .11
            	\\            	
            	8
            	&  .99 $\pm$ .01	
            	&  .86 $\pm$ .10 	
            	&  1.00 $\pm$ .00 
            	& \textbf{.96 $\pm$ .01}
            	& \textbf{.91 $\pm$ .05}	
            	& \textbf{.99 $\pm$ .01}
            	& .79 $\pm$ .10
            	& .80 $\pm$ .04
            	& .96 $\pm$ .03
            	\\   
            	9
            	&  .56 $\pm$ .17	
            	&  .46 $\pm$ .05 	
            	&  .57 $\pm$ .15 
            	& .52 $\pm$ .12	
            	& .36 $\pm$ .06	
            	& .42 $\pm$ .13  
            	& \textbf{.74 $\pm$ .10}
            	& \textbf{.77 $\pm$ .16}
            	& \textbf{.64 $\pm$ .10}
            	\\             	
            	10
            	&  .97 $\pm$ .04	
            	&  .96 $\pm$ .03 	
            	&  1.0 $\pm$ .00
            	& \textbf{.99 $\pm$ .01}	
            	& \textbf{.95 $\pm$ .05}	
            	& \textbf{1.0 $\pm$ .00}  
            	& .93 $\pm$ .04
            	& .94 $\pm$ .04
            	& .95 $\pm$ .04
            	\\           	
            	11
            	&  .86 $\pm$ .13	
            	&  .93 $\pm$ .03 	
            	&  .94 $\pm$ .07 
            	& \textbf{.85 $\pm$ .14}	
            	& \textbf{.90 $\pm$ .03}	
            	& \textbf{.86 $\pm$ .08}  
            	& .66 $\pm$ .04
            	& .84 $\pm$ .08
            	& .69 $\pm$ .17
            	\\             	
            	12
            	&  .36 $\pm$ .19	
            	&  .67 $\pm$ .06 	
            	&  .65 $\pm$ .18 
            	& \textbf{.49 $\pm$ .08}	
            	& .53 $\pm$ .10	
            	& \textbf{.61 $\pm$ .17}  
            	& .38 $\pm$ .22
            	& \textbf{.60 $\pm$ .08}
            	& .51 $\pm$ .11
            	\\             	
            	13
            	&  .88 $\pm$ .09	
            	&  .99 $\pm$ .02 	
            	&  .85 $\pm$ .11 
            	& \textbf{.91 $\pm$ .08}	
            	& \textbf{.98 $\pm$ .02}	
            	& \textbf{.86 $\pm$ .10}  
            	& .80 $\pm$ .15
            	& .87 $\pm$ .03
            	& .86 $\pm$ .03
            	\\   
            	14
            	&  .85 $\pm$ .03	
            	&  .89 $\pm$ .05 	
            	&  .97 $\pm$ .03 
            	& .81 $\pm$ .00	
            	& .89 $\pm$ .02	
            	& \textbf{1.0 $\pm$ .00}  
            	& \textbf{.87 $\pm$ .03}
            	& \textbf{.94 $\pm$ .03}
            	& .98 $\pm$ .01
            	\\   
            	15
            	&  .82 $\pm$ .10	
            	&  .87 $\pm$ .09 	
            	&  .92 $\pm$ .08 
            	& \textbf{.84 $\pm$ .07}	
            	& \textbf{.93 $\pm$ .05}	
            	& \textbf{.95 $\pm$ .07}  
            	& .60 $\pm$ .11
            	& .85 $\pm$ .08
            	& .83 $\pm$ .13
            	\\   
            	16
            	&  .96 $\pm$ .03	
            	&  .99 $\pm$ .01 	
            	&  .95 $\pm$ .01 
            	& \textbf{.96 $\pm$ .03}	
            	& .99 $\pm$ .01	
            	& \textbf{.96 $\pm$ .03}  
            	& .85 $\pm$ .11
            	& \textbf{1.0 $\pm$ .00}
            	& .93 $\pm$ .05
            	\\   
            	17
            	&  1.0 $\pm$ .00	
            	&  .92 $\pm$ .05 	
            	&  .83 $\pm$ .07 
            	& .91 $\pm$ .04	
            	& \textbf{.83 $\pm$ .04}	
            	& \textbf{.88 $\pm$ .04}  
            	& \textbf{.95 $\pm$ .02}
            	& .47 $\pm$ .12
            	& .81 $\pm$ .15
            	\\   
            	18
            	&  .81 $\pm$ .14	
            	&  .68 $\pm$ .11 	
            	&  .81 $\pm$ .08 
            	& .63 $\pm$ .14	
            	& .58 $\pm$ .17	
            	& .59 $\pm$ .07  
            	& \textbf{.68 $\pm$ .07}
            	& \textbf{.59 $\pm$ .08}
            	& \textbf{.70 $\pm$ .05}
            	\\   
            	19
            	&  .85 $\pm$ .01	
            	&  .94 $\pm$ .04 	
            	&  .85 $\pm$ .09 
            	& .89 $\pm$ .03	
            	& \textbf{.93 $\pm$ .00}	
            	& \textbf{.84 $\pm$ .08}  
            	& \textbf{.88 $\pm$ .04}
            	& .81 $\pm$ .04
            	& .67 $\pm$ .14
            	\\   
            	23
            	&  .70 $\pm$ .05	
            	&  .85 $\pm$ .06 	
            	&  .72 $\pm$ .07 
            	& \textbf{.81 $\pm$ .03}	
            	& \textbf{.81 $\pm$ .03}	
            	& \textbf{.85 $\pm$ .07}  
            	& .58 $\pm$ .02
            	& .66 $\pm$ .09
            	& .75 $\pm$ .09
            	\\   
            	25
            	&  .84 $\pm$ .06	
            	&  .95 $\pm$ .03 	
            	&  .99 $\pm$ .01 
            	& \textbf{.76 $\pm$ .04}	
            	& .93 $\pm$ .07	
            	& \textbf{.99 $\pm$ .01}  
            	& .61 $\pm$ .12
            	& \textbf{.94 $\pm$ .02}
            	& .90 $\pm$ .04
            	\\   
            	26
            	&  .84 $\pm$ .16	
            	&  .53 $\pm$ .03 	
            	&  .82 $\pm$ .10 
            	& .64 $\pm$ .11	
            	& \textbf{.76 $\pm$ .05}	
            	& .81 $\pm$ .09  
            	& \textbf{.77 $\pm$ .05}
            	& .59 $\pm$ .00
            	& \textbf{.95 $\pm$ .05}
            	\\   
            	28
            	&  .78 $\pm$ .07	
            	&  .97 $\pm$ .03 	
            	&  1.0 $\pm$ .00 
            	& \textbf{.96 $\pm$ .05}	
            	& \textbf{.97 $\pm$ .00}	
            	& \textbf{1.0 $\pm$ .00}  
            	& .83 $\pm$ .15
            	& .87 $\pm$ .05
            	& .99 $\pm$ .01
            	\\   
            	29
            	&  .95 $\pm$ .07	
            	&  .94 $\pm$ .04 	
            	&  1.0 $\pm$ .00 
            	& .87 $\pm$ .09	
            	& .93 $\pm$ .04	
            	& .93 $\pm$ .05  
            	& \textbf{.90 $\pm$ .03}
            	& \textbf{.95 $\pm$ .01}
            	& \textbf{.95 $\pm$ .04}
            	\\   
        \midrule  
        \end{tabular}
    \end{center}
    }

    \end{table*}

\begin{table}[!h]
    {
\caption{\textbf{Classification Details} on Lepping dataset. ``Voxel'' indicates the ANOVA-SVM strategy on input data. NTFA performs consistently better than HTFA and closer to performing a completely supervised feature selection + classification pipeline on the input data.}
\label{table:svm_lepping}
\tiny
    \begin{center}
    \setlength\tabcolsep{3.0pt}
        \begin{tabular}{c|c|c|c|c||c|c|c|c|c|c|c|c}
        & \multicolumn{4}{c||}{Voxel}
        & \multicolumn{4}{c}{NTFA}
        & \multicolumn{4}{|c}{HTFA}\\
        \midrule
            \textbf{Subject}   
            & \textbf{-Music}
            & \textbf{-Sounds}
            & \textbf{+Music}
            & \textbf{+Sounds}
            & \textbf{-Music}
            & \textbf{-Sounds}
            & \textbf{+Music}
            & \textbf{+Sounds}
            & \textbf{-Music}
            & \textbf{-Sounds}
            & \textbf{+Music}
            & \textbf{+Sounds}
            \\           
        \midrule
 
            	CTRL-1
            	&  .75 $\pm$ .20	
            	&  .42 $\pm$ .43 	
            	&  .58 $\pm$ .24 
            	&  .79 $\pm$ .12
            	
             	&  .75 $\pm$ .20	
            	&  .29 $\pm$ .21  
            	&  \textbf{.96 $\pm$ .06} 
            	&  .58 $\pm$ .31
            	
            	&  \textbf{.96 $\pm$ .06}
            	&  \textbf{.37 $\pm$ .27}
            	&  .92 $\pm$ .12
            	&  \textbf{.83 $\pm$ .24}
            	\\  
            	CTRL-2
            	&   .62 $\pm$ .27
            	&   .92 $\pm$ .12  	
            	&   .92 $\pm$ .12 
            	&   .54 $\pm$ .06	
            	
            	&   \textbf{.79 $\pm$ .21}	
            	&   \textbf{1.0 $\pm$ .00}  
            	&   \textbf{.79 $\pm$ .16} 
            	&   \textbf{.75 $\pm$ .20}
            	
            	&   .67 $\pm$ .31
            	&   .92 $\pm$ .12
            	&   .71 $\pm$ .21
            	&   .58 $\pm$ .24
            	\\
            	CTRL-3
            	&   .12 $\pm$ .10 	
            	&   .54 $\pm$ .29 	
            	&   .71 $\pm$ .21
            	&   .29 $\pm$ .21
            	
            	&   .58 $\pm$ .12	
            	&   .79 $\pm$ .29  
            	&   \textbf{.87 $\pm$ .10} 
            	&   \textbf{.58 $\pm$ .12}
            	
            	&   \textbf{.71 $\pm$ .21}
            	&   \textbf{.83 $\pm$ .12}
            	&   .79 $\pm$ .212
            	&   .42 $\pm$ .12
            	\\   
            	CTRL-4
            	&   .79 $\pm$ .16 	
            	&   .79 $\pm$ .16  	
            	&   .67 $\pm$ .12
            	&   .67 $\pm$ .12	
            	
            	&   \textbf{.92 $\pm$ .12}	
            	&   \textbf{.62 $\pm$ .10}  
            	&   .75 $\pm$ .20 
            	&   \textbf{.67 $\pm$ .12}
            	
            	&   .87 $\pm$ .18
            	&   .58 $\pm$ .12
            	&   \textbf{.83 $\pm$ .12}
            	&   .58 $\pm$ .12
            	\\   
            	CTRL-5
            	&   .71 $\pm$ .25 	
            	&   .54 $\pm$ .33 	
            	&   .83 $\pm$ .24 
            	&   .62 $\pm$ .10 
            	
            	&   .67 $\pm$ .12	
            	&   \textbf{.67 $\pm$ .31}  
            	&   \textbf{.87 $\pm$ .12} 
            	&   .62 $\pm$ .18
            	
            	&   \textbf{.75 $\pm$ .20}
            	&   \textbf{.67 $\pm$ .24} 
            	&   .83 $\pm$ .20
            	&   \textbf{.83 $\pm$ .12}
            	\\             	
            	CTRL-6           
            	&   .50 $\pm$ .41
            	&   .79 $\pm$ .16  	
            	&   .67 $\pm$ .24 
            	&   .83 $\pm$ .24 
            	
            	&   .42 $\pm$ .24	
            	&   \textbf{.83 $\pm$ .12}  
            	&   .92 $\pm$ .12
            	&   \textbf{1.0 $\pm$ .00}
            	
            	&   \textbf{.58 $\pm$ .42}
            	&   .75 $\pm$ .20
            	&   \textbf{.96 $\pm$ .06}
            	&   \textbf{1.0 $\pm$ .00}
            	\\            	
            	CTRL-7
            	&   .79 $\pm$ .16 	
            	&   .42 $\pm$ .42  	
            	&   .58 $\pm$ .12 
            	&   .50 $\pm$ .20 
            	
            	&   \textbf{.92 $\pm$ .12}	
            	&   .42 $\pm$ .42  
            	&   .33 $\pm$ .12 
            	&   \textbf{.25 $\pm$ .20}
            	
            	&   .83 $\pm$ .12
            	&   \textbf{.46 $\pm$ .41}
            	&   \textbf{.50 $\pm$ .20}
            	&   .17 $\pm$ .24
            	\\   
            	CTRL-8
            	&   1.0 $\pm$ .00 	
            	&   .83 $\pm$ .12  	
            	&   .86 $\pm$ .10
            	&   .75 $\pm$ .20 
            	
            	&   \textbf{1.0 $\pm$ .00}	
            	&   .75 $\pm$ .00  
            	&   \textbf{1.0 $\pm$ .00} 
            	&   \textbf{1.0 $\pm$ .00}
            	
            	&   .67 $\pm$ .12
            	&   \textbf{.92 $\pm$ .12}
            	&   \textbf{1.0 $\pm$ .00}
            	&   .92 $\pm$ .12
            	\\             	
            	CTRL-9
            	&   .58 $\pm$ .12 	
            	&   .96 $\pm$ .06  	
            	&   .71 $\pm$ .21 
            	&   .50 $\pm$ .00 
            	
            	&   .42 $\pm$ .31	
            	&   \textbf{.92 $\pm$ .12}  
            	&   \textbf{.92 $\pm$ .12} 
            	&   \textbf{.67 $\pm$ .12}
            	
            	&   \textbf{.75 $\pm$ .00}
            	&   \textbf{.92 $\pm$ .12}
            	&   .87 $\pm$ .10
            	&   .58 $\pm$ .24
            	\\           	
            	CTRL-10
            	&   .62 $\pm$ .31 	
            	&   .87 $\pm$ .10  	
            	&   .37 $\pm$ .10 
            	&   .17 $\pm$ .12 
            	
            	&   .37 $\pm$ .44	
            	&   \textbf{.83 $\pm$ .12}  
            	&   .25 $\pm$ .20
            	&   .21 $\pm$ .16
            	
            	&   \textbf{.92 $\pm$ .12}
            	&   \textbf{.83 $\pm$ .12}
            	&   \textbf{.75 $\pm$ .20}
            	&   \textbf{.50 $\pm$ .20}
            	\\             	
            	CTRL-11
            	&   .46 $\pm$ .21 	
            	&   .87 $\pm$ .18  	
            	&   .54 $\pm$ .16 
            	&   1.0 $\pm$ .00	
            	
            	&   .33 $\pm$ .12	
            	&   \textbf{1.0 $\pm$ .00}  
            	&   .33 $\pm$ .12
            	&   \textbf{1.0 $\pm$ .00}
            	
            	&   \textbf{.58 $\pm$ .12}
            	&   .79 $\pm$ .16
            	&   \textbf{.50 $\pm$ .20}
            	&   \textbf{1.0 $\pm$ .00}
            	\\             	
            	CTRL-12
            	&   .83 $\pm$ .12 	
            	&   .67 $\pm$ .24  	
            	&   .87 $\pm$ .10 
            	&   .92 $\pm$ .12 
            	
            	&   \textbf{.75 $\pm$ .20}	
            	&   \textbf{.67 $\pm$ .31}  
            	&   \textbf{.83 $\pm$ .12} 
            	&   .71 $\pm$ .06
            	
            	&   \textbf{.75 $\pm$ .20}
            	&   .58 $\pm$ .12
            	&   .79 $\pm$ .83 
            	&   \textbf{.83 $\pm$ .24}
            	\\   
            	CTRL-13
            	&   .71 $\pm$ .26 	
            	&   .67 $\pm$ .31  	
            	&   .92 $\pm$ .12 
            	&   .29 $\pm$ .26
            	
            	&   \textbf{.92 $\pm$ .12}	
            	&   \textbf{.75 $\pm$ .35}  
            	&   \textbf{.75 $\pm$ .20} 
            	&   .29 $\pm$ .26
            	
            	&   .67 $\pm$ .24
            	&   .25 $\pm$ .20 
            	&   .71 $\pm$ .21
            	&   \textbf{.75 $\pm$ .20}
            	\\   
            	CTRL-14
            	&   .58 $\pm$ .31 	
            	&   .67 $\pm$ .31  	
            	&   .33 $\pm$ .12 
            	&   .83 $\pm$ .12 
            	
            	&   .62 $\pm$ .31	
            	&   \textbf{.87 $\pm$ .10}  
            	&   .33 $\pm$ .12
            	&   .75 $\pm$ .20
            	
            	&   \textbf{.67 $\pm$ .24}
            	&   .46 $\pm$ .16
            	&   \textbf{.42 $\pm$ .12} 
            	&   \textbf{.87 $\pm$ .18}
            	\\   
            	CTRL-15
            	&   .62 $\pm$ .18 	
            	&   .87 $\pm$ .18  	
            	&   .58 $\pm$ .31 
            	&   .92 $\pm$ .12	
            	
            	&   \textbf{.58 $\pm$ .24}	
            	&   .67 $\pm$ .47  
            	&   \textbf{.50 $\pm$ .20} 
            	&   .62 $\pm$ .10
            	
            	&   \textbf{.58 $\pm$ .12}
            	&   \textbf{.79 $\pm$ .29}
            	&   .08 $\pm$ .12
            	&   \textbf{.87 $\pm$ .18}
            	\\   
            	CTRL-16
            	&   .87 $\pm$ .18 	
            	&   .79 $\pm$ .21  	
            	&   .46 $\pm$ .16 
            	&   .58 $\pm$ .31 
            	
            	&   \textbf{.92 $\pm$ .12}	
            	&   .67 $\pm$ .12  
            	&   .33 $\pm$ .31 
            	&   \textbf{.50 $\pm$ .20}
            	
            	&   .87 $\pm$ .10
            	&   \textbf{.83 $\pm$ .12}
            	&   \textbf{.50 $\pm$ .41}
            	&   .42 $\pm$ .24
            	\\   
            	CTRL-17
            	&   .67 $\pm$ .31 	
            	&   .25 $\pm$ .20  	
            	&   .79 $\pm$ .16
            	&   .62 $\pm$ .10 
            	
            	&   .37 $\pm$ .10	
            	&   .21 $\pm$ .21  
            	&   \textbf{.83 $\pm$ .12} 
            	&   .37 $\pm$ .27
            	
            	&   \textbf{.71 $\pm$ .26}
            	&   \textbf{.29 $\pm$ .26} 
            	&   .79 $\pm$ .16
            	&   \textbf{.79 $\pm$ .21}
            	\\   
            	CTRL-18
            	&   .71 $\pm$ .21 	
            	&   .62 $\pm$ .37  	
            	&   .83 $\pm$ .12 
            	&   .83 $\pm$ .12 
            	
            	&   .83 $\pm$ .24	
            	&   .67 $\pm$ .12  
            	&   \textbf{.92 $\pm$ .12} 
            	&   \textbf{.62 $\pm$ .17}
            	
            	&   \textbf{.92 $\pm$ .12}
            	&   \textbf{1.0 $\pm$ .00}
            	&   \textbf{.92 $\pm$ .12}
            	&   .46 $\pm$ .33
            	\\   
            	CTRL-19
            	&   .87 $\pm$ .10 	
            	&   1.0 $\pm$ .00  	
            	&   .54 $\pm$ .36 
            	&   .92 $\pm$ .12 
            	
            	&   \textbf{.96 $\pm$ .06}	
            	&   .75 $\pm$ .202  
            	&   \textbf{.50 $\pm$ .41} 
            	&   \textbf{.92 $\pm$ .12}
            	
            	&   .83 $\pm$ .24
            	&   \textbf{.96 $\pm$ .06}
            	&   .33 $\pm$ .24
            	&   .58 $\pm$ .24
            	\\ 
            	CTRL-20
            	&   .83 $\pm$ .24 	
            	&   .83 $\pm$ .12  	
            	&   .54 $\pm$ .33 
            	&   .50 $\pm$ .00 
            	
            	&   \textbf{.79 $\pm$ .16}	
            	&   \textbf{.87 $\pm$ .12}  
            	&   .42 $\pm$ .42 
            	&   .37 $\pm$ .10
            	
            	&   .58 $\pm$ .31
            	&   .67 $\pm$ .12
            	&   \textbf{.46 $\pm$ .39} 
            	&   \textbf{.62 $\pm$ .31}
            	\\ 
            	MDD-1
            	&   .54 $\pm$ .36	
            	&   .75 $\pm$ .20  	
            	&   .83 $\pm$ .12 
            	&   .08 $\pm$ .12 
            	
            	&   \textbf{.79 $\pm$ .16}	
            	&   \textbf{.75 $\pm$ .20}  
            	&   .833 $\pm$ .12 
            	&   .17 $\pm$ .12
            	
            	&   .67 $\pm$ .12
            	&   .67 $\pm$ .31
            	&   \textbf{1.0 $\pm$ .00}
            	&   .17 $\pm$ .12
            	\\   
            	MDD-2
            	&   .67 $\pm$ .24 	
            	&   .25 $\pm$ .00  	
            	&   .83 $\pm$ .24 
            	&   .67 $\pm$ .31 
            	
            	&   \textbf{.67 $\pm$ .31}	
            	&   .17 $\pm$ .12  
            	&   .75 $\pm$ .20 
            	&   .67 $\pm$ .31
            	
            	&   \textbf{.67 $\pm$ .47}
            	&   \textbf{.42 $\pm$ .24}
            	&   \textbf{.79 $\pm$ .29}
            	&   \textbf{.71 $\pm$ .06}
            	\\   
            	MDD-3
            	&   .11 $\pm$ .16 	
            	&   1.0 $\pm$ .00  	
            	&   .58 $\pm$ .31 
            	&   .25 $\pm$ .35 
            	
            	&   \textbf{.44 $\pm$ .34}	
            	&   \textbf{.89 $\pm$ .16}  
            	&   .56 $\pm$ .42 
            	&   \textbf{.25 $\pm$ .20}
            	
            	&   .19 $\pm$ .14
            	&   \textbf{.89 $\pm$ .16}
            	&   \textbf{.64 $\pm$ .31}
            	&   .00 $\pm$ .00
            	\\             	
            	MDD-4           
            	&   .79 $\pm$ .16 	
            	&   .62 $\pm$ .10  	
            	&   .75 $\pm$ .20 
            	&   .83 $\pm$ .12 
            	
            	&   \textbf{.83 $\pm$ .12}	
            	&   .58 $\pm$ .31  
            	&   \textbf{.75 $\pm$ .20} 
            	&   .75 $\pm$ .00
            	
            	&   .71 $\pm$ .06
            	&   \textbf{.67 $\pm$ .24}
            	&   .54 $\pm$ .33
            	&   \textbf{.87 $\pm$ .10}
            	\\
            	MDD-6
            	&   .67 $\pm$ .31 	
            	&   .58 $\pm$ .42  	
            	&   .87 $\pm$ .10 
            	&   .92 $\pm$ .12 
            	
            	&   .67 $\pm$ .31	
            	&   \textbf{.75 $\pm$ .35}  
            	&   .92 $\pm$ .12
            	&   .62 $\pm$ .31
            	
            	&   \textbf{.71 $\pm$ .21}
            	&   .62 $\pm$ .18
            	&   \textbf{.96 $\pm$ .06}
            	&   \textbf{.79 $\pm$ .21}
            	\\             	
            	MDD-7
            	&   .54 $\pm$ .16 	
            	&   .79 $\pm$ .16  	
            	&   .42 $\pm$ .31 
            	&   .58 $\pm$ .31 
            	
            	&   .54 $\pm$ .21	
            	&   \textbf{.83 $\pm$ .12}  
            	&   \textbf{.75 $\pm$ .00} 
            	&   \textbf{.62 $\pm$ .10}
            	
            	&   \textbf{.62 $\pm$ .10}
            	&   .58 $\pm$ .12
            	&   .71 $\pm$ .06
            	&   .42 $\pm$ .31
            	\\           	
            	MDD-8
            	&   .83 $\pm$ .12 	
            	&   .58 $\pm$ .31  	
            	&   .79 $\pm$ .21 
            	&   .42 $\pm$ .31
            	
            	&   \textbf{.92 $\pm$ .19}	
            	&   .42 $\pm$ .12  
            	&   .71 $\pm$ .33 
            	&   \textbf{.33 $\pm$ .471}
            	
            	&   .87 $\pm$ .10
            	&   \textbf{.54 $\pm$ .26}
            	&   \textbf{.83 $\pm$ .12}
            	&   \textbf{.33 $\pm$ .47}
            	\\             	
            	MDD-9
            	&   1.0 $\pm$ .00 	
            	&   .67 $\pm$ .12  	
            	&   .87 $\pm$ .18
            	&   .96 $\pm$ .06 
            	
            	&   \textbf{1.0 $\pm$ .00}	
            	&   .25 $\pm$ .20  
            	&   \textbf{1.0 $\pm$ .00}
            	&   \textbf{.92 $\pm$ .12}
            	
            	&   \textbf{1.0 $\pm$ .00}
            	&   \textbf{.29 $\pm$ .33}
            	&   .75 $\pm$ .20
            	&   .83 $\pm$ .12
            	\\             	
            	MDD-10
            	&   1.0 $\pm$ .00 	
            	&   .87 $\pm$ .18  	
            	&   .83 $\pm$ .24 
            	&   .58 $\pm$ .12 
            	
            	&   \textbf{1.0 $\pm$ .00}	
            	&   \textbf{.92 $\pm$ .12}  
            	&   \textbf{.83 $\pm$ .24} 
            	&   .50 $\pm$ .00
            	
            	&   .92 $\pm$ .12
            	&   .87 $\pm$ .18
            	&   .79 $\pm$ .21
            	&   \textbf{.71 $\pm$ .21}
            	\\   
            	MDD-11
            	&   .62 $\pm$ .10 	
            	&   .67 $\pm$ .12  	
            	&   .79 $\pm$ .21
            	&   .83 $\pm$ .12 
            	
            	&   \textbf{.54 $\pm$ .16}	
            	&   \textbf{.83 $\pm$ .12}  
            	&   .87 $\pm$ .18 
            	&   \textbf{.79 $\pm$ .16}
            	
            	&   .46 $\pm$ .06
            	&   .37 $\pm$ .31
            	&   \textbf{1.0 $\pm$ .00}
            	&   .75 $\pm$ .20
            	\\   
            	MDD-12
            	&   .83 $\pm$ .24 	
            	&   .42 $\pm$ .12  	
            	&   .58 $\pm$ .12
            	&   .79 $\pm$ .16
            	
            	&   \textbf{.71 $\pm$ .33}	
            	&   \textbf{.33 $\pm$ .12}  
            	&   \textbf{.71 $\pm$ .21} 
            	&   \textbf{.58 $\pm$ .12}
            	
            	&   .46 $\pm$ .16
            	&   .17 $\pm$ .12
            	&   .67 $\pm$ .12
            	&   .37 $\pm$ .10
            	\\   
            	MDD-13
            	&   .67 $\pm$ .31 	
            	&   .71 $\pm$ .26  	
            	&   .25 $\pm$ .00 
            	&   .58 $\pm$ .12 
            	
            	&   .92 $\pm$ .12	
            	&   \textbf{1.0 $\pm$ .00}  
            	&   .42 $\pm$ .12 
            	&   .58 $\pm$ .12
            	
            	&   \textbf{1.0 $\pm$ .00}
            	&   .92 $\pm$ .12
            	&   \textbf{.58 $\pm$ .12}
            	&   \textbf{.75 $\pm$ .20}
            	\\   
            	MDD-14
            	&   .58 $\pm$ .24 	
            	&   .83 $\pm$ .12  	
            	&   .58 $\pm$ .31
            	&   .58 $\pm$ .12
            	
            	&   .33 $\pm$ .12	
            	&   \textbf{.79 $\pm$ .16}  
            	&   .50 $\pm$ .20 
            	&   \textbf{.67 $\pm$ .31}
            	
            	&   \textbf{.37 $\pm$ .10}
            	&   .62 $\pm$ .10
            	&   \textbf{.67 $\pm$ .12}
            	&   .58 $\pm$ .24
            	\\   
            	MDD-15
            	&   1.0 $\pm$ .00 	
            	&   .58 $\pm$ .24  	
            	&   .75 $\pm$ .20 
            	&   .58 $\pm$ .12 
            	
            	&   \textbf{.83 $\pm$ .12}	
            	&   \textbf{.75 $\pm$ .20}  
            	&   \textbf{.83 $\pm$ .24} 
            	&   \textbf{.75 $\pm$ .20}
            	
            	&   .58 $\pm$ .31
            	&   .50 $\pm$ .41
            	&   .79 $\pm$ .21
            	&   .67 $\pm$ .12
            	\\   
            	MDD-16
            	&   .92 $\pm$ .12 	
            	&   .46 $\pm$ .39  	
            	&   .29 $\pm$ .26 
            	&   .37 $\pm$ .10 
            	
            	&   \textbf{.96 $\pm$ .06}	
            	&   .46 $\pm$ .39   
            	&   \textbf{.54 $\pm$ .39} 
            	&   \textbf{.54 $\pm$ .06}
            	
            	&   \textbf{.96 $\pm$ .06}
            	&   \textbf{.62 $\pm$ .31}
            	&   .42 $\pm$ .24
            	&   .37 $\pm$ .10
            	\\   
            	MDD-17
            	&   1.0 $\pm$ .00 	
            	&   .21 $\pm$ .06 	
            	&   .67 $\pm$ .24 
            	&   .75 $\pm$ .20 
            	
            	&   \textbf{1.0 $\pm$ .00}	
            	&   \textbf{.25 $\pm$ .20}  
            	&   .50 $\pm$ .00 
            	&   .83 $\pm$ .12
            	
            	&   .87 $\pm$ .18
            	&   \textbf{.25 $\pm$ .00}
            	&   \textbf{.79 $\pm$ .21}
            	&   \textbf{1.0 $\pm$ .00}
            	\\   
            	MDD-18
            	&   .83 $\pm$ .24
            	&   1.0 $\pm$ .00  	
            	&   .75 $\pm$ .00 
            	&   .92 $\pm$ .12
            	
            	&   \textbf{.83 $\pm$ .24}	
            	&   \textbf{.33 $\pm$ .24}  
            	&   \textbf{.92 $\pm$ .12} 
            	&   \textbf{.92 $\pm$ .12}
            	
            	&   .42 $\pm$ .12
            	&   .29 $\pm$ .21
            	&   .83 $\pm$ .12
            	&   .58 $\pm$ .31
            	\\   
            	MDD-19
            	&   .75 $\pm$ .00 	
            	&   .62 $\pm$ .18  	
            	&   .58 $\pm$ .24 
            	&   .92 $\pm$ .12 	
            	
            	&   .75 $\pm$ .00	
            	&   .62 $\pm$ .10  
            	&   \textbf{.75 $\pm$ .00} 
            	&   \textbf{.83 $\pm$ .12}
            	
            	&   \textbf{.92 $\pm$ .12}
            	&   \textbf{.75 $\pm$ .20}
            	&   .58 $\pm$ .12
            	&   .58 $\pm$ .24
            	\\  
        \midrule  
        \end{tabular}
    \end{center}
    }

    \end{table}

\end{document}